\documentclass[letterpaper, 12pt]{article}
\usepackage[pass,paperwidth=8.5in,paperheight=11in]{geometry}
\usepackage{amsmath, amsfonts, amssymb, bm, graphicx, isomath}
\usepackage[round]{natbib}
\usepackage[colorlinks,citecolor=blue,urlcolor=blue]{hyperref}
\usepackage{wrapfig, lscape, rotating, epstopdf}
\usepackage{setspace}

\newcommand{\vect}[1]{\mathbfit{#1}}

\newcommand{\D}[3]{\mathrm{D}_{#1}\!\left[{#2} \vert\vert {#3}\right] }

\newcommand{\tr}{\mathop{\mathrm{tr}}}
\newcommand{\BE}[1]{ \left\langle #1 \right\rangle}
\newcommand{\BR}[1]{ \left( #1 \right)}
\newcommand{\BS}[1]{ \left[ #1 \right]}
\newcommand{\BC}[1]{ \left\{ #1 \right\}}

\newcommand{\sumlow}[1]{\sum\limits_{#1}}

\newcommand{\prodlow}[1]{\prod\limits_{#1}}

\newcommand{\appropto}{\mathrel{\vcenter{
  \offinterlineskip\halign{\hfil$##$\cr
    \propto\cr\noalign{\kern2pt}\sim\cr\noalign{\kern-2pt}}}}}

\title{Sparse Approximate Inference for Spatio-Temporal Point Process Models}
\usepackage{authblk}
\author[1]{Botond Cseke\thanks{Email: bcseke@inf.ed.ac.uk. B. Cs. is funded by BBSRC under grant BB/I004777/1.}}
\author[2]{Andrew Zammit-Mangion\thanks{Email: a.zammitmangion@bristol.ac.uk.  A. Z.-M. is funded by NERC under grant NE/I027401/1.}}
\author[3]{Tom Heskes\thanks{Email: t.heskes@science.ru.nl. T.H. is a professor in Artificial Intelligence at R.U.N.}}
\author[1]{Guido Sanguinetti\thanks{Email: gsanguin@inf.ed.ac.uk. G.S. is funded by ERC under grant MLCS-306999.}}
\affil[1]{School of Informatics, University of Edinburgh}
\affil[2]{School of Geographical Sciences, University of Bristol}
\affil[3]{Faculty of Science, Radboud University Nijmegen}

\begin{document}


\maketitle

\newpage

\begin{abstract}
Spatio-temporal point process models play a central role in the analysis of spatially distributed systems in several disciplines. Yet, scalable inference remains computationally challenging both due to the high resolution modelling generally required and the analytically intractable likelihood function. Here, we exploit the sparsity structure typical of (spatially) discretised log-Gaussian Cox process models by using approximate message-passing algorithms. The proposed algorithms scale well with the state dimension and the length of the temporal horizon with moderate loss in distributional accuracy. They hence provide a flexible and faster alternative to both non-linear filtering-smoothing type algorithms and to approaches that implement the Laplace method or expectation propagation on (block) sparse latent Gaussian models. We infer the parameters of the latent Gaussian model using  a structured variational Bayes approach. We demonstrate the proposed framework on simulation studies with both Gaussian and point-process observations and use it to reconstruct the conflict intensity and dynamics in Afghanistan from the WikiLeaks Afghan War Diary.

\end{abstract}

\section{Introduction}\label{SecIntro}

Dynamic models of spatially distributed point processes are widespread in scientific applications of computational statistics, ranging from environmental sciences \citep{Wikle_2001} to epidemiology \citep{Diggle_2005, Ahn2014} and ecology \citep{Hooten_2008} to name but a few. The prevalence of such data is dramatically increasing due to advances in remote sensing technologies, and novel application domains are fast emerging in the social sciences due to the large scale data sets collected, for example through social networks. Log-Gaussian Cox processes (LGCPs) introduced in \citep{Moller_1998} are an important modelling paradigm for such systems, due to their ability to elegantly explain event-based data through the introduction of an auxiliary latent Gaussian field.

Despite their importance, inference in LGCPs remains computationally challenging. Markov chain Monte Carlo (MCMC) is frequently employed and has desirable asymptotic properties; 
however, despite considerable advances \citep{ADH-PMCMC2010, GiroRiemann2011, Yuan2012}, the computational costs of sampling approaches remain high for high-dimensional latent fields, and in the presence of heterogeneous data sets. 
Deterministic approximations can provide a computationally effective alternative for computing the posterior distribution over the latent field, which is often the computational bottleneck in high dimensions. Current approaches can be broadly classified as 
blocked or dynamic.  Blocked approaches cast the (time-discretised) model as a latent (sparse) Gaussian model with all state variables concatenated into a single large vector, and apply the Laplace method \citep{RueMartinoChopin2009, Lindgren_2011} or a corresponding expectation propagation (EP) algorithm  \citep[e.g.,][]{CsekeHeskes2010b}. The computational cost of inference in the blocked approach is dominated by a sparse partial matrix inversion \citep{Taka1973}; these costs may become untenable for very high dimensions, and it is not always clear what further (robust) approximations could be used to alleviate these problems \citep[e.g.,][]{WisrRue2006, Simpson2013}. Dynamic approaches address the state inference problem using a filtering-smoothing (forward-backward) dynamic programming approach within a variational or EP approximation framework \citep[e.g.,][]{AZM2012b,ypma2005novel,finns2011}.  Due to the non-conjugate likelihood, dynamic approaches also have to resort to approximations, typically exploiting the message passing formulation of inference in graphical models \citep[e.g.,][]{Lauritzen1996, KollerFriedman20009}. The cost of the forward-backward algorithm is typically cubic in the dimension of the state space due to the predictive update step in the Kalman filter. 

In this paper we build on the dynamic approach to inference in spatio-temporal LGCPs, extending it in several ways in order to achieve efficiency in high-dimensional settings. First, we cast the model as a dynamic latent Gaussian model using time discretisation and basis-function projection, the weights of which define the state variables in a latent state-space model representation. Following this we derive a variational, joint state-parameter inference method for approximating the full posterior distribution over the states and unknown parameters. This approximate distribution is factored over states, parameters governing the state-interaction structure, and noise parameters, respectively. In the case of LGCPs, the message-passing algorithm for computing the approximate posterior distribution over the states is not analytically tractable since the likelihood is non-Gaussian. The key contribution of the paper is the derivation of an approximate message-passing algorithm for dealing with this intractability that does not suffer from the computational limitations arising from high-dimensional state spaces. We achieve this by enforcing a sparse structure of the messages, and adopting efficient  sparse linear algebraic methods \citep{Davis2006} in the local computations of the message-passing algorithm. This circumvents the limitation of typical forward-backward algorithms that invariably involve operations that destroy sparsity, for example due to matrix multiplication and marginalisation (as in the Kalman filter's update step).  We show that the approximate message passing scheme we propose is an instance of expectation propagation \citep{Minka2001} that can also be derived from the view of an expectation constrained approximate inference framework \citep{Heskes2005, OpperWinther2005}. 


The method naturally allows for a compromise between computation speed and accuracy. To show this we introduce a class of constraints that result in Gaussian messages having precision structures that are increasingly representative: (i) diagonal (factored messages), (ii) spanning tree (iii) chordal and finally (iv) fully connected (full messages). The latter case (iv) corresponds to the filtering-smoothing type algorithm that uses expectation propagation to cater for the non-Gaussian parts of the model. 
Comparisons in simulated case studies show that the proposed algorithms scale well with state dimension and, depending on the complexity of the messages, we can carry out approximate inference on thousands of state variables and hundreds of time-steps with reasonable time and memory requirements.

The text is structured as follows. In Section~\ref{SecModel} we introduce log-Gaussian Cox processes and present the discretisation and numerical approximation steps that simplify this model to a dynamic latent Gaussian model with non-Gaussian likelihood terms. In Section~\ref{SecInf} we describe the variational inference framework applied to this problem, and derive a class of dynamic message-passing algorithms that exploit the sparsity resulting from the discretisation. In Section~\ref{SecExp} we carry out extensive simulation studies, discuss the performance of these algorithms and use them to extract the micro-dynamics of conflict events in the Afghan war \citep{AZM2012b}. Section \ref{SecConc} concludes the work.

\section{Model}\label{SecModel}

In this paper we are interested in the dynamic modelling of two-dimensional point patterns. The data consists of  location- and time-stamped events $\mathcal{Y} = \{ (\vect{s}_i, {t}_i)\}_{i}$ where the locations $\vect{s}_i$ are points in a two-dimensional compact domain $\mathcal{S} \subset \mathbb{R}\!\times\!\mathbb{R}$ and the time-stamps $t_i$ are in a time interval $\mathcal{T}=[0, {\text{max}}(\{t_i\}_i)]$. In order to model this type of data, we use log-Gaussian Cox process models \citep{Moller_1998} discretised in both space and time. We discretise the domain $\mathcal{S}$ by using a triangular lattice and using the corresponding piecewise linear finite element functions as basis functions. We discretise time by first dividing the time interval $\mathcal{T}$ into $T$ time windows $\{\mathcal{T}_t\}_t$ of equal size $\Delta_t$, that is $\mathcal{T}_t = [t\Delta_t, (t+1)\Delta_t)$ and $T\Delta_t =  {\text{max}}(\{t_i\}_i)$. We then treat the data $\mathcal{Y}$ as a set of spatial point processes indexed by $t$. Specifically, we  let $\mathcal{Y} = \cup_{t} \mathcal{Y}_t$ where each $\mathcal{Y}_t$ contains the (spatial-only) events occurring in the window $\mathcal{T}_t$. The choice of $\Delta_t$ is often determined by the application and although it can be critical from the computational point of view, it is beyond the scope of this article to address this choice in detail.


We define the log intensity function of the point process as a linear combination of the $n$ piecewise linear basis functions $\phi_{j}: \mathcal{S} \rightarrow \mathbb{R}; j = 1,\dots, n$. That is,  $\lambda(\vect{s}, t) \approx \exp\{\vect{x}_{t}^{T}\vect{\phi}(\vect{s})\}$, where $\vect{\phi}(\vect{s}) = (\phi_1(\vect{s}),\dots,\phi_n(\vect{s}))^T$ and the weights (states) $\vect{x}_t \in \mathbb{R}^n$. We further assume that the weights $\vect{x}_{t}$ follow a linear dynamical system  
\begin{equation}\label{eq:LDS}
\vect{x}_{t+1}  =  {\vect{A}}\vect{x}_t  + \vect{\epsilon}_t,
\end{equation}
where $\vect{\epsilon}_t \sim \mathcal{N}({\bf 0},\vect{Q}^{-1})$ with both $\vect{A}$ and $\vect{Q}$ sparse. This linear dynamical model for the log intensity $u(\vect{s}, t) = \log \lambda(\vect{s}, t)$ can be derived from spatio-temporal models commonly employed in practice, such as the integro-difference equation (IDE) \citep{Wikle_2002},  and the stochastic partial differential equation (SPDE) \citep{Zammit_2012}. Sparsity in $\vect{A}$ and $\vect{Q}$ follows either from girdding the domain or from employing a Galerkin reduction on an infinite-dimensional system in $u(\vect{s}, t), \vect{s} \in \mathcal{S}, t \in \mathcal{T}$  using the basis functions $\{\phi_{j}(\vect{s})\}_{j}$.  

\begin{figure}[t]
	\begin{center}
		\begin{tabular}{ccc}
			\resizebox{0.45\textwidth}{!}{\includegraphics{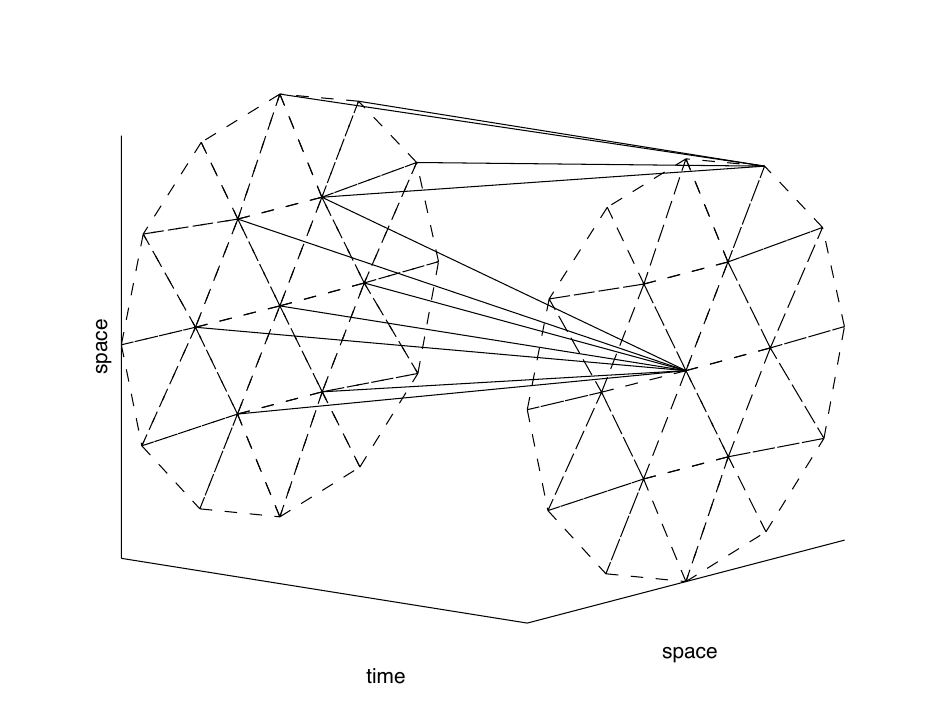}}
			&
			\resizebox{0.45\textwidth}{!}{\includegraphics{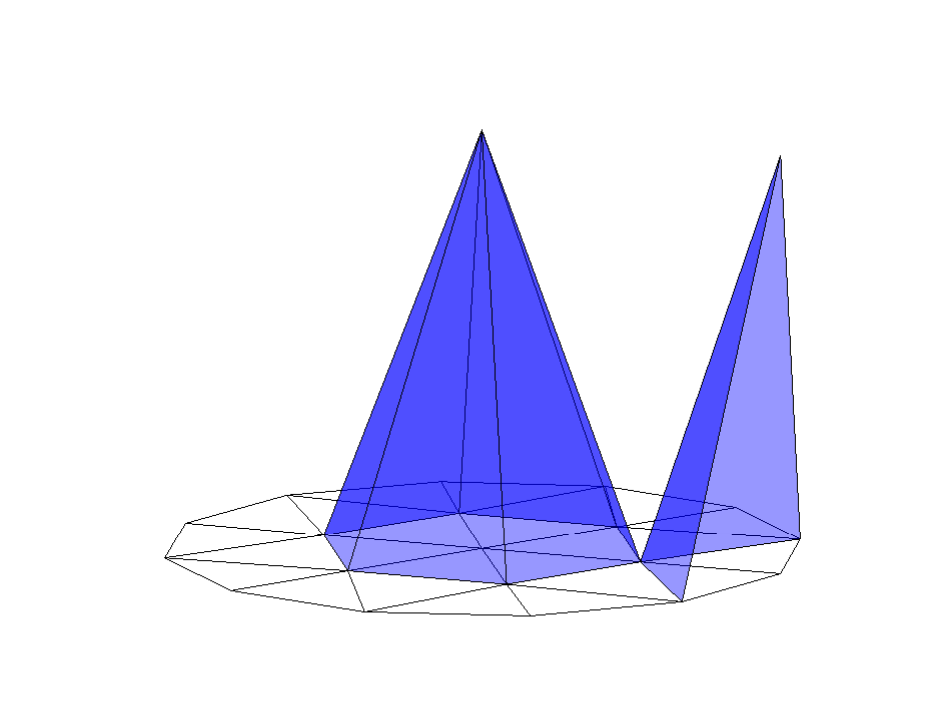}}
		\end{tabular}
	\end{center}
\caption{\small An illustration of the spatio-temporal discretisation employed. The right panel illustrates two basis functions defined according to the triangular finite element spatial discretisation. The bases are shown for two nodes/vertices, one in the interior and one on the boundary of the domain. The left panel illustrates the temporal connectivity for some of the nodes resulting from the spatio-temporal discretisation described in Section~\ref{SecModel}. Similarly, the temporal connectivities are shown only for an interior and a boundary node/vertex.}
\label{FigSTdemo}
\end{figure}

The likelihood of the intensity  $\lambda(\vect{s}, t)$ can be written as
\begin{align*}
	p(\mathcal{Y} \mid \lambda) \propto \exp\Big\{ -\int\limits_{\mathcal{S}\times\mathcal{T}}\!d\vect{s}dt \:\lambda(\vect{s},t) \Big\} \times\prod_{i} \lambda(\vect{s}_i, t_i).
\end{align*}
Using the spatial and temporal discretisation schemes outlined above, we re-write this likelihood as $p(\mathcal{Y} \mid \lambda) \propto \prod_{t} p(\mathcal{Y}_t \mid \vect{x}_t)$ where 
\begin{align}
p(\mathcal{Y}_t|\vect{x}_t) &\appropto \exp\Big\{ -\Delta_t\int\limits_{\mathcal{S}}\!d\vect{s} \:e^{\vect{x}_{t}^{T}\vect{\phi}(\vect{s})} \Big\} \times \prod_{\vect{s} \in \mathcal{Y}_{t} }e^{\vect{x}_{t}^{T}\vect{\phi}(\vect{s})}\nonumber \\
 & = L_1(\vect{x}_t) \times L_2(\vect{x}_t;\mathcal{Y}_t).
\end{align}
This likelihood can be split into two components: the first component $L_1(\vect{x}_t)$ is directly related to the \emph{void probability} of the process. We adopt the approach in \cite{Simpson_2011} and numerically approximate the integral as:
\begin{align}
\log L_1(\vect{x}_t) 
 &\approx -\Delta_t\sum_{j=1}^p\widetilde{\eta}_j\exp(\vect{\phi}^T(\bar{\vect{s}}_j)\vect{x}_t)\nonumber \\
 &= -\vect{\eta}^T\exp(\vect{W}\vect{x}_t), \label{eq:L1}
\end{align}
\noindent where the vector $\vect{\eta}$ denotes the integration weights $\Delta_t \widetilde{\vect{\eta}}$ and the matrix $\vect{W} = [\vect{\phi}(\bar{\vect{s}}_1), \dots, \vect{\phi}(\bar{\vect{s}}_p)]^T$ contains the values of the basis at the chosen $p$ integration points $\{\bar{\vect{s}}_j\}_{j}$. The second component of the likelihood, $L_2(\vect{x}_t;\mathcal{Y}_t)$, denotes the contributions from the observed events and can be represented as
\begin{align}
\log L_2(\vect{x}_t;\mathcal{Y}_t) &= \sum_{\vect{s} \in \mathcal{Y}_t} \vect{ \phi}^T(\vect{s})\vect{x}_t = \vect{h}_t^T\vect{x}_t, \label{eq:L2} 
\end{align}
where  $h_t^j = \sum_{\vect{s} \in \mathcal{Y}_t} \phi_j(\vect{s})$ is the sum of basis functions evaluated at the events' spatial coordinates. The approximate log-likelihood can hence be written, up to a constant, as
\begin{equation} \label{eq:L3}
\log p(\mathcal{Y}_t | \vect{x}_t) \approx - \vect{\eta}^T\exp(\vect{W}\vect{x}_t) + \vect{h}_t^T\vect{x}_t.  
\end{equation}
Both compact basis functions and gridded domains induce sparsity in $\vect{W}$. In particular, if one chooses the integration points to be the vertices of a triangulation or the centres of gridded cells, then $\vect{W}$ simplifies to the identity matrix of size $n \times n$, $\vect{I}_{n \times n}$, where $n=p$. The integration weights $\vect{\eta}$ then correspond to the volumes (scaled by $\Delta_t$) of the basis with unit weight \citep{Simpson_2011}. Note from \eqref{eq:L3} that, since $\vect{W}$ is diagonal, the non-Gaussian terms depend only on $x_{t}^{j}$. In the following we use $\psi_{t,j}({x}_{t}^{j}) = \exp\{-\eta_{j}\exp(x_{t}^{j})\}$ to denote the non-Gaussian component of the likelihood terms $\tilde\psi_{t,j}(x_t^j) = \psi_{t,j}(x_t^j)\exp(h_t^jx_t^j)$ appearing~in~\eqref{eq:L3}.

In our setting $\vect{Q}$ is a diagonal matrix, while we assume that the structure of the transition matrix $\vect{A}$ follows that of the neighbourhood graph that results from the discretisation. The matrix $\vect{A}$ hence describes small-scale, possibly directional, spatio-temporal dynamics, and it is reasonable to assume that $\vect{A}$ only has a select amount of non-zero elements on the neighbourhood structure.  For this reason we  impose a spike and slab prior on these structural elements by introducing a set of binary auxiliary variables $\vect{Z}=(z_{ij})_{i,j}$ with $z_{ij} \in \{0,1\}$ and where $\vect{Z}$, like $\vect{A}$, has the same sparsity structure resulting from the discretisation. We then define the conditional prior $p({a}_{ij} \mid z_{ij}, v_{\text{slab}}) = \mathcal{N}(a_{ij}; 0,v_{\text{slab}})^{z_{ij}}\delta(a_{ij})^{1 - z_{ij}}$ where $\delta(\cdot)$ is the Dirac delta function, and assume a Bernoulli prior $p(z_{ij} \mid p_{\text{slab}}) = {Ber}(z_{ij}; p_{\text{slab}})$ (spike and slab prior). Consequently, we can use the posterior distribution of the variable $z_{ij}$ to quantify the relevance of the transition coefficient $a_{ij}$. We use a Gamma prior ${Gam}(q_{ii}; k, \tau)$ for the diagonal elements of the precision matrix $\vect{Q}$. We conclude our model specification by letting $\vect{x}_1$ be Gaussian with mean $\vect{m}_1$ and covariance matrix $\vect{V}_1$. The hyper-parameters $\vect{\theta} = \{\vect{m}_{1}, \vect{V}_{1}, v_{\text{slab}}, p_{\text{slab}}, k, \tau \}$ are fixed.

With the above assumptions we can write down the joint probabilistic model as
\begin{align}\label{EqnJointModel}
 	p(\mathcal{Y}, \vect{X}, \vect{A}, \vect{Z}, \vect{Q} \mid \vect{\theta}) =\: & p(\vect{x}_{1})\prodlow{t} p(\vect{x}_{t+1}\mid \vect{x}_{t}, \vect{A}, \vect{Q}) \prodlow{j}\tilde{\psi}_{t+1,j}(x_{t+1}^j) 
	\\
	& \times \prodlow{i\sim j} p(a_{ij}\mid z_{ij}, \vect{\theta})p(z_{ij} \mid \vect{\theta}) \prodlow{j} p(q_{jj} \mid \vect{\theta}), \nonumber
\end{align}
where $\vect{X} = \{\vect{x}_t\}_t$ and $\{a_{ij}\}_{i\sim j}$ denotes all the structurally non-zero elements (following from the neighbourhood graph) in $\vect{A}$. The prior distributions are given by
\begin{align*}
 	p(\vect{x}_{1}) = &\: \mathcal{N}(\vect{x}_{1}; \vect{m}_{1}, \vect{V}_1),  
	\\
	p(\vect{x}_{t+1}\mid \vect{x}_{t}, \vect{A}, \vect{Q}) = & \: \mathcal{N}(\vect{x}_{t+1} ;\vect{A}\vect{x}_{t}, \vect{Q}^{-1}),  
	\\
	p({a}_{ij} \mid z_{ij}, v_{\text{slab}}) = & \:  \mathcal{N}(a_{ij}; 0,v_{\text{slab}})^{z_{ij}}\delta(a_{ij})^{1 - z_{ij}},
	\\
	p(z_{ij} \mid \vect{\theta}) = &\: {Ber}(z_{ij}; p_{\text{slab}}),  
	\\
	 p(q_{jj} \mid \vect{\theta}) = & \: {Gam}(q_{jj}; k, \tau).
\end{align*}

%
%

\section{Inference}\label{SecInf}

In many application areas such as the ones mentioned in Section \ref{SecIntro}, the dimension of the state space and the length of the time horizon is in the range of  hundreds or thousands which makes MCMC sampling from the posterior distribution computationally demanding. For this reason, we resort to variational approximate inference methods; we seek structured factorised approximations to the posterior distribution. Variational approximate inference methods formulate inference as an optimisation problem by using the Kullback-Leibler divergence $\D{}{\cdot}{p}$ as optimisation objective \citep[e.g.,][]{Jordan1999}. These methods have been successfully applied in many areas of engineering and machine learning where large scale analytically  intractable probabilistic models are common.  In latent Gaussian models they provide tractable Gaussian approximations to analytically intractable Bayesan posteriors \citep[e.g.,][]{Opper2009, Saul1996}.  

To apply variational inference to our problem, we approximate the posterior distribution $p( \vect{X}, \vect{A}, \vect{Z}, \vect{Q} \mid \mathcal{Y}, \vect{\theta})$ in \eqref{EqnJointModel} with a factored distribution 
\begin{align*} 
q(\vect{X}, \vect{A}, \vect{Z}, \vect{Q} \mid \vect{\theta})= q_{X}(\vect{X}\mid\vect{\theta}) q_{AZ}(\vect{A}, \vect{Z} \mid \vect{\theta}) q_{Q}(\vect{Q} \mid \vect{\theta}),
\end{align*}
 where the factors are the solution to the optimisation 
\begin{align} \label{EqnKL}
	\mathop{\text{minimise}}\limits_{q_{X}, q_{AZ},q_{Q} } \D{}{q_{X}(\vect{X}\mid\vect{\theta}) q_{AZ}(\vect{A}, \vect{Z} \mid \vect{\theta}) q_{Q}(\vect{Q} \mid \vect{\theta})\:}{\:p(\vect{X}, \vect{A}, \vect{Z}, \vect{Q} \mid \mathcal{Y}, \vect{\theta})}.
\end{align}
 To simplify notation, hereafter we omit the dependence of the distribution of interest on the hyper-parameters $\vect{\theta}$.

The optimality conditions of \eqref{EqnKL} can be used to define a component-wise fixed point iteration that is known to correspond to a coordinate descent towards a local optimum of  the objective. These updates are 
\begin{align}
	q_{X}(\vect{X})^{new} \propto & \: p(\vect{x}_{1})  \prodlow{t,j}\psi_{t+1,j}(x_{t+1}^{j})  \exp \Big\{ \sumlow{t} \BE{\log p(\vect{x}_{t+1}\mid \vect{x}_{t}, \vect{A}, \vect{Q})}_{q_{AZ}, q_{Q}} \Big\}, 
	\label{EqnX}
	\\
	q_{AZ}(\vect{A}, \vect{Z})^{new} \propto & \:  \prodlow{i\sim j} p(a_{ij}\mid z_{ij})p(z_{ij})  \exp \Big\{ \sumlow{t} \BE{\log p(\vect{x}_{t+1}\mid \vect{x}_{t}, \vect{A}, \vect{Q})}_{q_{X}, q_{Q}} \Big\}, 
	\label{EqnVarAZ}
	\\
	q_{Q}(\vect{Q})^{new} \propto  & \: \prodlow{i} p(q_{ii} \mid \vect{\theta})  \exp \Big\{  \sumlow{t}\BE{\log p(\vect{x}_{t+1}\mid \vect{x}_{t}, \vect{A}, \vect{Q})}_{q_{X}, q_{AZ}} \Big\},
	\label{EqnVarQ}
\end{align}
where we use $\BE{\cdot}_q$ to denote the expectation with respect to a distribution $q$. These updates are performed in a circular fashion to achieve a coordinate descent in each step. If the first and second moments of $q_{X}$ are known, then \eqref{EqnVarAZ} and \eqref{EqnVarQ}, and hence also the exponential term in the update \eqref{EqnX}, can be easily found. However, non-Gaussian components of \eqref{EqnX} prevent us from directly computing the moments of $q_X$.

To deal with the analytical intractability of $q_{X}$, we propose to compute the required expectations by applying further approximate inference techniques. We view $q_{X}$ from a graphical model \citep{Lauritzen1996} or factor graph \citep{Kshischang2001} perspective and propose a novel large-scale extension of an approximate inference technique called expectation propagation \citep{OpperWinther2000, Minka2001}.  We show that the approximations we arrive at can be embedded into the wider, principled framework of variational approximate inference by using expectation constraints  \citep{Heskes2005, OpperWinther2005}. We extend  this framework to accommodate our structured variational inference approach for the joint model $p( \vect{X}, \vect{A}, \vect{Z}, \vect{Q} \mid \mathcal{Y}, \vect{\theta})$. In the following sections we present the algorithmic approach. A  detailed, generic derivation of the proposed method is given in the Supplementary Material.

\subsection{The model $q_{X}$ } \label{SecInfX}

The model for $q_{X}$ is a dynamic latent Gaussian model where the non-Gaussian terms $\psi_{t,j}(x_{t}^{j})$ depend on only one state-space component $x_{t}^{j}$. We collect the Gaussian terms into $\Psi_{t, t+1}(\vect{x}_{t}, \vect{x}_{t+1}) \propto \exp \{ \BE{\log p(\vect{x}_{t+1}\mid \vect{x}_{t}, \vect{A}, \vect{Q})}_{q_{AZ}, q_{Q}} \}\times \exp(\vect{h}_{t+1}^T\vect{x}_{t+1})$ and define $q_X$ as 
\begin{align}\label{EqnQX}
	q_{X}(\vect{X}) \propto \prodlow{t} \Psi_{t, t+1}(\vect{x}_{t}, \vect{x}_{t+1})  \times \prodlow{t,j} \psi_{t+1,j}(x_{t+1}^{j}). 
\end{align}
The model is tree structured, and thus inference can be done by using the message-passing algorithm \citep[e.g.,][]{Lauritzen1996} or the sum-product algorithm \citep[e.g.,][]{Kshischang2001}. 

Message passing algorithms operate on so-called factor graphs. These are bipartite graphs for which the node sets consist of factors of a probability distribution and the corresponding variables (or groups thereof). A factor is connected to all variables (or groups thereof) that are subsets of its arguments. Message passing is a dynamic programming algorithm that computes marginal probability distributions in factor graphs with tree structure, that is, that do not contain any loops. To each edge of this graph we assign a pair of messages, one in each direction. The message from a  variable node to a factor node is computed as the product of all incoming messages from the other factors. The message from a factor node to a variable node is the marginal of the product of the factor and all other incoming messages to that factor. The dynamic nature of the algorithm guarantees that once all messages are computed, the marginals over the variables (sometimes termed beliefs) can be formed as the product of the factors and the incoming messages \citep[e.g.,][]{Kshischang2001}.

\begin{figure}[t]
	\begin{center}
		\begin{tabular}{c}
			\resizebox{0.8\textwidth}{!}{\includegraphics{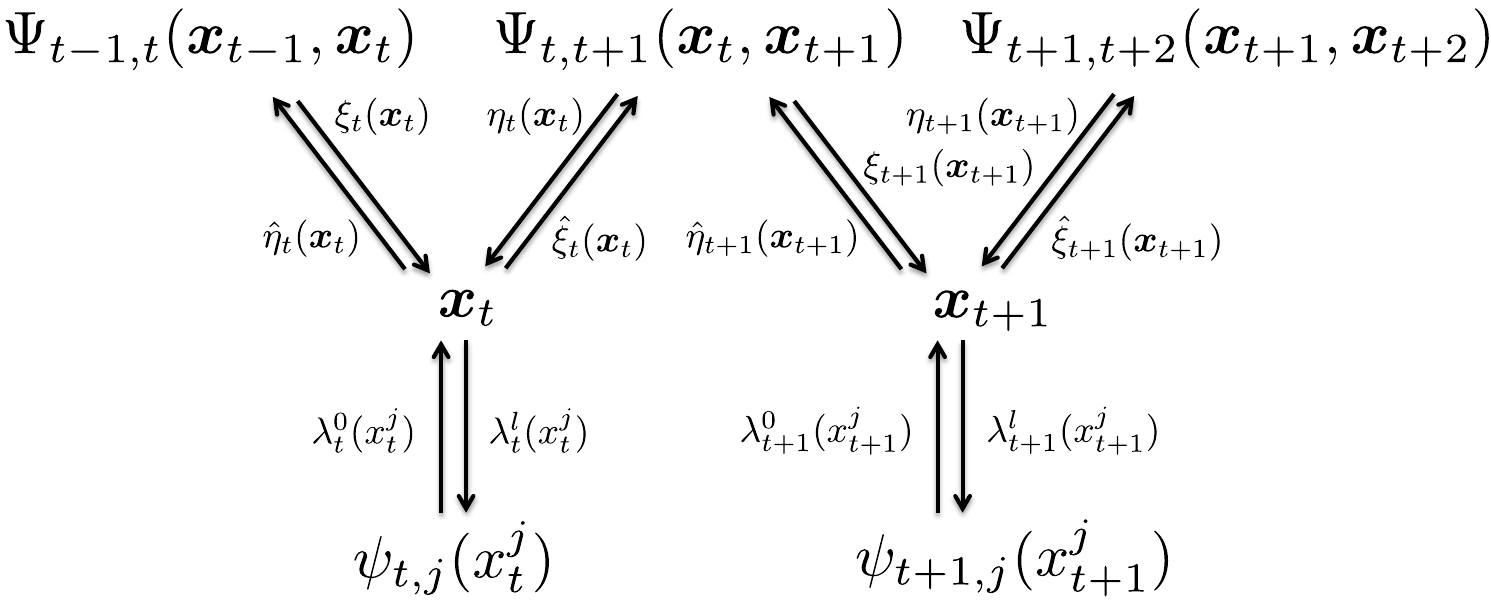}}
		\end{tabular}
	\end{center}
\caption{\small Illustration of the factor graph and the message-passing algorithm for the model $q_{X}$.}
\label{FigQX1}
\end{figure}

In our case, the factors are $\{\Psi_{t,t+1}\}_t$ and $\{\prod_j \psi_{t,j}\}_t$, while the groups of variables are the states $\{\vect{x}_t\}_t$. The message passing updates corresponding to the factor graph representation  of this model (illustrated in Figure~\ref{FigQX1}), read
\begin{align*}
	\lambda_{t+1, j}^{0}(x_{t+1}^{j}) \propto & \: \psi_{t+1,j}(x_{t+1}^{j}), 
	\\
	\xi_{t+1}(\vect{x}_{t+1}) \propto & \int\!d\vect{x}_{t}\:  \Psi_{t, t+1}(\vect{x}_{t}, \vect{x}_{t+1}) \hat{\xi}_{t}(\vect{x}_{t}), 
	\\	
	\eta_{t}(\vect{x}_{t}) \propto & \int\!d\vect{x}_{t+1}\:  \Psi_{t, t+1}(\vect{x}_{t}, \vect{x}_{t+1}) \hat{\eta}_{t+1}(\vect{x}_{t+1}),
	\\
	\lambda_{t+1, j}^{l}(x_{t+1}^{j}) \propto & \: \int\!d\vect{x}_{t+1}^{\backslash j} \: \xi_{t+1}(\vect{x}_{t+1}) \eta_{t+1}(\vect{x}_{t+1})  \prodlow{k \neq j } \lambda_{t+1, k}^{0}(x_{t+1}^{k}),
	\\
	\hat{\xi}_{t}(\vect{x}_{t})  \propto & \:	\xi_{t}(\vect{x}_{t})  \prodlow{j} \lambda_{t, j}^{0}(x_{t}^{j}),
	\\
	\hat{\eta}_{t+1}(\vect{x}_{t+1})  \propto & \: \eta_{t+1}(\vect{x}_{t+1}) \prodlow{j} \lambda_{t+1, j}^{0}(x_{t+1}^{j}).
\end{align*}
The messages $\xi_{t+1}, \eta_{t}$ and  $\lambda_{t+1, j}^{0}$ are factor-to-variable messages sent from $\Psi_{t, t+1}$ and  $\psi_{t+1,j}$ to $\vect{x}_{t+1}$ and $\vect{x}_{t}$, while 
$\hat{\xi}_{t}, \hat{\eta}_{t+1}$ and $\lambda_{t+1, j}^{l}$ are the corresponding variable-to-factor messages.

By denoting $\alpha_{t} = \hat{{\xi}}_{t}$ and $\beta_{t} = \eta_{t}$ and writing the marginal densities corresponding to the factors as
\begin{align}
	q_{X}(\vect{x}_{t}, \vect{x}_{t+1}) \propto &\: \Psi_{t, t+1}(\vect{x}_{t}, \vect{x}_{t+1}) \alpha_{t}(\vect{x}_{t}) \beta_{t+1}(\vect{x}_{t+1})  \prodlow{j} \lambda_{t+1, j}^{0}(x_{t+1}^{j}), 
	\label{EqnTwoSlice}
	\\
	q_{X}(x_{t+1}^{j}) \propto &  \: \psi_{t+1,j}(x_{t+1}^{j})\lambda_{t+1, j}^{l}(x_{t+1}^{j}),
	\label{EqnNGMarg}
\end{align}
we can rewrite the algorithm in the following form:\footnote{See Supplementary Material for details.}
\begin{align}
	\lambda_{t+1, j}^{0}(x_{t+1}^{j})^{new}  \: \lambda_{t+1, j}^{l}(x_{t+1}^{j})\propto & \: q_{X}(x_{t+1}^{j}), \label{EqnDWfree}
	\\
	 \lambda_{t+1, j}^{0}(x_{t+1}^{j}) \: \lambda_{t+1, j}^{l}(x_{t+1}^{j})^{new}\propto & \int\!d\vect{x}_{t}d\vect{x}_{t+1}^{\backslash j}\:  q_{X}(\vect{x}_{t}, \vect{x}_{t+1})  \label{EqnUPfree},
	\\
	\alpha_{t+1}(\vect{x}_{t+1})^{new} \: \beta_{t+1}(\vect{x}_{t+1}) \propto & \int\!d\vect{x}_{t}\: q_{X}(\vect{x}_{t}, \vect{x}_{t+1}), \label{EqnFWfree}
	\\
	 \alpha_{t}(\vect{x}_{t}) \: \beta_{t}(\vect{x}_{t})^{new} \propto & \int\!d\vect{x}_{t+1}\: q_{X}(\vect{x}_{t}, \vect{x}_{t+1}). \label{EqnBWfree}
\end{align}
In the typical approach, one performs  forward-backward updates w.r.t $\alpha_t$ and $\beta_t$ whilst also doing a $\lambda_{t, j}^{l}$ and $\lambda_{t,j}^{0}$ update at each time step. This algorithm is analogous to the well known Rauch--Tung--Striebel smoothing algorithm for linear dynamical systems with a Gaussian likelihood.

In order to make the message passing tractable with a non-Guassian likelihood, we use Gaussian messages between the Gaussian and non-Gaussian terms.  Moreover, we propose to use messages with restricted precision structures between the Gaussian factors to exploit sparsity and significantly reduce computing time; details  are presented in the following sub-sections and in the Supplementary Material.

 \subsubsection{Approximations for non-Gaussian likelihoods}\label{SecLK}
 
Due to the non-Gaussianity of $\psi_{t+1,j}$, the updates \eqref{EqnDWfree}--\eqref{EqnBWfree} cannot be computed analytically. To deal with this, we recast  $\lambda_{t+1, j}^{0}(x_{t+1}^{j})$ as a Gaussian message by defining  $\tilde{q}_{t+1}(x_{t+1}^j) \propto  \psi_{t+1,j}(x_{t+1}^{j})\lambda_{t+1, j}^{l}(x_{t+1}^{j})$, and, based on \eqref{EqnDWfree}, introducing the approximation
\begin{align}\label{EqnEP4}
	\lambda_{t+1, j}^{0}(x_{t+1}^{j})^{new}\lambda_{t+1, j}^{l}(x_{t+1}^{j}) = \text{Project}\big[\tilde{q}_{t+1}(x_{t+1}^j);\mathcal{N}\big].
\end{align}
The operation $\text{Project}[q(\vect{x}); \, \mathcal{N}]$ is the projection of a distribution $q(\vect{x})$ into the Gaussian family $\mathcal{N}$ in the moment matching Kullback-Leibler sense, that is,
\begin{align*}
	\text{Project}\big[q(\vect{x}); \mathcal{N}\big] = \mathop{\text{argmin}}\limits_{\tilde q \in \mathcal{N}}\D{}{q(\vect{x})}{\tilde q(\vect{x})}.
\end{align*}
This projection finds a Gaussian distribution $\tilde{q}(\vect{x})$ that has the first and second moments identical to those of $q(\vect{x})$. Note that the method operates by projecting the marginal distributions resulting from the message passing and not the messages themselves \citep{HeskesZoeter2002, Minka2005}.

A natural consequence of the projection \eqref{EqnEP4} is that the messages no longer compute the marginals $q_X(\vect{x}_t,\vect{x}_{t+1})$ and $q_X(x_{t+1}^j)$ given in \eqref{EqnTwoSlice} and \eqref{EqnNGMarg}, but instead local approximate marginals, which we denote as $\tilde q_{t,t+1}(\vect{x}_t,\vect{x}_{t+1})$ and $\tilde q_{t+1}(x_{t+1}^j)$, respectively.  A further consequence of the approximation is that the resulting marginals only satisfy the weak consistency conditions $\text{Project}[ \tilde{q}_{t+1}(x_{t+1}^{j});\mathcal{N}] = \tilde{q}_{t,t+1}(x_{t+1}^j)$, that is, $\tilde q_{t+1}(x_{t+1}^j)$ and $\tilde q_{t,t+1}(x_{t+1}^j)$ only match in their first two moments. This becomes apparent when the approach is derived from an expectation constrained inference perspective, see Supplementary Material for more details.

The resulting algorithm that caters for the non-Gaussian components of the likelihood can be written as follows:
\begin{align}
	\lambda_{t+1, j}^{0}(x_{t+1}^{j})^{new}  \: \lambda_{t+1, j}^{l}(x_{t+1}^{j})\propto & \: \text{Project}\big[\tilde{q}_{t+1}(x_{t+1}^j);\mathcal{N}\big], \label{EqnDWfree2}
	\\
	 \lambda_{t+1, j}^{0}(x_{t+1}^{j}) \: \lambda_{t+1, j}^{l}(x_{t+1}^{j})^{new}\propto & \int\!d\vect{x}_{t}d\vect{x}_{t+1}^{\backslash j}\:  \tilde q_{t,t+1}(\vect{x}_{t}, \vect{x}_{t+1})  \label{EqnUPfree2},
	\\
	\alpha_{t+1}(\vect{x}_{t+1})^{new} \: \beta_{t+1}(\vect{x}_{t+1}) \propto & \int\!d\vect{x}_{t}\: \tilde q_{t,t+1}(\vect{x}_{t}, \vect{x}_{t+1}), \label{EqnFWfree2}
	\\
	 \alpha_{t}(\vect{x}_{t}) \: \beta_{t}(\vect{x}_{t})^{new} \propto & \int\!d\vect{x}_{t+1}\: \tilde q_{t,t+1}(\vect{x}_{t}, \vect{x}_{t+1}), \label{EqnBWfree2}
\end{align}
where, similar to \eqref{EqnTwoSlice} and \eqref{EqnNGMarg}, 
\begin{align}
	\tilde q_{t,t+1}(\vect{x}_{t}, \vect{x}_{t+1}) \propto &\: \Psi_{t, t+1}(\vect{x}_{t}, \vect{x}_{t+1}) \alpha_{t}(\vect{x}_{t}) \beta_{t+1}(\vect{x}_{t+1})  \prodlow{j} \lambda_{t+1, j}^{0}(x_{t+1}^{j}), 
	\label{EqnTwoSlice2}
	\\
	\tilde q_{t}(x_{t+1}^{j}) \propto &  \: \psi_{t+1,j}(x_{t+1}^{j})\lambda_{t+1, j}^{l}(x_{t+1}^{j}).
	\label{EqnNGMarg2}
\end{align}
Note how, since the newly defined $\lambda_{t+1, j}^{0}(x_{t+1}^{j})$ is Gaussian, the computations for the forward and backward approximate messages in \eqref{EqnFWfree2} and \eqref{EqnBWfree2} are now tractable. As such, from an implementation point of view, the only modification to the algorithm of \eqref{EqnDWfree}--\eqref{EqnBWfree} is the replacement of \eqref{EqnDWfree} with \eqref{EqnEP4}. The $\text{Project}[\tilde q_{t+1}(x_{t+1}^{j}); \mathcal{N}]$ step in \eqref{EqnEP4}  can be done by univariate numerical quadrature. Due to the accuracy of these univariate methods, the numerical error in computing the moments is negligible. The \eqref{EqnDWfree2} and \eqref{EqnUPfree2}  updates are performed for all $j$ at once; this corresponds to the so-called parallel EP scheduling in \cite{Gerven2009} and \cite{CsekeHeskes2010b}.

Although the resulting message-passing algorithm is not exact, similar approximate message-passing algorithms have been successfully used in various models  \citep[e.g.,][]{MurphyWeissJordan1999, Minka2001,HeskesZoeter2002}.  These have been derived from different perspectives for various tasks such as  the cavity method in statistical physics \citep{OpperWinther2000},  assumed density filtering-based factor-graph inference \citep{Minka2001, Minka2005} or expectation constrained approximate inference \citep{HeskesZoeter2002, Heskes2005, OpperWinther2005}. In latent Gaussian models with log-concave likelihoods (such as the model considered here), the fixed point iteration over the messages typically exhibits fast convergence and provides good quality approximations \citep[e.g.,][]{Minka2001,KussRasmussen2005, Seeger2008}.  


\subsubsection{Exploiting sparsity}\label{SecSparsity}

The approximate messages introduced above make inference tractable since all messages ($\alpha_t$,  $\beta_{t}, \lambda_{t+1, j}^{0},\lambda_{t+1, j}^{l}$) and the relevant approximate marginals $\tilde{q}_{t,t+1}(\vect{x}_t, \vect{x}_{t+1})$ become Gaussian. However, in the case of large state spaces, the $O(n^3)$ computational and $O(n^2)$ storage costs resulting from the computation of the temporal messages $\alpha_t$ and $\beta_{t}$ can become prohibitive. To lessen these costs, and thus render inference scalable, we propose further approximations made possible by exploiting the structural sparsity of $\vect{A}$ and $\vect{Q}$. 

From \eqref{EqnTwoSlice2} we can see that, by restricting the precision matrix of the temporal messages $\alpha_t$ and $\beta_{t+1}$ to be sparse, one can keep $\tilde q_{t,t+1}(\vect{x}_{t}, \vect{x}_{t+1})$'s precision matrix sparse. This allows us to use fast sparse linear algebraic methods, such as the sparse Cholesky factorisation and partial matrix inversions, to compute the required moments. To introduce our proposed approximations which result in temporal messages with sparse precision structures, we proceed as follows.


Let $\vect{f}(\vect{x}) = (x_1, \ldots, x_{n}, \{-x_{i}x_{j}/2\}_{i\sim j})$ denote the sufficient statistic of a Gaussian Markov random field where $i\! \sim \! j$ follows the connectivity of  a graph with $n$ vertices, $G(\vect{f})$, to be specified later. Let $\text{Project}[q(\vect{x});\mathcal{N}_{\vect{f}}]$ denote the Kullback-Leibler moment matching projection to the Gaussian family with precision structure defined by $G(\vect{f})$. We make use of the form of  \eqref{EqnFWfree2}  and  \eqref{EqnBWfree2} to define the approximate message updates
\begin{align}
	\alpha_{t+1}(\vect{x}_{t+1})^{new} \beta_{t+1}(\vect{x}_{t+1})\propto & \: \text{Project}\big [\int\!d\vect{x}_{t}\:\tilde{q}_{t,t+1}(\vect{x}_{t}, \vect{x}_{t+1});\mathcal{N}_{\vect{f}} \big],
	\label{EqnAlpha} 
	\\
	\beta_{t}(\vect{x}_{t})^{new} \alpha_{t}(\vect{x}_{t})  \propto & \: \text{Project}\big [\int\!d\vect{x}_{t+1}\:\tilde{q}_{t,t+1}(\vect{x}_{t}, \vect{x}_{t+1});\mathcal{N}_{\vect{f}} \big].
	\label{EqnBeta} 
\end{align} 
These projections ensure that the forward and backward messages $\alpha_{t}$ and $\beta_{t}$ have a sparse precision structure, defined by $G(\vect{f})$, at all times. Similarly to the approach presented in the previous section, this new approximate message-passing algorithm (i.e., equations \eqref{EqnDWfree2}, \eqref{EqnUPfree2},\eqref{EqnAlpha} and \eqref{EqnBeta}) is computed iteratively until a fixed point is reached.  As in Section \ref{SecLK}, once convergence is achieved, the above definition of the messages results in the weak consistency conditions 
\begin{align*}
	\text{Project}\big [\int\!d\vect{x}_{t-1}\:\tilde{q}_{t-1,t}(\vect{x}_{t-1}, \vect{x}_{t});\mathcal{N}_{\vect{f}} \big] = \text{Project}\big [\int\!d\vect{x}_{t+1}\:\tilde{q}_{t,t+1}(\vect{x}_{t}, \vect{x}_{t+1});\mathcal{N}_{\vect{f}} \big].
\end{align*}
See Supplementary Material for details.

In the following we detail the computational issues related to the newly introduced approximate message-passing algorithm. Specifically, we show that (i) fast linear algebraic methods can be applied to exploit sparsity and (ii) when and under which conditions on $G(\vect{f})$ we can carry out $\text{Project}\big [\cdot;\mathcal{N}_{\vect{f}} \big]$ efficiently.


\vspace{0.1in}

\noindent {\em A. Efficient methods for moment computations.}


In Section \ref{SecLK} we have already shown that we carry out \eqref{EqnDWfree2} by computing the first and second moments of $\tilde{q}_{t,j}$ using univariate numerical quadrature. The update \eqref{EqnUPfree2} is performed by computing the univariate canonical parameters corresponding to the marginal $\tilde{q}_{t,t+1}(x_{t+1}^{j})$, while updates in \eqref{EqnAlpha} and  \eqref{EqnBeta} are performed by computing the canonical parameters resulting from the projections to Gaussians with restricted precision structure. In the following we show how  the computation of these canonical parameters can  be carried out using efficient moment computations through sparse matrix inversion and log-determinant optimisation.

The computation of the marginal  $\int\!d\vect{x}_{t}d\vect{x}_{t+1}^{\backslash j}\:\tilde{q}_{t,t+1}(\vect{x}_{t}, \vect{x}_{t+1})$ in \eqref{EqnUPfree2}, for all $j$, reduces to the computation of the marginal means and variances of $\vect{x}_{t+1}$ in 
$\tilde{q}_{t,t+1}(\vect{x}_{t}, \vect{x}_{t+1})$,  which can be computationally expensive. The crucial idea that leads to significant computational savings is that we carry out the computations on the joint $\tilde{q}_{t,t+1}(\vect{x}_{t}, \vect{x}_{+1})$ that now has a sparse precision structure.
Let $(\vect{h}_{{\alpha}_t}, \vect{Q}_{{\alpha}_t})$ and $(\vect{h}_{{\beta}_{t+1}}, \vect{Q}_{{\beta}_{t+1}})$ denote the canonical parameters of the messages $\alpha_{t}$ and ${\beta}_{t+1}$, respectively. Further, concatenate the parameters of  ${\lambda}^{0}_{t+1,j}$ into the representation 
$(\vect{h}_{{\lambda}^{0}_{t+1,\cdot}}, \vect{Q}_{{\lambda}^{0}_{t+1,\cdot}})$ where, $\vect{Q}_{{\lambda}^{0}_{t+1,\cdot}}$ is diagonal. Recall the definition of $\Psi_{t,t+1}$ from \eqref{EqnQX} and that of $\tilde{q}_{t,t+1}$ in \eqref{EqnTwoSlice2}. The linear parameter $\vect{h}_{t,t+1}$ and the precision matrix $\vect{Q}_{t,t+1}$ of $\tilde{q}_{t,t+1}(\vect{x}_{t}, \vect{x}_{t+1})$  can be written as $\vect{h}^{T}_{t,t+1} = [\vect{h}_{{\alpha}_t}^T,  \vect{h}^T_{t+1} + \vect{h}^T_{{\beta}_{t+1}}  + \vect{h}^T_{{\lambda}^{0}_{t+1,\cdot}}]$ and
\begin{equation}\label{EqnTwoPrec}
	\vect{Q}_{t,t+1} =
		\left[ 
			\begin{array}{cc}
				\BE{\vect{A}^{T}\vect{Q}\vect{A}}_{q_{AZ},q_{Q}} + \vect{Q}_{{\alpha}_t}& -\BE{\vect{A}}_{q_{AZ}}^{T}\BE{\vect{Q}}_{q_{Q}}
				\\
				- \BE{\vect{Q}}_{q_{Q}}\BE{\vect{A}}_{q_{AZ}}            & \BE{\vect{Q}}_{q_{Q}} + \vect{Q}_{{\beta}_{t+1}} + \vect{Q}_{{\lambda}^{0}_{t+1,\cdot}}
			\end{array}
		\right]. 
\end{equation}

To compute the required moments of $\vect{x}_{t+1}$, we (i) solve the system $[\vect{Q}_{t, t+1}]^{-1}[\vect{h}_{t,t+1}]$ and (ii) compute the diagonal of $[\vect{Q}_{t, t+1}]^{-1}$. We do this by a sparse Cholesky factorisation of a fill-in reduction reordering of $\vect{Q}_{t, t+1}$ \citep{Davis2006} followed by (i) solving triangular sparse linear systems and (ii) doing a partial inversion by solving the  Takahashi equations \citep{Taka1973}. The Takahashi equations compute all the covariance elements that correspond to non-zeroes in the Cholesky factor, and thus to all non-zeros in $\vect{Q}_{t,t+1}$. This property is pivotal to rendering the $\text{Project}[\,\cdot\,;\, \mathcal{N}_{\vect{f}}]$ step computationally efficient.



In the following, we show how to derive the canonical parameters for $\alpha_{t+1}$; a similar procedure is used to derive those for $\beta_t$. From \eqref{EqnAlpha}, it follows that we have to project $\tilde q_{t,t+1}(\vect{x}_{t+1})$ into the Gaussian family $\mathcal{N}_\vect{f}$. Let $\vect{m}_{t,t+1} = \vect{Q}_{t,t+1}^{-1}\vect{h}_{t,t+1}$ and $\vect{V}_{t,t+1} = \vect{Q}_{t,t+1}^{-1}$. Further, let $\vect{m}_{t,t+1}^{[t+1]}$ and  $\vect{V}_{t,t+1}^{[t+1]}$ denote the  marginal mean and variance of $\vect{x}_{t+1}$. Then, $\text{Project}[\tilde q_{t,t+1}(\vect{x}_{t+1});\mathcal{N}_{\vect{f}}]$ reduces to finding the matrix $\vect{Q}_{\alpha_{t+1}}$ which solves 
\begin{eqnarray}
	\mathop{\text{minimise}}\limits_{\vect{Q}_{\alpha_{t+1}}} && \tr\BR{\vect{V}_{t,t+1}^{[t+1]}\vect{Q}_{\alpha_{t+1}}} - \log \det \vect{Q}_{\alpha_{t+1}} \label{EqnMaxDet}
		\\
		\text{s.t.} && [\vect{Q}_{\alpha_{t+1}}]_{ij} = 0, \: \text{for all} \: (i,j) \not\in {G}(\vect{f}),  \nonumber
\end{eqnarray}
and $\vect{h}_{\alpha_{t+1}} = \vect{Q}_{\alpha_{t+1}}^{-1}\vect{m}_{t,t+1}^{[t+1]}$. 

The optimisation \eqref{EqnMaxDet} can be solved by gradient-based methods or the Newton method and  the calculations are  computationally expensive.
However, when the graph $G(\vect{f})$ is \emph{chordal} \citep[e.g.,][]{Lauritzen1996}, optimality conditions lead to equations that can be solved exactly (without expensive optimisation) by using only the values $[\vect{V}_{t,t+1}^{[t+1]}]_{ij}$ with $(i,j) \in G(\vect{f})$ \citep{Dahl2008}. Recall that the partial matrix inversion of $\vect{Q}_{t,t+1}$ always computes these values, and hence no further covariance computations are needed. For this reason, in this paper we especially consider restricting temporal  messages to have chordal precision structure (that is, we set $G(\vect{f})$ to be chordal). The algorithm for computing $\text{Project}[\cdot ;\mathcal{N}_{\vect{f}}]$ for chordal graphs follows from \cite{Dahl2008} and is presented in the Supplementary Material. It has a complexity that scales approximately cubically with the largest clique size in $G(\vect{f})$, which is generally much less than $n$.

\vspace{0.1in}

\noindent {\em B. Choosing chordal structures for spatial applications.}

\begin{figure*}[!ht]
	\begin{center}
		\begin{tabular}{cc}
			{\tiny $G(\vect{f})$} & {\tiny $ \vect{Q}_{t,t+1}\:  \text{($G(\vect{f})$ based on {\tt amd})}$}
			\\			
			\resizebox{!}{0.45\textwidth}{\includegraphics{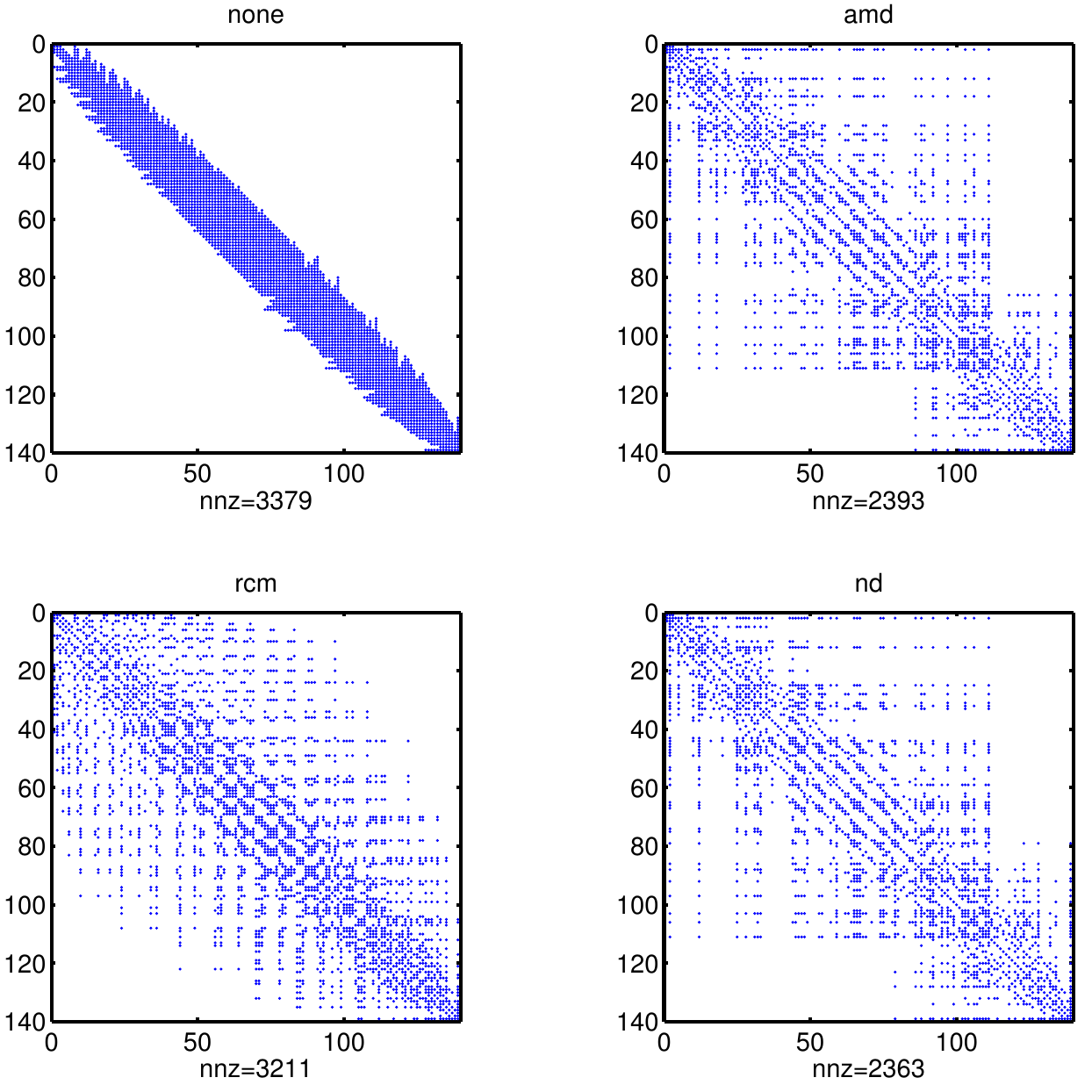}}
			&
			\resizebox{!}{0.45\textwidth}{\includegraphics{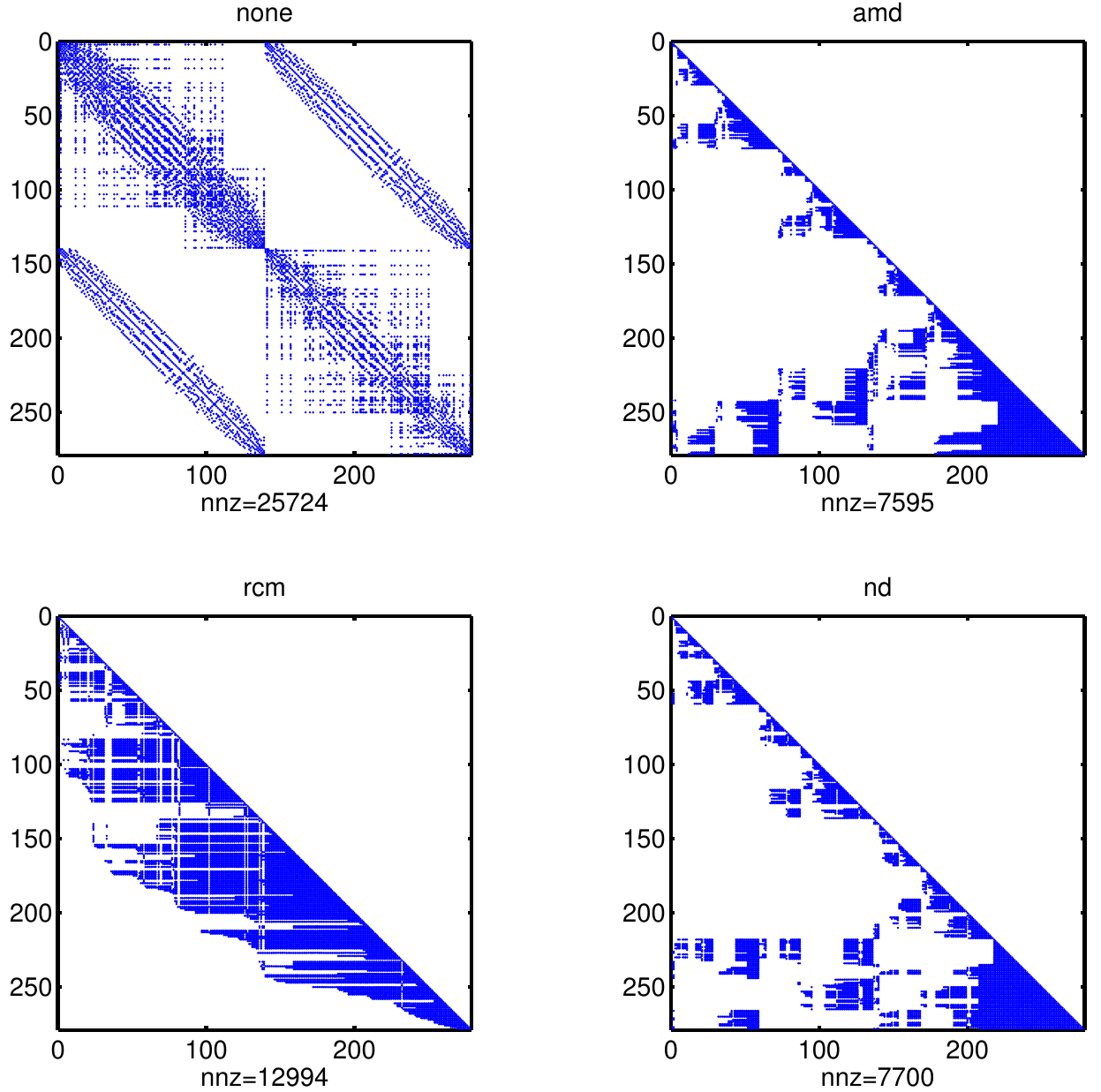}}
			\\
			(a) & (b)	
			\\
			\\
			{\tiny $\vect{Q}_{t,t+1}\: \text{ ($G(\vect{f})$ based on  {\it diag})}$} & {\tiny $\vect{Q}_{t,t+1}\: \text{($G(\vect{f})$ based on {\it tsp}})$}
			\\
			\resizebox{!}{0.45\textwidth}{\includegraphics{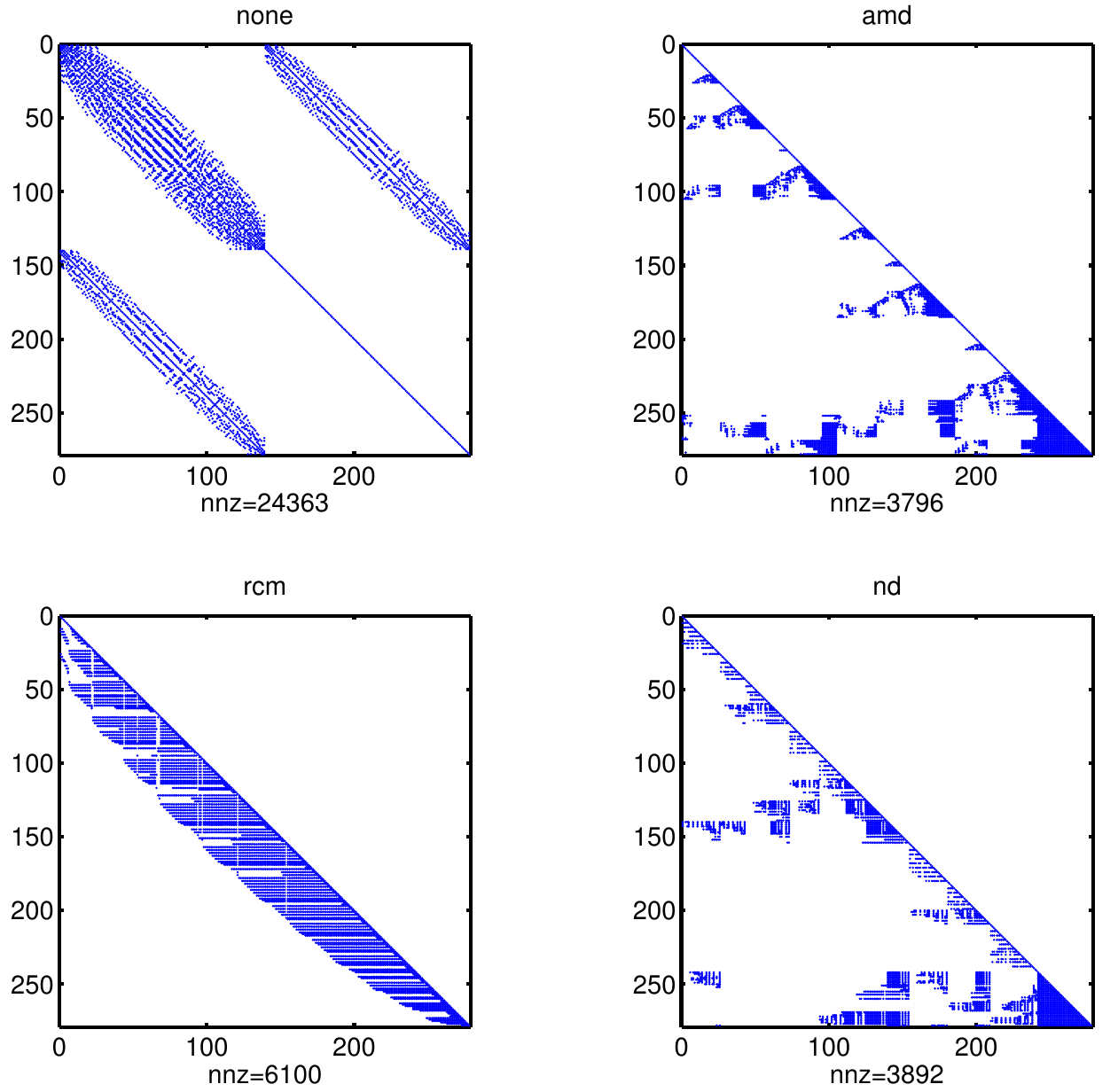}}
			&
			\resizebox{!}{0.45\textwidth}{\includegraphics{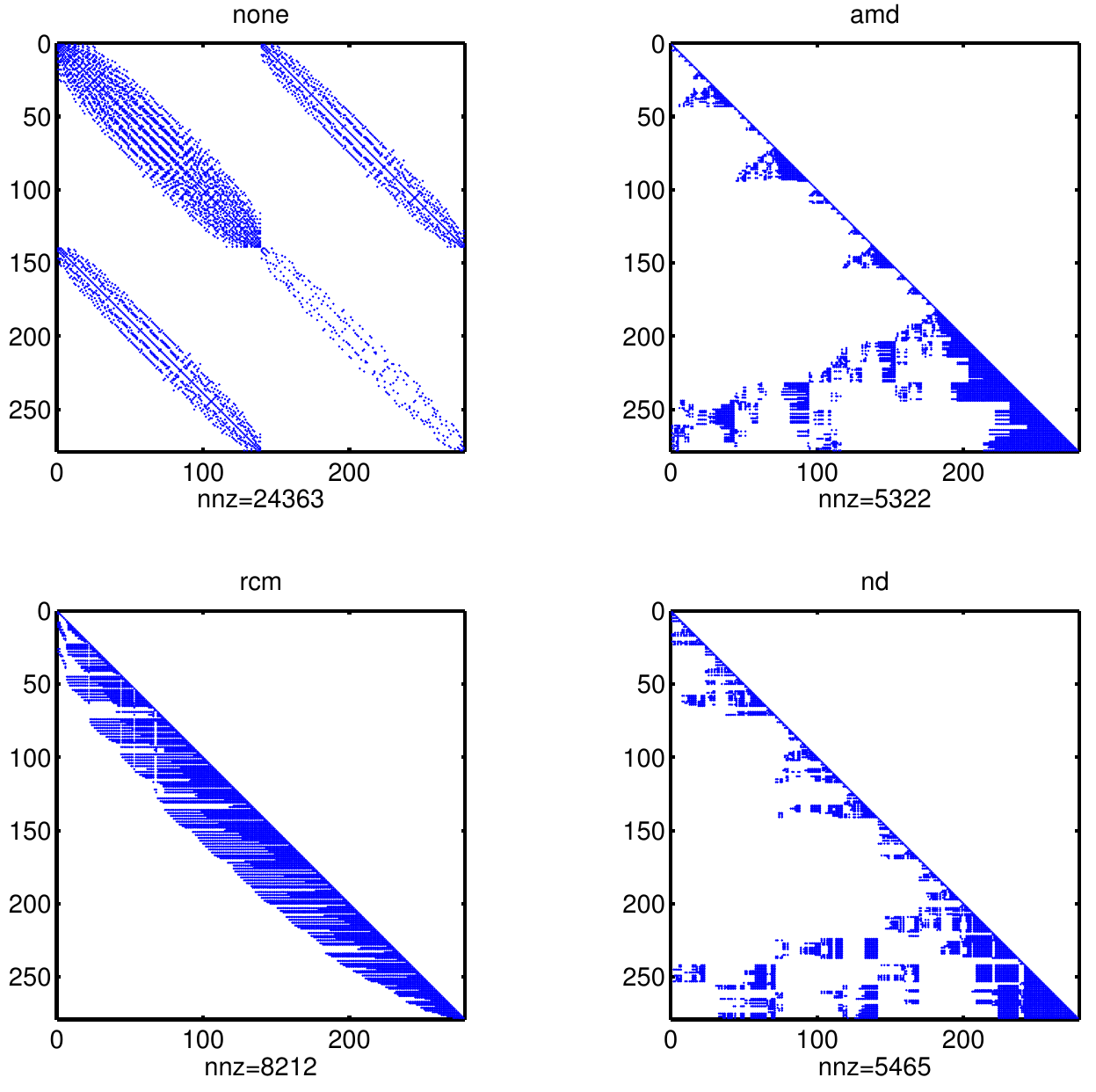}}	
			\\
			(c) & (d)
		\end{tabular}
	\end{center}
	\caption{\small Illustration of the sparsity structures of the graph $G(\vect{f})$ and the matrix $\vect{Q}_{t,t+1}$ on the lattice model illustrated in Figure~\ref{FigSTdemo}. The panels in (a) show the chordal completions of the sparsity structure of the lattice obtained by symbolic Cholesky factorisations using ``fill-in'' reducing permutations. Panels (b), (c) and (d) show the structure of $\vect{Q}_{t,t+1}$ for a choice of $G(\vect{f})$ as well as the structure of its Cholesky factors for various reordering permutations. For more details see Sections~\ref{SecSparsity} and \ref{SecScheduling}.}
	\label{FigSparseMat}
\end{figure*}	

\begin{figure}
	\begin{center}
		\begin{tabular}{cc}
			\resizebox{0.45\textwidth}{!}{\includegraphics{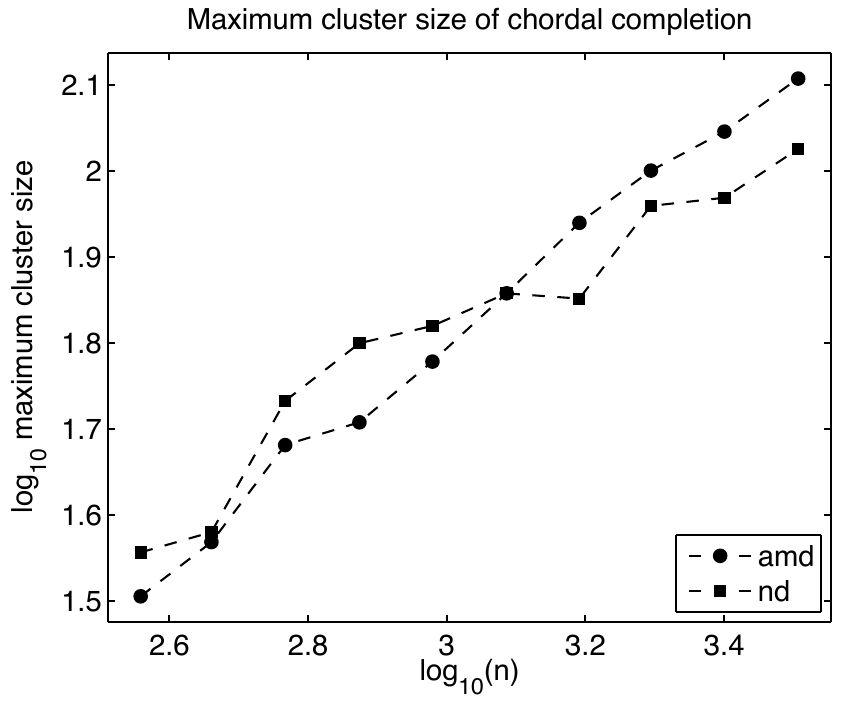}}
			&
			\resizebox{0.45\textwidth}{!}{\includegraphics{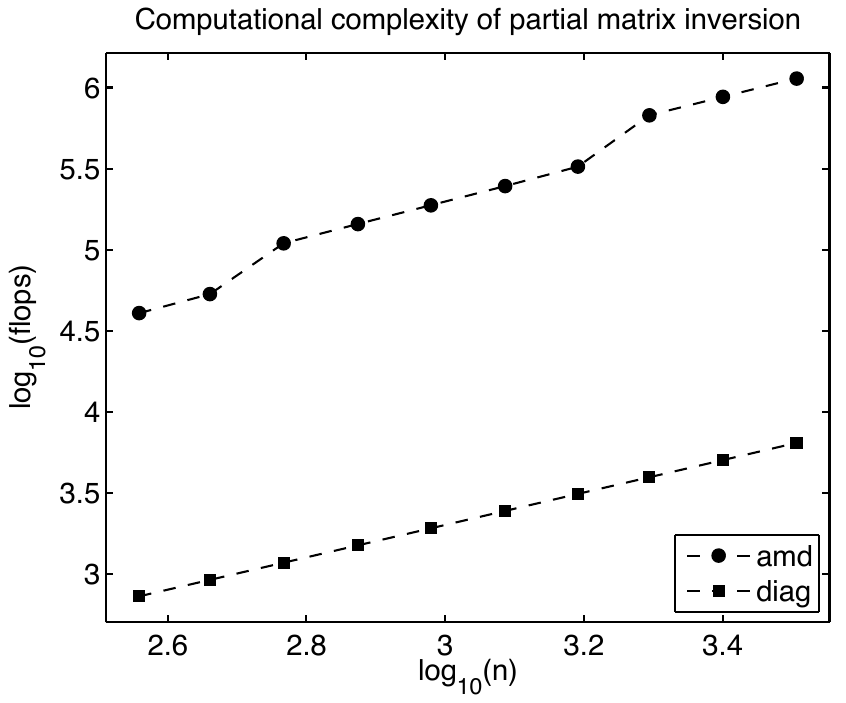}} 
		\end{tabular}
	\end{center}
\caption{\small Computational complexities of sparse matrix operations as a function of state dimension. The left panel shows the maximum clique size of the chordal graphs generated from a triangular mesh by using {\tt amd} and {\tt nd} permutations. The right panel shows the empirical computational complexity estimates for the partial matrix inversion of $\vect{Q}_{t,t+1}$.}
\label{FigCompCplx}
\end{figure}

In principle, any chordal graph structure can be used, however, since in this work we are concerned with spatial applications, it is natural to motivate the choice of $G(\vect{f})$ based on the neighbourhood graph of the spatial lattice. In general, the neighbourhood graph corresponding to a spatial lattice is \emph{not} chordal. Therefore, in order to take advantage of the efficient optimisation resulting from the use of chordal structures, we need to include extra edges in the neighbourhood graph such that the resulting $G(\vect{f})$ is chordal. 

It is well known that sparse Cholesky factorisations create chordal matrices that contain the original sparse matrix structure \citep{Davis2006}. For this reason, we propose to construct chordal graph structures by carrying out (symbolic) sparse Cholesky factorisations of the adjacency  matrix given by the spatial lattice. Our choice of chordal graphs is motivated by computational arguments: we aim to construct chordal structures that are maximally sparse, so that the precision matrix in \eqref{EqnTwoPrec} is as sparse as possible. There is a substantial literature on maximising the sparsity of the sparse Cholesky factors by row-column permutations, see \cite{Davis2006} and references therein. In this paper we  use (i) the approximate minimum degree permutation \citep{Amestoy1996}, denoted by {\tt amd}, (ii) the symmetric reverse Cuthill-McKee permutation \citep{Cuthill19698}, denoted by {\tt rcm}, and  (iii) the nested dissection permutation \citep{brainman2002nested}, denoted~by~{\tt nd}.

The panels of Figure~\ref{FigSparseMat} show the sparsity structures of $G(\vect{f})$, $\vect{Q}_{t,t+1}$ and the corresponding Cholesky factor for various choices of row-column permutations for a triangular lattice with $n = 140$. The matrix $\vect{A}$ corresponds to a lattice structure as illustrated in Figure~\ref{FigSTdemo} and $\vect{Q}$ is diagonal.  The  group of panels (a) show the chordal graphs $G(\vect{f})$ generated from the lattice using the various permutations. The group of panels (b) show the structure of the $\vect{Q}_{t,t+1}$s for which $G(\vect{f})$ is the chordal completion of the neighbourhood graph following an {\tt amd} permutation. The top-left panel of this group shows $\vect{Q}_{t,t+1}$, while the other panels in this group show its corresponding Cholesky factors following the {\tt amd}, {\tt rcm}, and {\tt nd} permutations. The panels (c) and (d) consider different graphs $G(\vect{f})$, \emph{diag} and \emph{tsp}, where \emph{diag} corresponds to a diagonal structure and \emph{tsp} corresponds to a spanning tree of the neighbourhood graph. These graph structures are discussed further in Section \ref{SecScheduling}. 

The panels of Figure~\ref{FigCompCplx} show the maximum clique size of the chordal graph generated from a triangular mesh and the empirical computational complexity estimates  for the partial matrix inversion of $\vect{Q}_{t,t+1}$. We generated triangular meshes of various size on a circular domain and used the {\tt amd} and {\tt nd} permutations to complete them to chordal graphs. The left panel shows that the maximum clique size scales approximately as $O(n^{\gamma})$ with $\gamma \sim 1/2$ for both permutations, implying that the time required to solve \eqref{EqnMaxDet} scales approximately as $O(n^{3/2})$ or $O(n^2)$ at most (recall the cubic complexity of the projection step w.r.t. maximum clique size).  The right panel shows an estimate of the number of flops required for solving the Takahashi equations given the sparse Cholesky factor of $\vect{Q}_{t,t+1}$. The \emph{chordal} component in this plot was generated by using {\tt amd} and {\tt nd} permutation-based chordal completions for $G(\vect{f})$ and an ${\tt amd}$ permutation for the sparse Cholesky factorisation on $\vect{Q}_{t,t+1}$. The flop-count estimates show that the computational complexity of the {\em chordal} methods scales approximately as $O(n^{3/2})$ or $O(n^2)$ at most, and that the computational complexity of the {\em diag} is significantly lower. These results are in line with the complexity estimates in \cite{George1973} and \cite{Rue_2005}.

\subsubsection{Approximations and message passing schedules} \label{SecScheduling}

The choice of $G(\vect{f})$, and the scheduling of the updates in the fixed point iteration of equations \eqref{EqnDWfree2}, \eqref{EqnUPfree2},\eqref{EqnAlpha} and \eqref{EqnBeta}, govern the accuracy and the computational complexity of the inference algorithm. In the following we detail our choices and relate some of these to the current literature.

The computational complexity of the approximations is dominated by the partial matrix inversion of $\vect{Q}_{t,t+1}$, that is in turn directly determined by the structure of $G(\vect{f})$. We thus consider three classes of $G(\vect{f})$: (i) {\em full}, where $G(\vect{f})$ is  fully connected, which corresponds to an approximate inference approach of propagating full Gaussian messages \citep{ypma2005novel}, (ii) {\em chordal}, where $G(\vect{f})$ is a chordal graph, corresponding to messages having precision matrices with restricted sparsity structure (the spanning tree structure \emph{tsp} is a special case), and (iii) {\em diag}, where $G(\vect{f})$ is a disconnected graph, corresponding to  factorised temporal messages. With \emph{diag}, only marginal means and variances are propagated, see \cite{MurphyWeiss2001} for an algorithmically similar approach for models with discrete variables. 

When $G(\vect{f})$ is fully connected, ({\em full} temporal messages), $\vect{Q}_{t,t+1}$ has dense diagonal blocks, and hence the computational complexity scales as $O(n^3 T)$. When $G(\vect{f})$ is chordal, this complexity is reduced; empirical studies showed that the resulting complexity is around $O(n^2 T)$. In case of \emph{tsp} and \emph{diag}, we expect a complexity of around $O(n^{3/2}T)$ \citep[][Section 2.1]{RueMartinoChopin2009}.

In terms of message scheduling we differentiate between the following choices: (i) {\em Static}. Here the forward backward updates  \eqref{EqnAlpha} and \eqref{EqnBeta} for all time steps are iterated until convergence and then the \eqref{EqnDWfree2} and \eqref{EqnUPfree2} updates are performed. In this scheduling, the forward-backward \eqref{EqnAlpha}--\eqref{EqnBeta} iteration corresponds to an approximate partial matrix inversion while the updates \eqref{EqnDWfree2} and \eqref{EqnUPfree2} correspond to the EP steps in a blocked model \citep{Minka2001,CsekeHeskes2010b}.  (ii) {\em Sequential}. Here the  \eqref{EqnDWfree2} and \eqref{EqnUPfree2} updates are  iterated until convergence at each time step, followed by a \eqref{EqnAlpha}-\eqref{EqnBeta} update; these steps are performed in a   forward backward fashion. In this scheduling, the \eqref{EqnDWfree2} and \eqref{EqnUPfree2} steps correspond to an EP algorithm approximating the Kalman update steps at each time point. (iii) {\em Dynamic}. Here, in order to minimise the number of expensive partial matrix inversions, we use a greedy scheduling strategy. With this strategy, at every step we select the message that has the largest (last) update (in terms of canonical parameters), and update both the receiver and  the source of this message, be it either $\tilde{q}_{t,t+1}(\vect{x}_t, \vect{x}_{t+1})$ or $\tilde{q}_{t+1}(x_{t+1}^{j})$.  For example, suppose that $\alpha_t$ has the largest recent update. Then, we update $\tilde q_{t,t+1}$ and its outgoing messages $\beta_t$ and $\alpha_{t+1}$ followed by an update of $\tilde q_{t-1,t}$ and its outgoing messages, thus providing a new update also for $\alpha_t$.  
Simulation studies showing the computation savings of the greedy algorithm are given in Section \ref{SecExp}, while the scheduling options are discussed further un the Supplementary Material.

By constructing longer scheduling queues (ranking the change in messages, instead of choosing the maximal change), one can distribute the computation in the dynamic scheduling scheme to several processing units and achieve a further reduction of computing time. We distribute the computation by selecting and scheduling locally independent receiver-source pair updates to different computational cores.  We adapt the greedy approach by keeping a ranking of the message updates and then in each cycle proceeding from the top of the ranking in selecting pairs of approximate marginals $\tilde q_{t,t+1}$ to update.  We select as many disjunct pairs as available computational units, proceed with the computation and repeat these cycles until convergence. Currently, we can achieve about a five fold reduction in computation time using eight cores in our Matlab\footnote{http://www.mathworks.com} implementation. We believe that this can be further improved by fine-tuning our ranking and scheduling scheme and by making better use of Matlab's distributed computing toolbox.

\subsection{The models $q_{AZ}$ and $q_{Q}$ }

In this paper we are interested in learning dynamics for models with diagonal $\vect{Q}$, therefore, we only present the inference for such models. For diagonal  $\vect{Q}$, the distribution $q_{AZ}$ factorises over the rows of $\vect{A}$ (denoted $\vect{a}_i$) and the rows of $\vect{Z}$ (denoted $\vect{z}_i$), that is, $q_{AZ}(\vect{A}, \vect{Z}) = \prod_{i} q_{AZ}^{i}(\vect{a}_{i}, \vect{z}_{i})$ (see \eqref{EqnVarAZ} in Section~\ref{SecModel}) with each $q_{AZ}^{i}$ being a multivariate conditional Gaussian. Due to the sparse lattice structure, $q_{AZ}^{i}$ simplifies to a distribution with low, say $6-10$, dimensions, and therefore the marginal moments can be computed exactly in reasonable time. The first and second moments of $\vect{a}_{i}$  needed to update $q_{X}$ and $q_Q$ are computed from ${q}_{AZ}^{i} (\vect{a}_{i})$, whereas ${q}_{AZ}^i( z_{ij})$ can be used as a measure of the relevance of $a_{ij}$.  In cases when $\vect{Q}$ is not diagonal, the model for $q_{AZ}$  has a high dimensionality and exact inference is intractable. In such cases we can resort once again to approximate message passing. The form of this non-factorising model and the corresponding message-passing algorithm  are outlined in the Supplementary Material.

In general we choose conjugate priors for $\vect{Q}$ or we keep $\vect{Q}$ fixed. When $\vect{Q}$ is diagonal, from \eqref{EqnVarQ} we can see that $q_{Q}$ factorises as $q_{Q}(\vect{Q}) = \prod_{i}q_{Q}(q_{ii})$ and that  due to conjugacy, the marginals $q_{Q}(q_{ii})$ are Gamma distributed.

\section{Experiments}\label{SecExp}


In this section we assess the speed and accuracy of the inference methods we introduced and show the potential use of this approach in the WikiLeaks Afghan War Diary data studied in~\cite{AZM2012b}. The algorithms have been coded in Matlab and for partial matrix inversion we use the implementation of \cite{fastinvc}, which is implemented in the~C programming language. 


\subsection{Accuracy of state inference in 1D models} \label{sec:Accuracy}

In this section we assess the accuracy of the state inference methods  in 1D Gaussian and Poisson models by using fixed parameters. Although, these models do not fall in the class of models we discussed so far, they are well suited for an empirical assessment of the quality of approximations we introduce.   We use the Gaussian model to assess the accuracy of restricted (temporal) message passing inference schemes {\it diag} and  {\it chordal} by comparing them to the exact {\em full}.  By replacing the Gaussian likelihood with a Poisson likelihood we assess the loss of accuracy due to non-Gaussian likelihoods. Note that the Poisson likelihood ay time $t$ and location $j$ is formally identical to $\tilde{\psi}_{t,j}$.

\subsubsection{Models and accuracy measures}

In both the Gaussian and the Poisson case, we consider a  diffusion model on a 1D grid with $n-1$ grid intervals ($n$ state space variables, $\vect{x}_t \in \mathbb{R}^{n}$) and $T$ time points. We define $\vect{A}$ as a symmetric banded matrix with various bandwidths $n_{\text{neighb}}$ and $a_{ij} = (1-\epsilon_{\vect{A}})/(1+2n_{\text{neighb}})$ with values for nodes close to the boundaries rescaled accordingly to obtain a constant row-sum $1-\epsilon_{\vect{A}}$. We define the system noise inverse covariance $\vect{Q}$ as a linear combination of a first order intrinsic field's precision matrix and  a unit diagonal matrix. We introduce a parameter $s$ to control the correlation decay in $\vect{Q}^{-1}$ and normalise $\vect{Q}^{-1}$ to obtain a chosen variance value~$v_{x}$. Formally, $\vect{Q}$ is defined as 
{\small
\begin{align*}
	 \vect{Q}(v_{x},s) =  v_{x}^{-1} \sqrt{\text{diag}(\vect{R}(s)^{-1})} \vect{R}(s) \sqrt{\text{diag}(\vect{R}(s)^{-1})},
	\quad \text{where} \quad \vect{R}(s) = \vect{I} + 10^s \vect{R}_{1},
\end{align*} 
}

\vspace{-0.75cm}
\noindent
and $\vect{R}_1$ denotes the tri-diagonal precision matrix corresponding to  the quadratic form $\sum_{i}(x_{i+1} - x_{i})^{2}$.  The left panel of Figure~\ref{FigAccGauss} shows how $s$ influences the correlation decay. 

The observation models are defined as follows. In the Gaussian case we assume that we observe the field with added Gaussian observation noise with variance $v_{obs}$, and that the field is only partially observed:  we sample the locations of the observations from the $n \times T$ space-time grid uniformly with probability $p_{\text{obs}}$. In the Poisson case the observations are Poisson random numbers with mean $\exp\{x_t^i\}$ and we sample at all locations of the space-time grid. 

As mentioned above, in the Gaussian case we focus on the accuracy of  {\it diag} and {\it chordal} and compare them to  the  exact {\it full} to assess the loss in accuracy due to the restricted temporal messages. Since the objective of the  state inference method is to approximate the two-time-slice marginals $\tilde{q}_{t,t+1}(\vect{x}_t, \vect{x}_{t+1})$, the accuracy measure we choose is the symmetric KL divergence w.r.t. the exact two time slice marginals $p_{t,t+1}(\vect{x}_t, \vect{x}_{t+1})$ computed by {\it full}.  Therefore, we define the accuracy measure
{\small
\begin{align}\label{EqnKL_acc_score}
	S(\{p_{t,t+1}\}_{t}, \{\tilde{q}_{t,t+1}\}_{t}) = \frac{1}{2 (T-1)}\sum\limits_{t=1}^{T-1}\Big\{\D{}{\tilde{q}_{t,t+1}}{\,p_{t,t+1}} + \D{}{p_{t,t+1}}{\,\tilde{q}_{t,t+1}}\Big\}.
\end{align}
}

\vspace{-0.5cm}
\noindent
In the Poisson case the quality of the approximation is affected both by the restricted temporal messages and the non-Gaussian nature of the problem. Our aim here is to assess the joint effect of both sources of inaccuracy.

In the Gaussian case the KL measure seems to be a reasonable choice to assess the distributional accuracy. However, in the Poisson case the exact marginals are not available, therefore,  we  opt for the quantile-quantile (Q-Q) summaries as a measure of accuracy. Since the accuracy measure should reflect the local nature (we only approximate marginals) of the algorithm, we use the local Gaussian approximation $\tilde{q}_{t,t+1}(\vect{x}_{t}, \vect{x}_{t+1})$ to compute the normalised residuals $\hat{\vect{\epsilon}}_{t,t+1}$ w.r.t. the  state values $\tilde{\vect{x}}_{t,t+1}$ used in the data generation. Formally, we define the residuals as 
\begin{eqnarray*}
	\hat{\vect{\epsilon}}_{t,t+1} = \vect{L}_{t,t+1}^{T}(\tilde{\vect{x}}_{t,t+1} - \vect{m}_{t,t+1}), 
\end{eqnarray*}
where $\vect{m}_{t,t+1}$ and $\vect{Q}_{t,t+1} = \vect{L}_{t,t+1}\vect{L}_{t,t+1}^{T}$ are the mean and precision corresponding to $\tilde{q}_{t,t+1}(\vect{x}_t, \vect{x}_{t+1})$. We then use the quantile values to assess how well  the standard normal distribution fits $\hat{{\epsilon}}_{t,t+1}^{j}$ for all $j$ and $t$. Specifically, we use the mean absolute deviation from the standard normal quantiles as a measure of accuracy.

\subsubsection{Simulation results}\label{SecAccUniExp}

\begin{figure}
\begin{center}
	\begin{tabular}{cc}
		\resizebox{!}{0.275\textheight}{\includegraphics{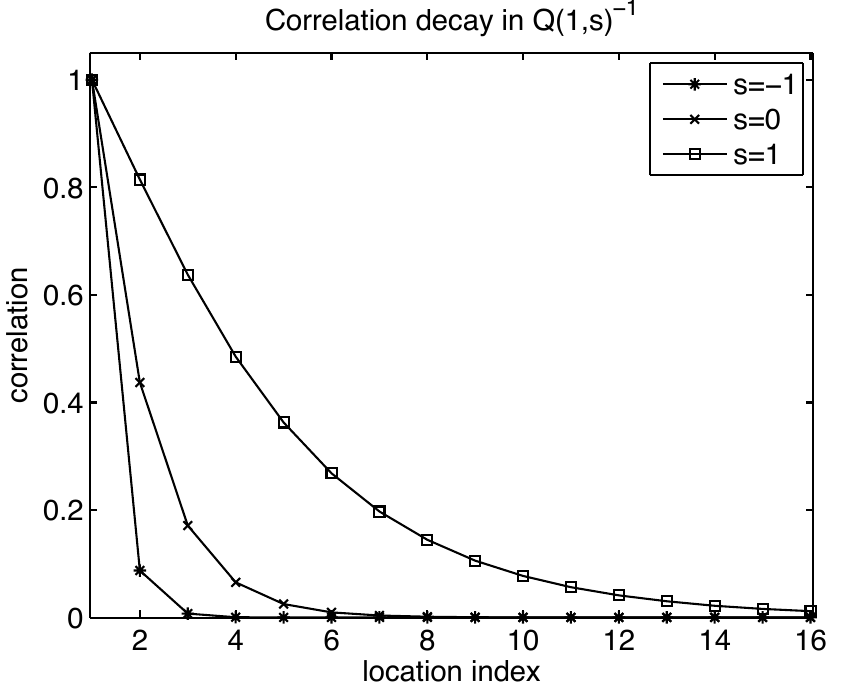}} 
		&
		\resizebox{!}{0.275\textheight}{\includegraphics{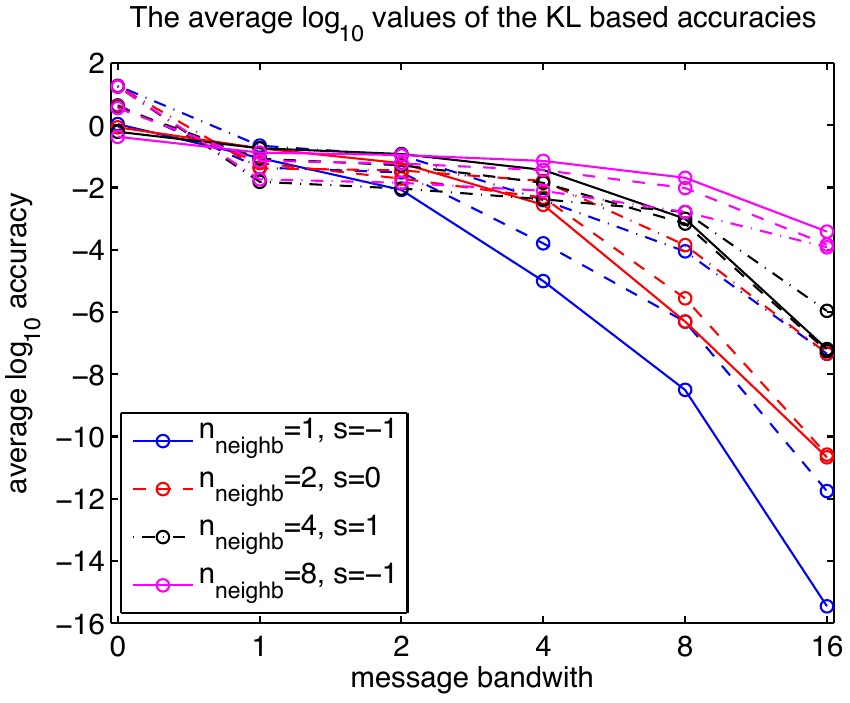}} 
	\end{tabular}
\end{center}	
\caption{\small Quantifying accuracy for the 1D  Gaussian model. The left panel shows the correlation decay in $\vect{Q}(1, s)^{-1}$  w.r.t. the location index $i$ for $s \in \{-1, 0, 1\}$ while the right panel shows the average over $n_{\text{exp}}=25$ runs of the $\log_{10}$ of KL accuracy score \eqref{EqnKL_acc_score} for various choices of $n_{\text{neighb}}$ (colour) and $s$ (line style) as a function of the message bandwidth $n_{\text{msg}}$. Lower values are indicative of better performance.}
\label{FigAccGauss}
\end{figure}

In the Gaussian case we considered models with $n = 64$, $T=100$ and  $n_{\text{neighb}} \in \{1,2,4,8\}$. We chose a system noise variance  $v_{x} = (0.5)^2$, an observation noise variance $v_{\text{obs}} = (0.25)^2$  and we set $p_{\text{obs}} = 0.75$. We chose $s \in \{-1, 0, 1\}$ leading to the correlation functions shown on the left panel of Figure~\ref{FigAccGauss}. 
We simulated the models starting from a sample from the stationary distribution. All these parameter choices together with $\epsilon_{\vect{A}} = 0.025$ lead to simulated samples with rich variations in the latent field $\{\vect{x}_{t}\}_t$ in the given time window.  We simulated $n_{\text{exp}}=25$ runs for each model and we computed the KL based score in \eqref{EqnKL_acc_score} for {\it diag} ($n_{\text{msg}}=0$) and for the {\it chordal} models corresponding to messages with bandwidths $n_{\text{msg}} \in \{1,2,4,8,16\}$ in their precision matrix, thus varying the accuracy of the approximation. Note that $n_{\text{msg}}=63$ corresponds to the exact {\it full} method and due to the univariate nature of the problem, all chordal methods ({\tt amd}, {\tt nd} and {\tt rcm}) lead to the same banded structure in the temporal messages' precision matrix. For each inference method, the inference scheme was run until the change in the maximum absolute value in the message parameters became smaller than~$10^{-8}$.  The right panel of Figure~\ref{FigAccGauss} shows the average log KL accuracies w.r.t. the message bandwidth $n_{\text{msg}}$ for various choices of $n_{\text{neighb}}$ and $s$. The accuracy plots show that for the {\it diag} method ($n_{\text{msg}}=0$) the accuracy is dominated by $s$ and that the chordal methods lead to a significant improvement in accuracy. The general pattern in the variation of the accuracy w.r.t. $n_{\text{neighb}}$ and $s$ is that, as expected, smaller correlation in $\vect{Q}^{-1}$ and fewer neighbours in $\vect{A}$ lead to better accuracy. For all cases the accuracy increases as the message bandwidth $n_{\text{msg}}$  increases thus validating the usefulness of the {\it chordal} inference methods.

\begin{figure}
\begin{center}
	\begin{tabular}{cc}
		\resizebox{!}{0.275\textheight}{\includegraphics{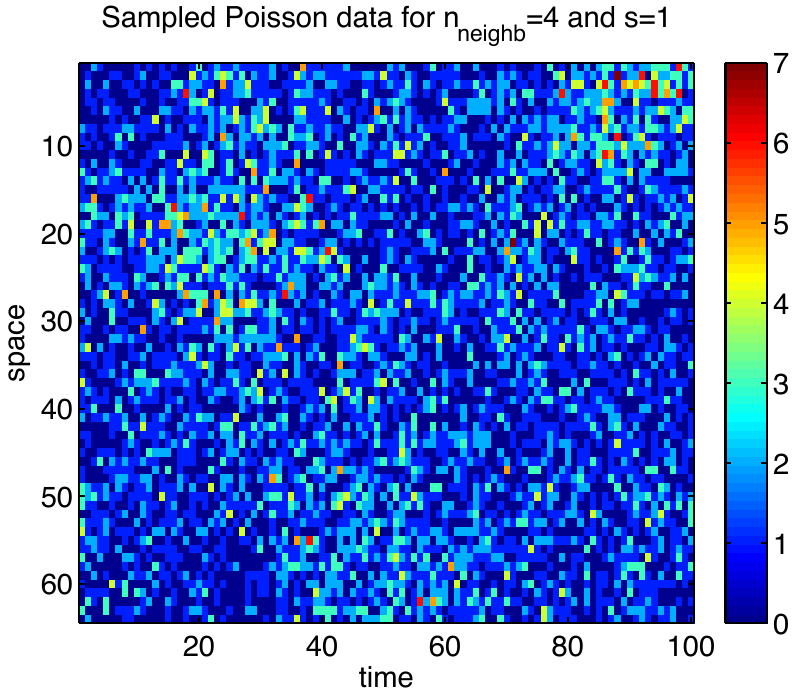}} 
		&
		\resizebox{!}{0.275\textheight}{\includegraphics{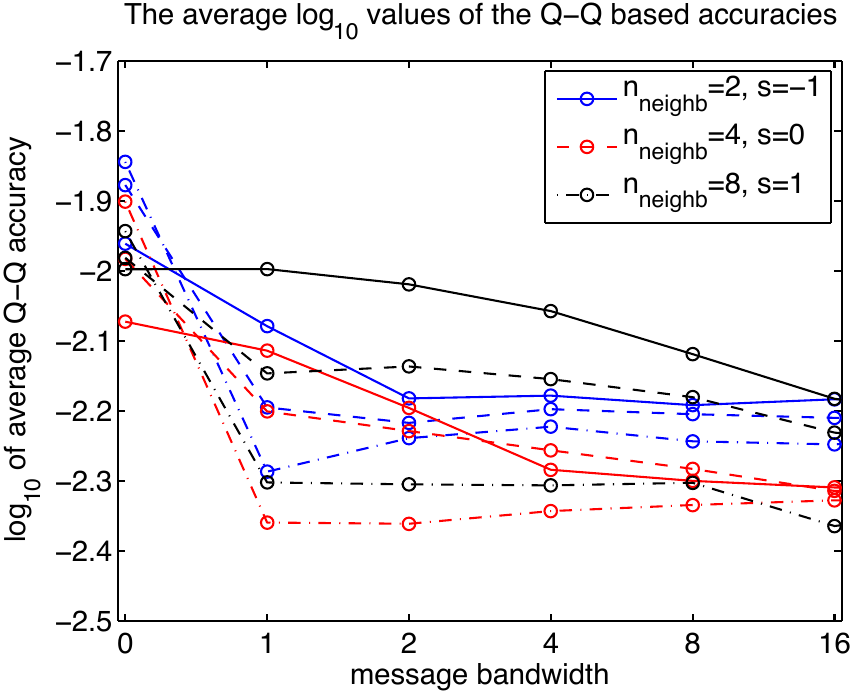}} 
	\end{tabular}
\end{center}	
\caption{\small Quantifying accuracy for the 1D Poisson model. The left panel shows sampled Poisson data from the model while the right panel shows the
logarithm of the average (over $n_{\text{exp}} = 25$ runs)  mean absolute  quantile deviations  ($50$ bins) for various choices of $n_{\text{neighb}}$ (colour) and $s$ (line style) as a function of message bandwidth $n_{\text{msg}}$. Lower values are indicative of better performance.}
\label{FigAccPoisson}
\end{figure}

In the Poisson case we used the same latent diffusion model, generated Poisson observations (see the left panel of Figure~\ref{FigAccPoisson}) and used the same inference schemes and stopping criteria as in the Gaussian case. For each run we constructed  Q-Q curves using the residuals $\hat{\epsilon}_{t,t+1}^{j}$ and 50 quantile bins and then measured the accuracy of  the inference by computing the mean absolute deviation of the curve from the diagonal.  We then averaged all the deviations over $n_{\text{exp}}=25$ experiments for each choice of $n_{\text{neighb}}$, $s$ and  message bandwidth~$n_{\text{msg}}$. The resulting accuracies are shown in the right panel of  Figure~\ref{FigAccPoisson}. The plots show that, similarly to the Gaussian case, the quality of the approximation improves as we increase the message bandwidth $n_{\text{msg}}$ and that, typically, there is a significant increase in accuracy when moving from {\it diag} to  {\it chordal} methods. As in the Gaussian case, the weaker the diffusion (smaller $n_{\text{neighb}}$) the more accurate the method is, however, it seems that in this case correlation in $\vect{Q}^{-1}$ leads to slightly improved accuracy---compare the performance of the {\it chordal} methods with the {\it diag} one and note the bad performance of {\em diag} for $s\in \{ 0,1\}$.

\interfootnotelinepenalty=10000

\subsection{Accuracy and structure recovery in a 2D spatial model}\label{SecArtData}

\begin{figure}
\begin{center}
	\begin{tabular}{cc}
		\resizebox{0.45\textwidth}{!}{\includegraphics{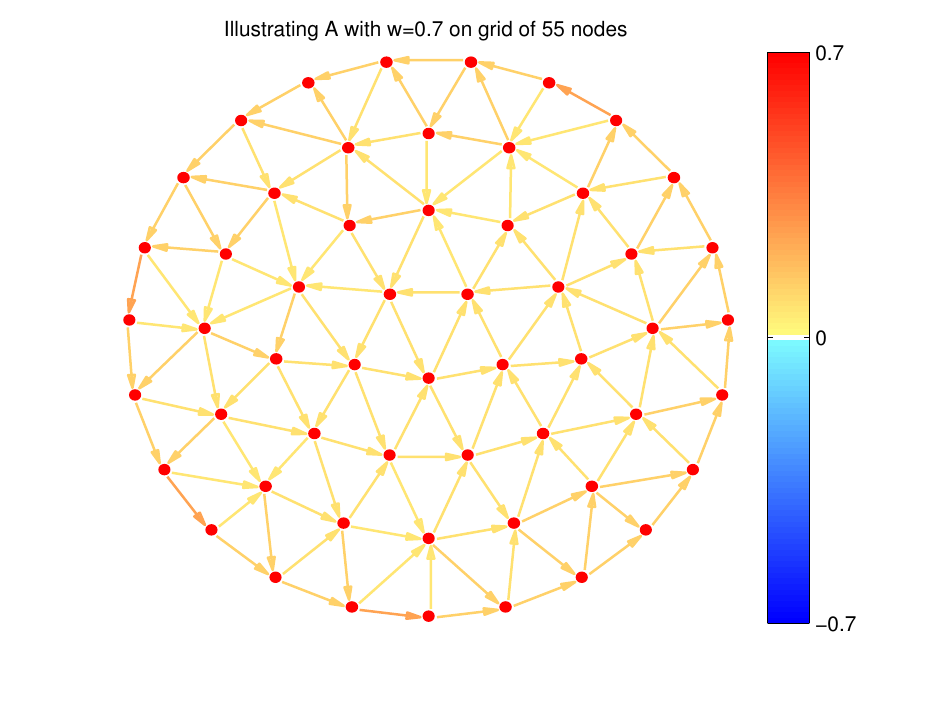}} 
		&
		\resizebox{0.45\textwidth}{!}{\includegraphics{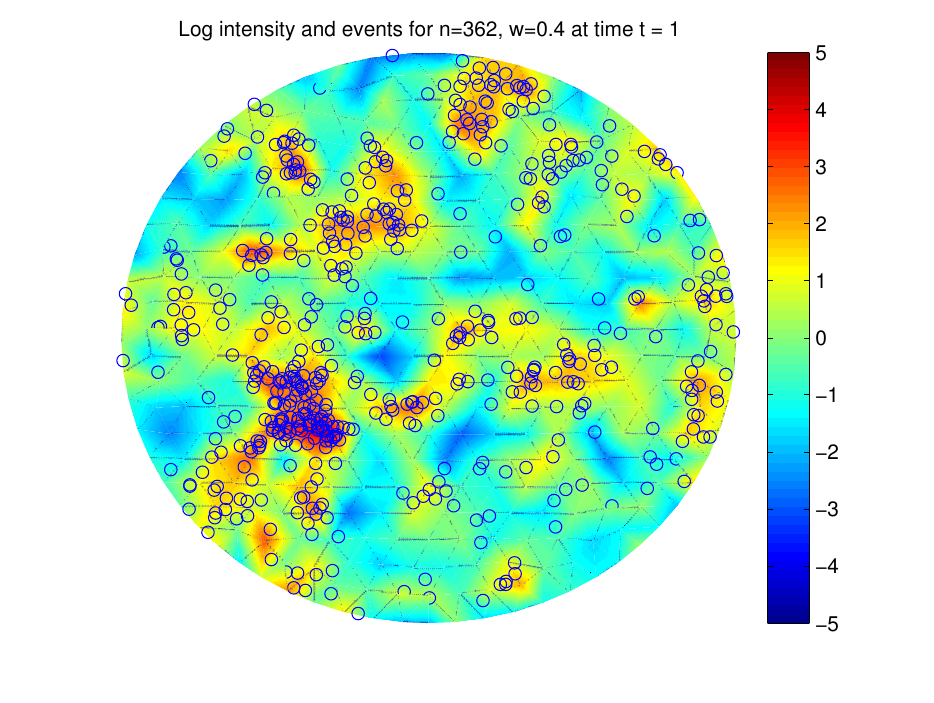}} 
		\\
		\resizebox{0.45\textwidth}{!}{\includegraphics{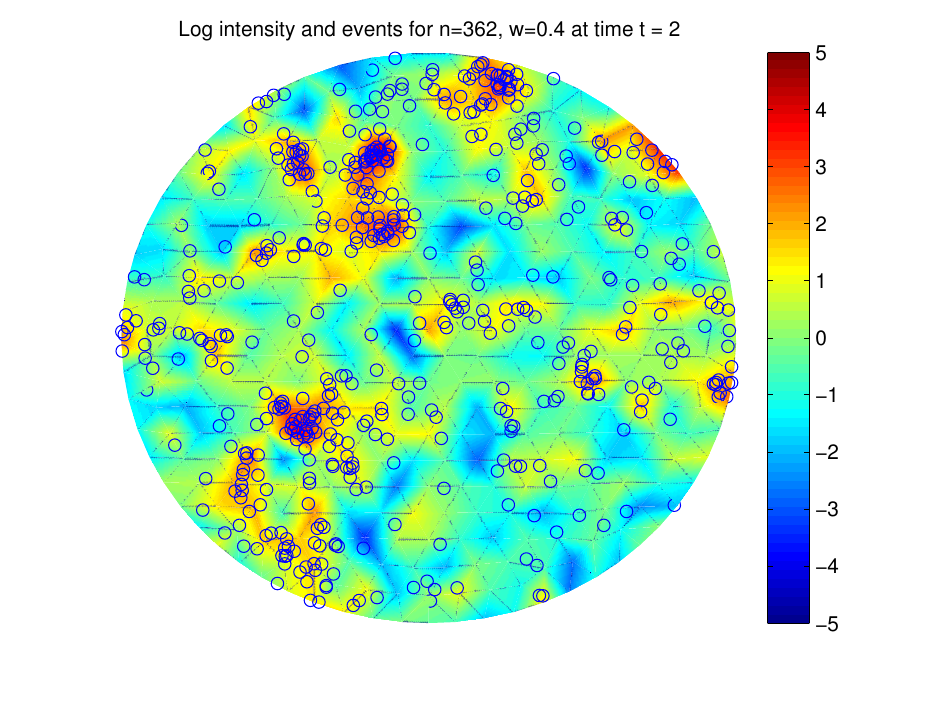}} 
		&
		\resizebox{0.45\textwidth}{!}{\includegraphics{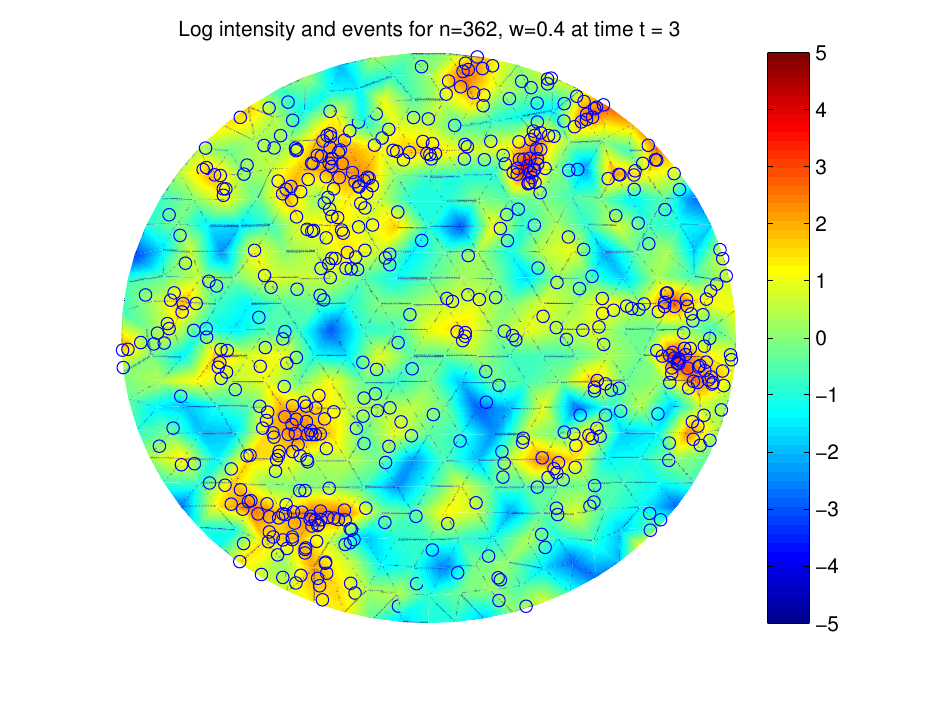}} 
	\end{tabular}
\end{center}	
\caption{\small The top left panel illustrates the transition matrix $\vect{A}$ for a 2D model with $w=0.7$ and a state space of size $n= 55$. The rest of the panels show the log intensity and the simulated events (open circles) corresponding to a sequence of 3 steps from a model with $n = 362, w = 0.4$ and $\sigma^2=1$ and a similarly structured transition matrix as in the top left panel. The resulting field, and consequently the point patterns, exhibit a rotation motion.}
\label{FigExpGrid}
\end{figure}

In this section we consider a two-dimensional spatio-temporal model where we vary the transition matrix $\vect{A}$ and a diagonal inverse covariance $\vect{Q}$ and we assess the accuracy of the state inference and the recovery rate of the network structure corresponding to $\vect{A}$. In order to obtain interesting dynamics, we define the field on a circular domain and choose the structure of the matrix $\vect{A}$ such that the model gives rise to a rotating ``motion" in the Gaussian field. To achieve this, we define $\vect{A}$ as follows: we start from a symmetric structure given by a triangular lattice, that is $a_{ij}  = a_{ji}$ for all $i,j= 1, \ldots, n$,  and for each node $i$ we eliminate all incoming edges $(i,j)$ for which the corresponding vector in the lattice is not in an anti-clockwise direction w.r.t. the domain's center, see Figure~\ref{FigExpGrid}, top-left panel. We then set the elements of the $i$-th row of $\vect{A}$ according to
\begin{eqnarray*}
a_{ii} = w, \quad \text{and} \quad a_{ij} = (1 - \epsilon_{w} -w)/\lvert \mathcal{N}(i)\rvert, 
\end{eqnarray*}
where $\mathcal{N}(i)$ denotes the set of neighbours of node $i$ in this newly defined directed structure.
We set $\epsilon_{w}$ to a small value such that the row $\vect{a}_{i}$ sums are lower than $1$, thus resulting in a zero stationary mean value for the states. We set  $\epsilon_{w} = 0.05$. The remaining panels of Figure~\ref{FigExpGrid} show samples from the field $u(s,t) = \sum_j \phi_j(s)x_{t}^{j}$ and the event data generated from it. We vary the diagonal values of $\vect{A}$ by choosing $w\in [0,1]$ and modify the system noise $\vect{Q}^{-1} = \sigma^2 \vect{I}$ by choosing $\sigma^{2} \in \{0.5, 1, 2\}$. The stationary mean and covariance of 
$\{\vect{x}_{t} \}_{t}$ are given by
\begin{eqnarray*}
	\vect{m}_{\infty} = \vect{A}\vect{m}_{\infty} \quad \text{and} \quad \vect{V}_{\infty} = \vect{A}\vect{V}_{\infty}\vect{A}^{T} + \sigma^{2}\vect{I},
\end{eqnarray*}
and the mean value of the stationary intensity is
\begin{eqnarray*}
	\BE{\lambda_{\infty}(\vect{s})} = \exp\Big\{ \vect{\phi}(\vect{s})^{T}\vect{m}_{\infty}  + \frac{1}{2} \vect{\phi}(\vect{s})^{T}\vect{V}_{\infty}\vect{\phi}(\vect{s})\Big\}.
\end{eqnarray*}
We simulated artificial event data by using initial samples from the stationarity distribution. The mean $\vect{m}_{\infty}$ is typically sufficiently close to zero to be negligible; thus the  mean intensity is determined by $\vect{V}_{\infty}$, that is,  by $w$ and $\sigma^2$. In the first experiment we assess how the accuracy of the state inference varies in terms of $w$ and $\sigma^2$, while in the second one we fix $\sigma^{2}=1$ and do joint inference for the parameters in $\vect{A}$ and the states in $\vect{X}$. 

\begin{figure}
\begin{center}
	\begin{tabular}{ccc}
		\resizebox{0.3\textwidth}{!}{\includegraphics{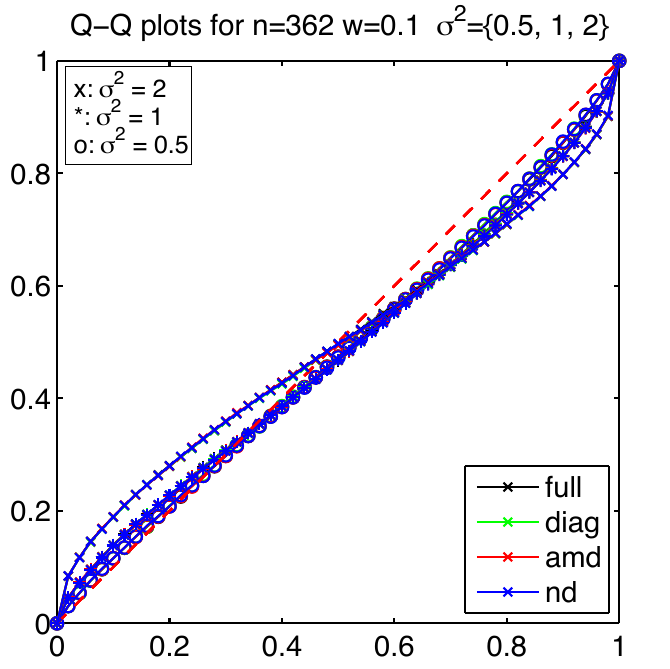}} 
		&
		\resizebox{0.3\textwidth}{!}{\includegraphics{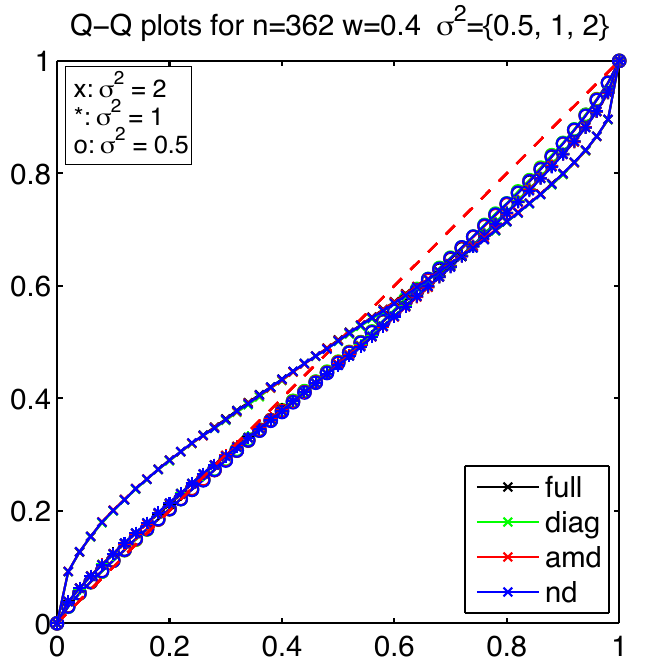}} 
		&
		\resizebox{0.3\textwidth}{!}{\includegraphics{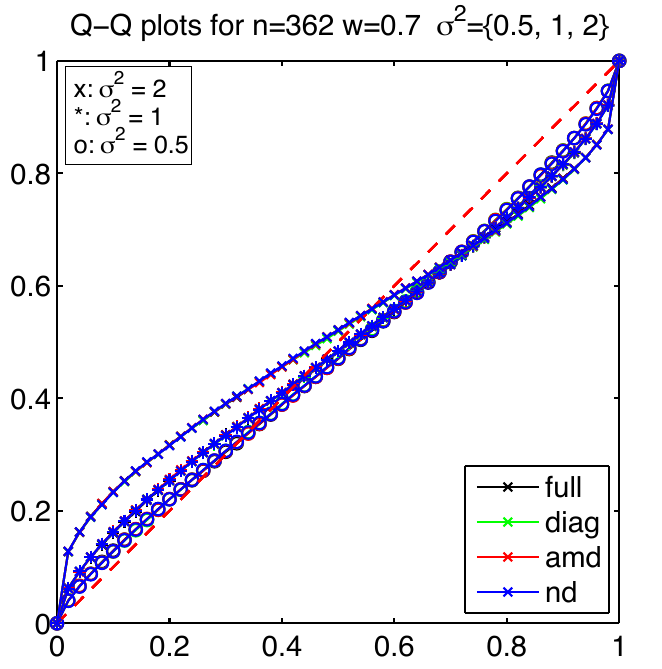}} 
		\\
		\resizebox{0.3\textwidth}{!}{\includegraphics{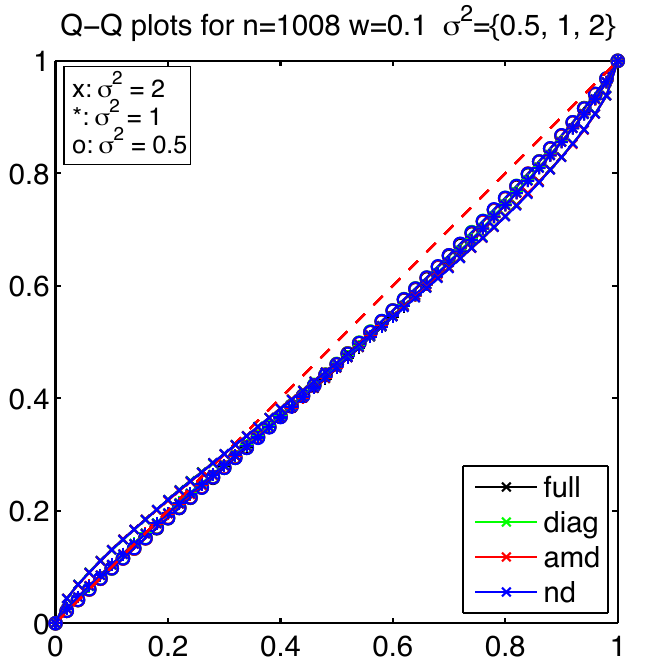}} 
		&
		\resizebox{0.3\textwidth}{!}{\includegraphics{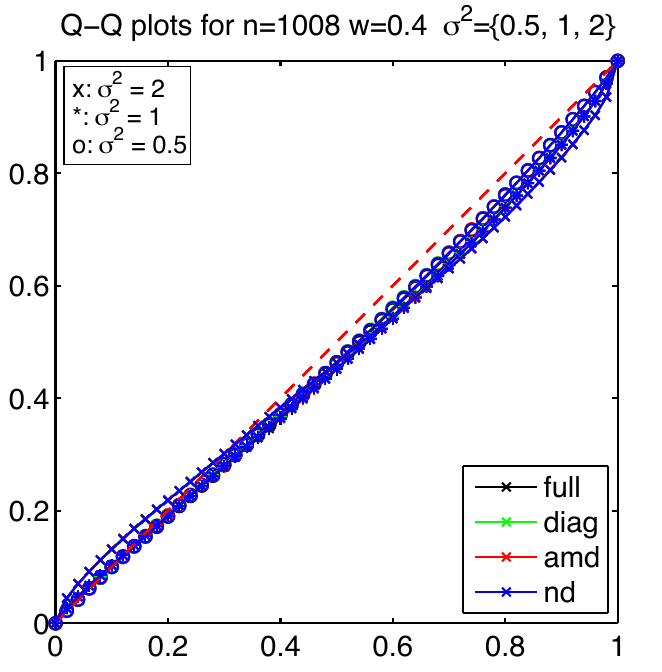}} 
		&
		\resizebox{0.3\textwidth}{!}{\includegraphics{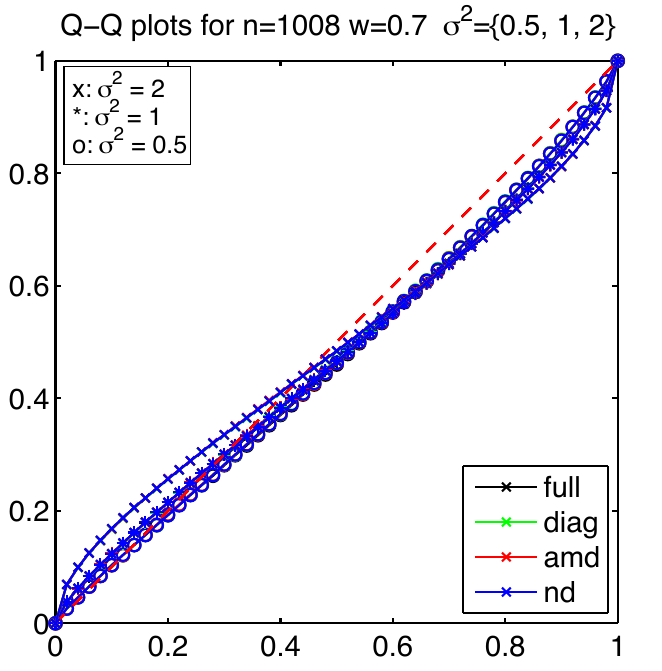}} 
	\end{tabular}
\end{center}	
\caption{\small Accuracy of the state space approximation in a 2D spatial model. The panels show the Q-Q plots for a variety of parameter settings and methods in models with $n=362$ (top) $1008$ (bottom) and $T=50$. The Q-Q plots were generated using the residuals $\hat{\epsilon}_{t,t+1}^{j}$, see Section~\ref{SecArtData}.}
\label{FigAcc}
\end{figure}

We assess the accuracy of the state inference on models with $n \in \{362,1008\}$. We generated a sequence of $T=50$ state samples $\{\tilde{\vect{x}}_{t}\}_{t}$ starting from the stationary distributions and sampled the event data by using a standard thinning method. We then ran the state inference methods using the $w$ and $\sigma^{2}$ parameters the data was generated by.  The panels in Figure~\ref{FigAcc} show the Q-Q plots using the residuals ${\hat\epsilon}_{t,t+1}^{j}$ for various settings of $w$ and $\sigma^2$. We can see that in this model the methods have very similar performance  (the Q-Q plots overlap) and there is a decrease in performance as the values of $w$ and $\sigma^2$ increase, as expected. The overlap of the Q-Q plots can be explained by the diagonal nature of $\vect{Q}$ (low noise correlation) and the magnitude of the Q-Q accuracies, see Figure~\ref{FigAccPoisson} right panel. The worsening of performance due to increasing $w$ is negligible compared to that due to increasing $\sigma^2$, which can be explained by the fact that higher system noise leads to less accurate approximations.

\begin{figure}
	\begin{center}
		\begin{tabular}{ccc}
			\resizebox{0.3\textwidth}{!}{\includegraphics{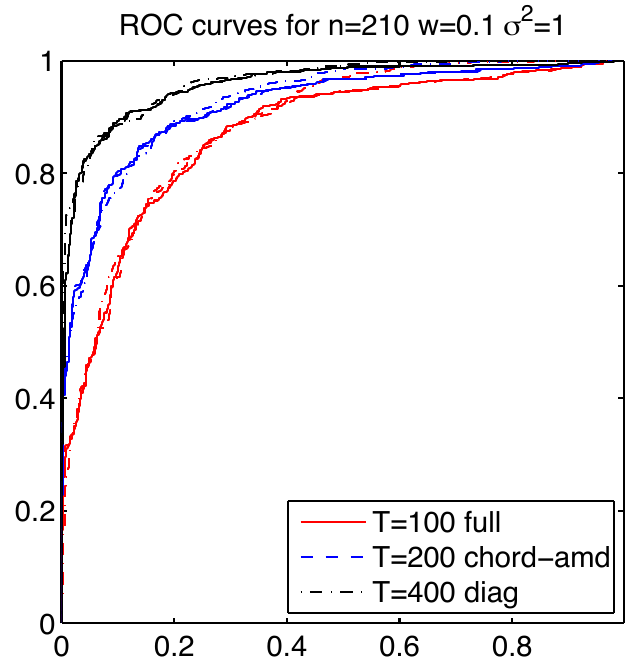}} 
			&
			\resizebox{0.3\textwidth}{!}{\includegraphics{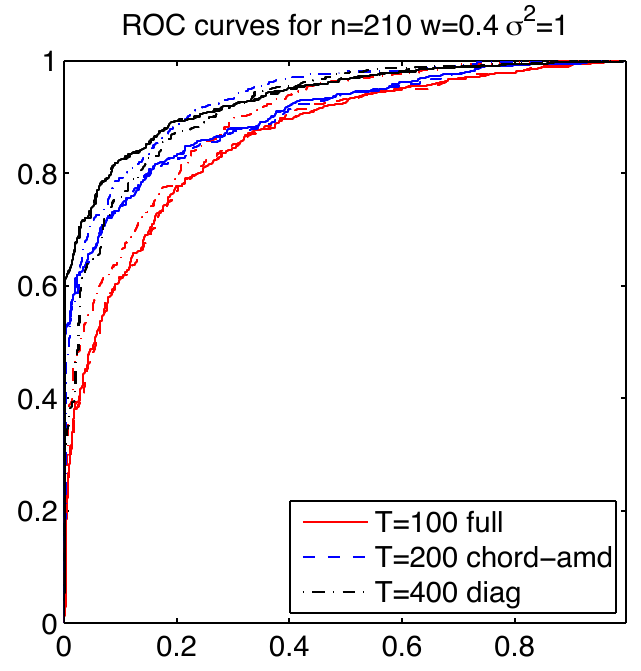}}
			&
			\resizebox{0.3\textwidth}{!}{\includegraphics{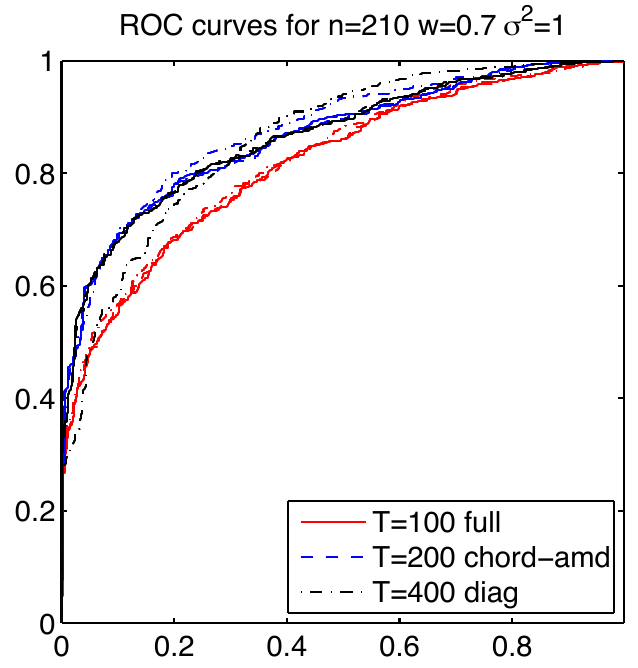}} 
		\end{tabular}
	\end{center}
\caption{\small Structure recovery. The panels of the figure show the ROCs for various settings of $w$ and $T$ and for a variety of state inference methods, see Section~\ref{SecArtData}. Colours denote the choices of $T$ while  line types denote the inference method used for~$q_{X}$.}
\label{FigRecovery}
\end{figure}

Recovering the structure of $\vect{A}$ is important in many spatio-temporal applications where we want to infer how events spread over a certain geographic area, see for example the data in Section~\ref{SecAFG}. 
To test the quality of  edge recovery, we  generated data by using the above described construction of $\vect{A}$ and a fixed $\sigma^{2}=1$, and inferred the (approximate) posterior distribution of $\vect{X}, \vect{A}$ and $\vect{Z}$. In Section~\ref{SecInf} we mentioned that the (approximate) posterior distribution of the variables $z_{ij}$ can be used to quantify the relevance of an edge $(i,j)$. Here we use $P(z_{ij}=1)$ as classification scores to assess whether the correct edges have been eliminated from the prior lattice structure, and we construct receiver-operator curves (ROC) to assess the quality of the structure inference. These curves show the true positive rate versus the false positive rate when varying the classification threshold between $0$ and $1$.  The panels in Figure~\ref{FigRecovery} show the ROC curves for various choices of $w$, $T$ and approximation methods. We can conclude that, as expected, the quality of the recovery increases as $T$ increases for all values of $w$. As expected lower values of $w$ and thus higher diffusion speeds lead to better performance on structure recovery in a limited time window. The difference in various state inference methods is hardly noticeable in these models and is well within the expected statistical variation. This lack of difference can be explained by the accuracy results discussed above. 

\subsection{Running times and scalability}



To determine the scalability of the algorithm as we vary $n$, we use the setting of Section \ref{SecArtData}  with $\vect{A}$ fixed to encode the neighbourhood graph. The results presented in this section are insensitive to the specific parameter values. We used a set of parameters that, at stationarity, result in around $1000$ events per time frame; a typical count for large datasets.

The algorithms were tested on domains with varying mesh density, $n \in \{362, 562, 1008\}$ and computing times were recorded using Matlab's profiler. The message passing was run until every parameter in the message was not changed by more than $10^{-4}$ in successive iterations. To ensure a fair comparison, all test results given here are with computations restricted to a single processor core.\footnote{All algorithms were tested on an Intel Core\texttrademark i7-2600S @ 2.80GHz personal computer with 8GB of RAM.}

\begin{figure*}[t]
	\begin{center}
		\begin{tabular}{ccc}
			\resizebox{0.51\textwidth}{!}{\includegraphics{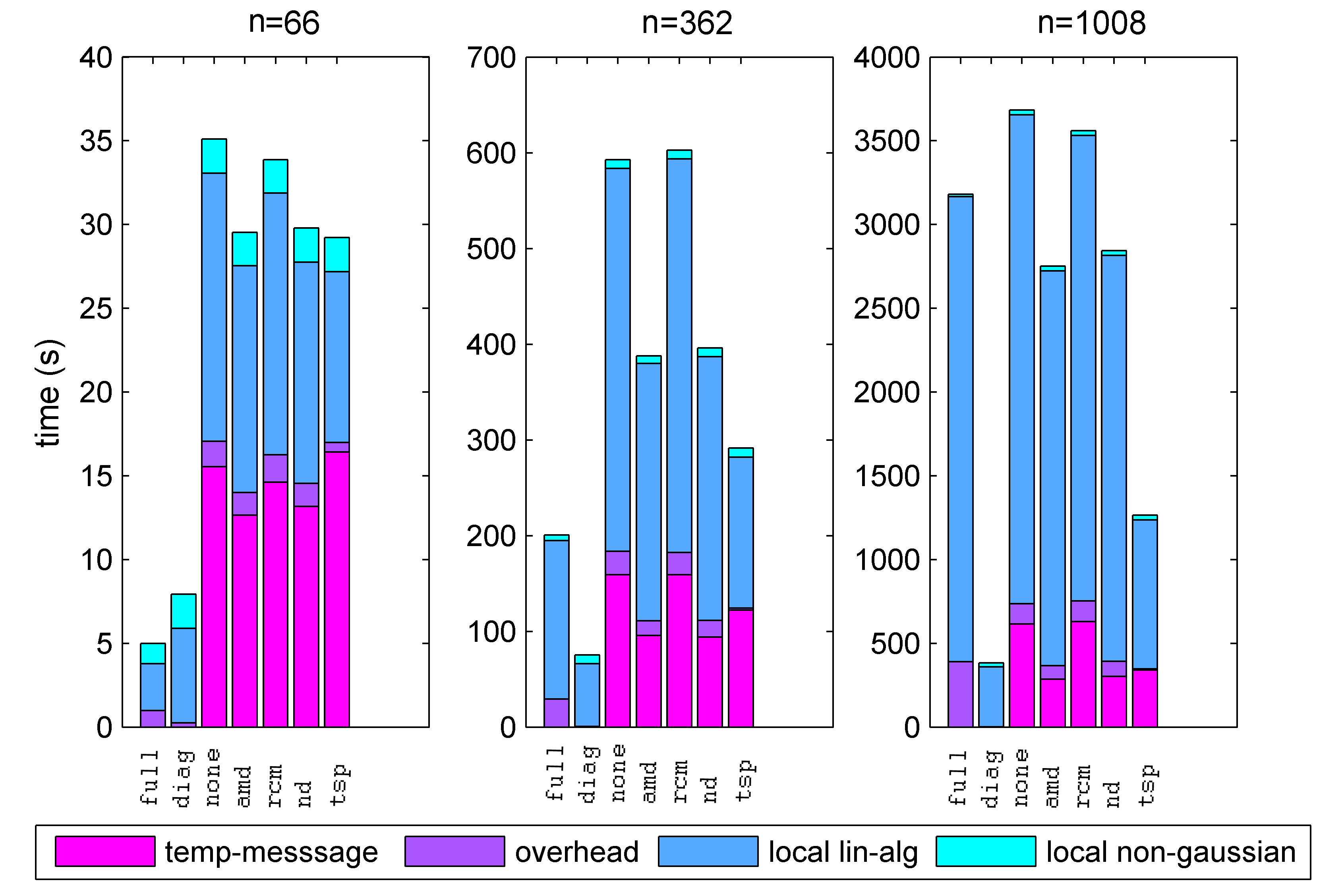}}
			\resizebox{0.44\textwidth}{!}{\includegraphics{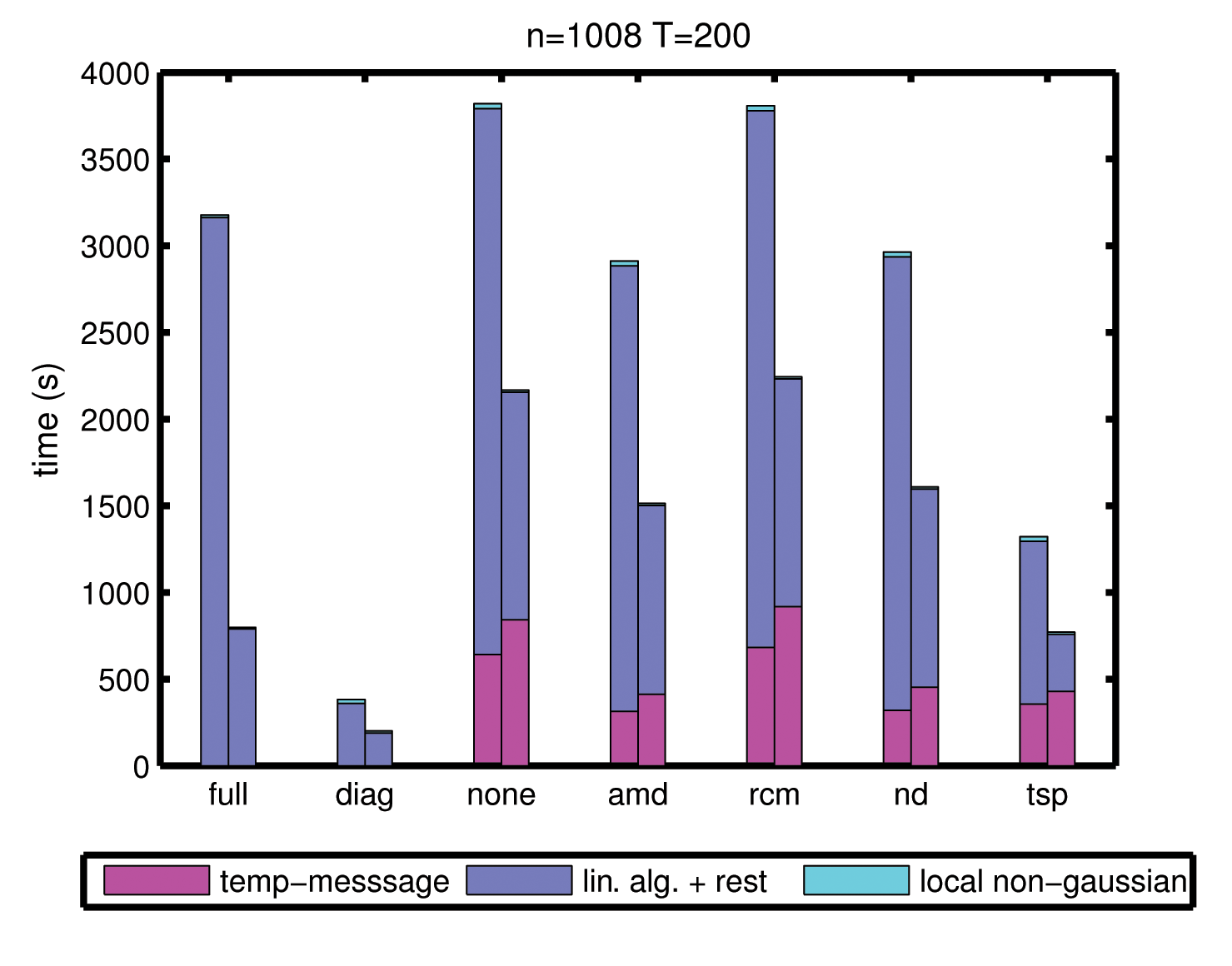}}
		\end{tabular}
	\end{center}
\caption{\small Left panel: Running times for various state space sizes and scheduling options. (left) Running times for the inference schemes {\it full}, {\it diag}, and chordal schemes {\tt none}, {\tt amd}, {\tt rcm}, {\tt nd} and {\it tsp}. Right panel: Comparison in terms of running times of the sequential (left bar) and greedy (right bar) scheduling strategies. Local operations refer to the local linear algebra whilst the temporal messages refer to the $\text{Project}[\cdot ;\mathcal{N}_{\vect{f}}]$ optimisation.}
\label{FigRT}
\end{figure*}

The computing times for the sequential scheduling scheme are plotted in the left panel of Figure~\ref{FigRT}. We segmented the computing times to correspond to the three main operations: (i) {\em temp-messages} stands for the $\text{Project}[\cdot ;\mathcal{N}_{\vect{f}}]$ operation \eqref{EqnMaxDet}, (ii) {\em overhead} accounts for initialisations, message updating and convergence monitoring, (iii) {\em local lin-alg} logs the time for linear algebraic operations (dominated by the Cholesky factorisation and partial matrix inversion), and (iv) {\em local non-Gaussian} stands for the univariate moment computations to update~${\lambda}^{0}_{t,j}$ in \eqref{EqnDWfree2}.  For clarity, we omit results for the static scheduling case which were up to an order of magnitude slower than the second worst-performing method.

The left panel in Figure~\ref{FigRT} shows that, for small $n$, the {\it full} inference scheme is faster than the other schemes due to the fact that it is implemented more efficiently in terms of dense matrix operations (Matlab/LAPACK core routines). However, the situation changes for $n = 1008$, where we see that the {\it full} is slower than the best {\it chordal} methods and much slower than the {\it tsp} and {\it diag}. Note that the increase in total computing time is well below cubic and at most quadratic for all methods other than the {\it full}. It is clear from this figure that \emph{full} will become untenable for large $n$ (note the rate of growth w.r.t. $n$).

Although the scheduling itself does not affect the scalability of the algorithm, it can be seen from Figure~\ref{FigRT}, right panel, that, as expected, the greedy scheduling can greatly reduce the computing time. For instance, after the initial forward-backward, the {\it full} needed only a few factor updates to achieve convergence within tolerance.

\subsection{The Afghan War Diary}\label{SecAFG}

Spatio-temporal point-process methods have recently been shown to be a valuable tool in the study of conflict. In \cite{AZM2012b}, a dynamic spatio-temporal model, inspired by the integro-difference equation, was used to obtain posterior estimates of conflict intensity in Afghanistan and to predict conflict levels using an iterative state-parameter update scheme on the WikiLeaks Afghan War Diary (AWD). Updates on $\vect{x}_t$ were found using an algorithm similar to the {\it full} described above. The spatial scales considered there were on the order of a 100 km and thus, modelling of micro-scale effects such as relocation or escalation diffusions in conflict were not possible. Conflict dynamics are known to occur at much smaller scales \citep{Schutte_2011}, even at resolutions of $\approx$10 km. The goal of this section is thus primarily to show that we can perform inference at such high resolutions and, in addition, estimate the dynamics on the required spatial and temporal~scales.

\subsubsection{State estimation using fixed parameters}

\begin{sidewaysfigure}
	\begin{center}
		\begin{tabular}{cc}
			\resizebox{0.48\textheight}{!}{\includegraphics{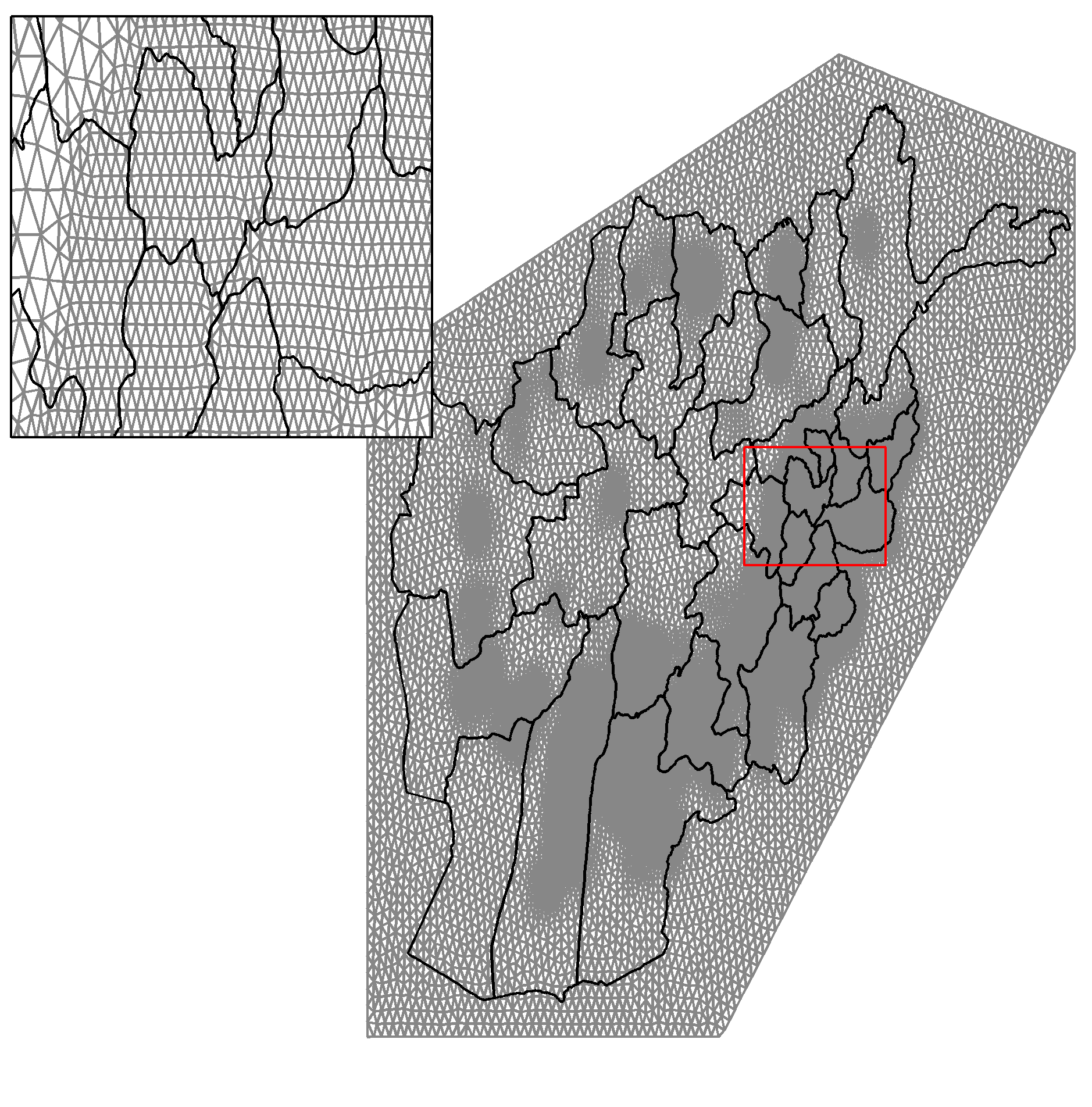}} 
			&
			\resizebox{0.48\textheight}{!}{\includegraphics{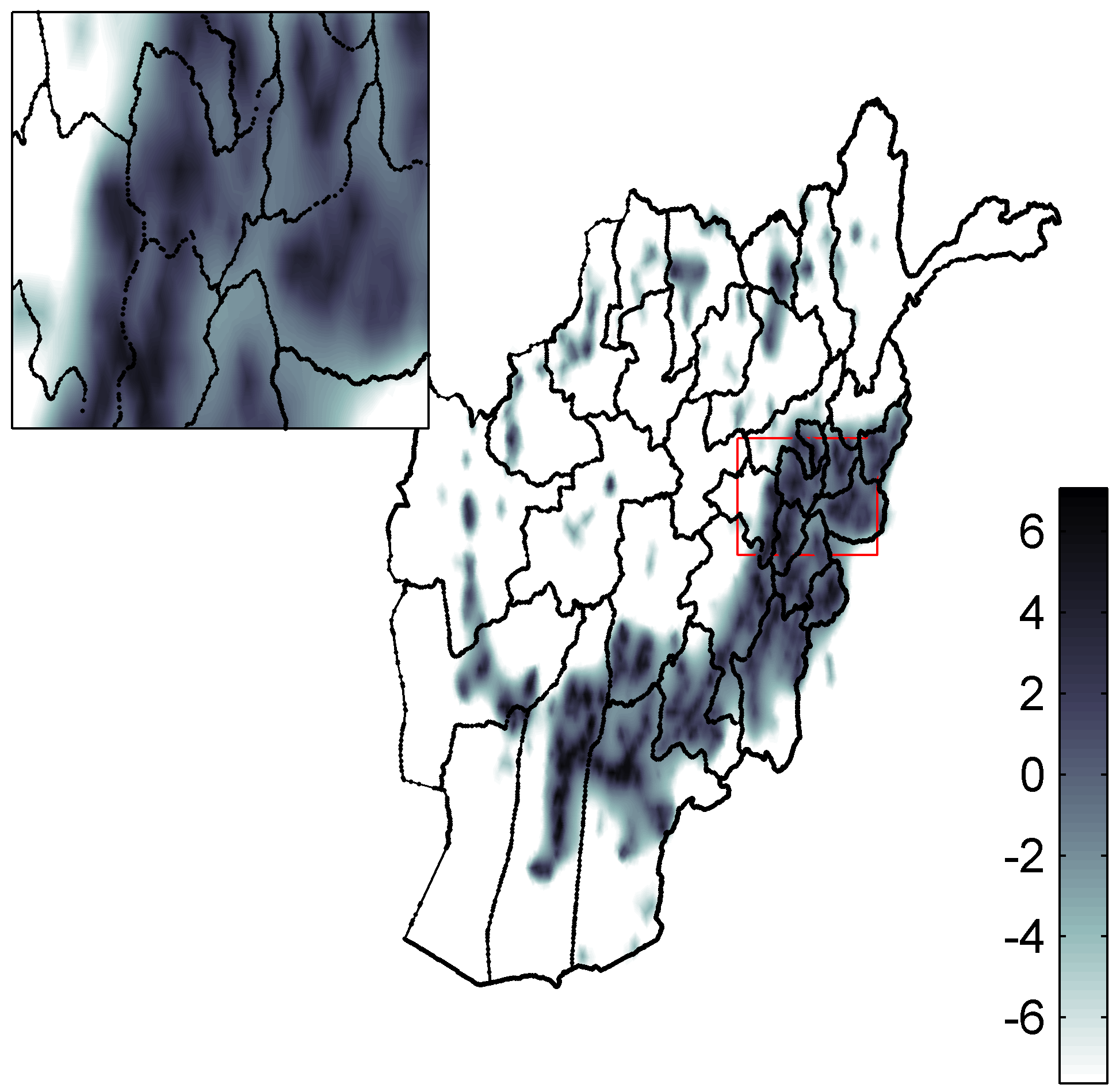}} 
		\end{tabular}
	\end{center}
\caption{\small The mesh and one time-slice log intensity map corresponding to the AWD on the first week of October 2009.}
\label{FigAFG}
\end{sidewaysfigure}

Afghanistan has an area of over 500000 km$^2$ and the WikiLeaks data set contains over 70000 events. The mesh we employed (using population density as a proxy for mesh density), shown in Figure~\ref{FigAFG} has the largest triangles with sides of 22km and the smallest ones with sides of 7km. The total number of vertices amounts to $n = 9398$ in a system with $T = 313$ time points (weeks). 

We constructed $\vect{A}$ using a Galerkin reduction with a mass lumping method \citep{Bueche_2000,Lindgren_2011} of a diffusion equation. For illustration purposes, the diffusion constant was set to $D = 10^{-4}$ with latitude/longitude as spatial units (all of $\vect{A}$ will be estimated in the next sub-section). The matrix $\vect{Q}= 0.2 \times \vect{I}$ was taken as rough value from the full joint analysis using a low resolution model, see \cite{AZM2012b} for details. We carried out inference in the AWD with the {\it diag} algorithm, which took only a few hours on a standard PC and consumed only about 4GB of memory.  

A characteristic plot showing one week of the conflict progression (first week of October 2009) is given in Figure~\ref{FigAFG}. At this point, in the conflict, activity in the south in Helmand and Kandahar was reaching its peak and conflict at the Pakistani border was intensifying considerably. The insets clearly show how detailed inferences can be made. 

\subsubsection{Learning conflict dynamics from the AWD}

\begin{sidewaysfigure}
	\begin{center}
		\resizebox{\textheight}{!}{\includegraphics{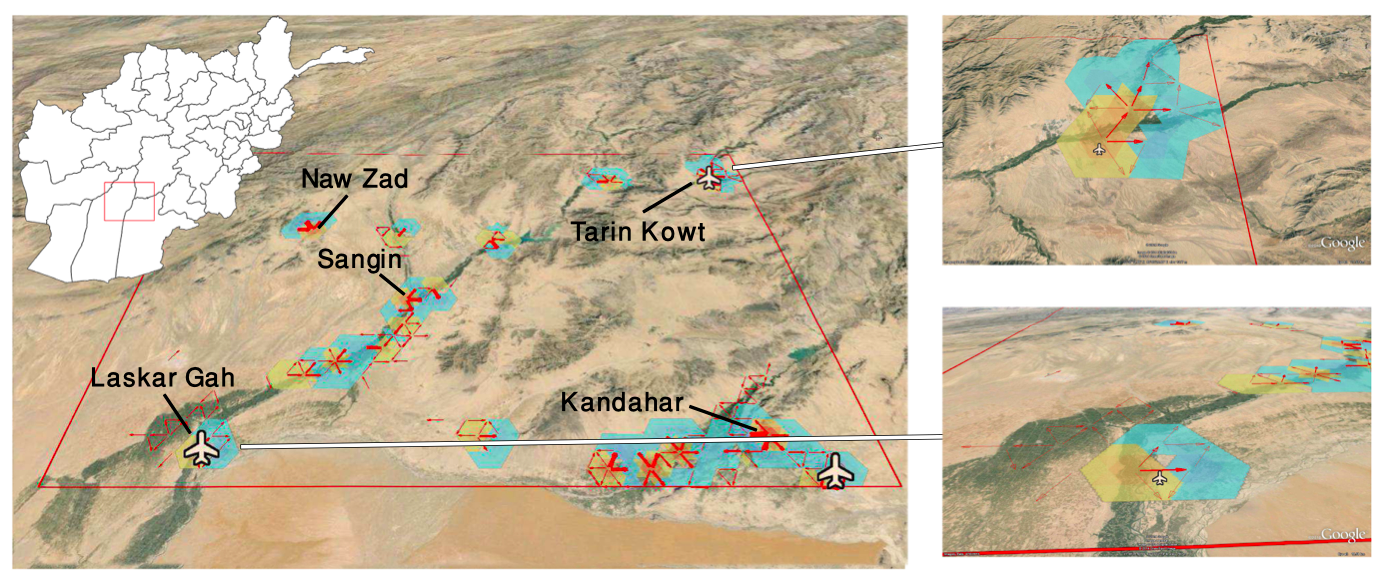}}		
	\end{center}
\caption{\small Left panel: The difference between the extent of sourcing (red) and sinking (blue) of conflict event intensity in part of Southern Afghanistan. The aeroplane symbols denote airport locations whilst the arrows denote the relevant edges in $\vect{A}$ with arrow thickness proportional to the element size. Right panel: Insets showing the detail at the airports near Laskar Gah and Tarin Kowt.}
\label{FigAFG2}
\end{sidewaysfigure}

In the context of conflict, the dynamic behaviour of events is usually extracted from the data by gridding the domain of interest in space and time and carrying out an empirical study on the events \emph{per se}. For example, one could analyse the pattern of cells which contain at least one event at two consecutive time frames \citep{Baudains_2013, Schutte_2011}. Unfortunately, due to an explicit reliance on the use of multiple observations to obtain a reliable estimator, these methods are only able to provide global assessments of the dynamics, that is, assert whether phenomena such as escalation or relocation are present everywhere on average. Here, on the other hand, we can provide spatially-resolved maps of conflict patterns and, moreover, are able to assign a probability to the presence or absence of dynamics.

The most important phenomena considered here are containment (events repeat in the same location), escalation (events repeat and also spill over into surrounding areas), and relocation (events move from one area to the next). All of these contagion phenomena may be interpreted directly from our posterior beliefs on $\vect{A}$ and the diagonal elements of $\vect{Q}$. For example, a vertex with low values of $q_{ii}$ and low values for the elements in its respective column in $\vect{A}$, is indicative of an area where events do occur, but that do not escalate or diffuse to surrounding areas. On the other hand, large values in $\vect{a}_{i}$ are indicative of an area that is susceptible to nearby conflict events whilst large values in a column of $\vect{A}$ are indicative of an area that is a large potential contributor to conflict contagion.

The last interpretation is particularly interesting, not only for retrospective analysis and prediction purposes, but also for generating insights into mechanisms which could be employed to contain contagion. One may summarise the role of a region in conflict by summing over the respective columns and rows in $\vect{A}$ (excluding the diagonal elements) in order to obtain a \emph{source} index and a \emph{sink} index for each vertex. The difference between these two indices can then be seen as a measure of how likely a region is to contribute to conflict in the surrounding areas and how likely conflict in a region is \emph{due to} conflict in a neighbouring area.

As a proof of concept, we employ the full model to obtain a map of contagion for Helmand using data between May 2006 and November 2009, shown in Fig. \ref{FigAFG2}. It is beyond the scope of this work to analyse the map in detail, however, three things are of note. First, although not evident from this figure, there is no direct correlation between event intensity and contagion, suggesting that the inference is able to distinguish between containment and relocation/escalation. Second, all airports in the vicinity (three in this case: Kandahar in the South East, Laskar Gah in the South West and Tarin Kowt in the North East) are highlighted as a \emph{source} of conflict, which is not surprising given the strategic importance of airbases. Third, several of the source hot spots are on towns and villages which have played a prominent role in the Afghan conflict, these include Naw Zad in the North West, the location of multiple offensive operations by the International Security Assistance Force (ISAF) in the latter part of the conflict and Sangin, one of the most hotly contested towns in the conflict. We note that the interpretation of the inferred dynamics is fully dependent on the spatial and temporal resolutions we employ, and may change for different mesh sizes and temporal discretisations.

To evaluate whether the inferred connectivity in $\vect{A}$ makes any improvement on the approach where the evolution of ${x}_{t}^{i}$ are considered independent (diagonal $\vect{A}$), we propose to use the 
one step ahead predictive probabilities. We proceed as follows: we infer a connected $\vect{A}$, $\vect{A}_{\text{conn}}$, using a {\it chordal} method for state inference, and we also infer a diagonal $\vect{A}$, $\vect{A}_{\text{indep}}$. We then use the mean values  $\BE{\vect{A}_{\text{conn}}}$ and $\BE{\vect{A}_{\text{indep}}}$ to compute the predictive probabilities. Note that in the latter case, the inference completely decouples into independent inference tasks for all $\{x_{t}^{i} \}_{t}$~and~$a_{ii}$. 

The one step ahead predictive probability at time $t$  is given by
{\small
\begin{align*}
	p(\mathcal{Y}_{t+1} \mid \mathcal{Y}_{1:t}, \vect{A}) = \int \!d\vect{x}_{t,t+1} \: p(\mathcal{Y}_{t+1} \mid \vect{x}_{t+1}) \mathcal{N}(\vect{x}_{t+1};\vect{A}\vect{x}_t, \vect{Q}^{-1}) \mathcal{N}(\vect{x}_{t};\vect{Q}_{\alpha_t}^{-1}\vect{h}_{\alpha_t}, \vect{Q}_{\alpha_t}^{-1}),
\end{align*}
}

\vspace{-0.75cm}
\noindent
where $\vect{h}_{\alpha_t}$ and $\vect{Q}_{\alpha_t}$ denote the canonical parameters of the $\alpha_t(\vect{x}_{t})$ forward message corresponding to the  filtering algorithm. Clearly, the above quantity is not tractable, therefore, we use the corresponding marginal likelihood approximation following from our approach. The integral itself corresponds to expectation propagation based marginal likelihood approximation in latent Gaussian models and is known to be a good quality approximation for a variety of (pseudo) likelihood terms  \citep[e.g.,][]{KussRasmussen2005, RWGP2005}. Similar latent Gaussian models where this approximation is shown to perform excellently are  the stochastic volatility and sparial log-Gaussian Cox process models in \cite{CsekeHeskes2010b}. The cumulative log values of the one step ahead prediction approximations approximate the log evidence, and thus we can also use them to do model comparison. In this way we can assess which model better explains the data. 

\begin{figure}
	\begin{center}
		\resizebox{0.75\textwidth}{!}{\includegraphics{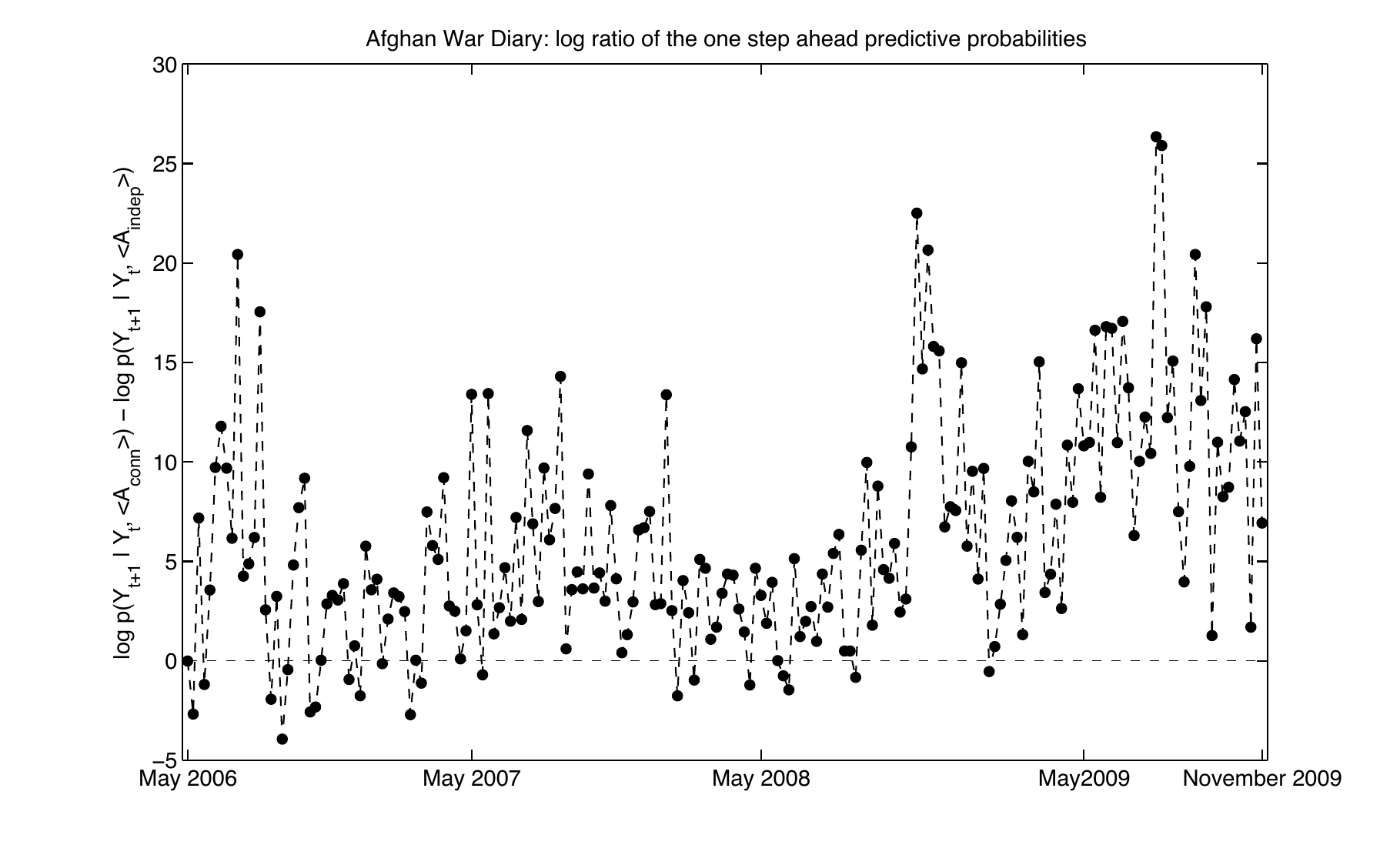}}		
	\end{center}
\caption{\small The log ratio of the (approximate) one step ahead predictive probabilities given by the approximations of $\log p(\mathcal{Y}_{t+1} \mid \mathcal{Y}_{1:t},   \BE{\vect{A}_{\text{conn}}}) - \log p(\mathcal{Y}_{t+1} \mid \mathcal{Y}_{1:t},  \BE{\vect{A}_{\text{indep}}})$ for the AWD data. The plot shows that the connected model generally achieves better predictive performance and that the connected model is more likely overall---the sum of log ratios is clearly positive.}
\label{FigAFGOSAP}
\end{figure}

Figure~\ref{FigAFGOSAP} shows how the one step ahead predictions $p(\mathcal{Y}_{t+1} \mid \mathcal{Y}_{1:t},   \BE{\vect{A}_{\text{conn}}})$ and $p(\mathcal{Y}_{t+1} \mid \mathcal{Y}_{1:t},  \BE{\vect{A}_{\text{indep}}})$ compare. The plot shows that the corresponding predictive  log likelihood ratio is positive for most times and  that the overall log likelihood ratio (sum of one step ahead log likelihood ratios) is positive. Therefore, our qualitative conclusions about the benefit of learning micro diffusions are supported  by quantitative evidence: the (approximated) predictive performance of the model increases and a connected model is more likely than an independent one. As  future work, we intend to  focus on areas of high conflict intensity to assess how the learned conflict dynamics varies w.r.t. the spatial resolution of the model.

\section{Conclusions} \label{SecConc}

In this paper we propose a family of approximate inference methods for spatio-temporal log-Gaussian Cox process models; the algorithms are based on variational approximate inference methods and approximate message passing. Note that the method can be applied to any similar latent Gaussian model \eqref{EqnJointModel} with general (pseudo) likelihood ${\psi}_{t,i}(x_{t}^{i})$.
We show how the sparsity in the underlying dynamic model can be exploited in order to overcome the limitations in the standard forward-backward and block inference methods which can become prohibitive for large $n$ and $T$.


In this paper we employ two layers of approximation. The first is the variational approximation to the posterior distribution, which is factored across the states and parameters. The second one addresses the non-tractability and computational issues associated with the resulting variational distribution over the states, $q_X$. Approximations due to the variational method and the EP updates for intractable likelihoods have been used extensively with considerable success in several applications. As such, one could argue that the additional approximation used in this work in order to retain sparsity in the message updates introduces further errors that may be hard to quantify. However, the advantage of the new methods is that they provide a  wide range of options w.r.t. accuracy and both computational and storage  complexity; in particular we propose using messages with chordal precision structures that serve as a good compromise in complexity between schemes using diagonal and full precision matrix structures. A beneficial aspect of this framework is that one can always choose to do away with these new approximations and revert to the \emph{full} scheme when this is not computationally prohibitive. Finally, as shown in the Supplementary Material, these layers of approximation naturally follow from embedding the variational method into the expectation constrained framework.

We applied the proposed methods to model conflict data and we showed that by using the increased resolution resulting from our methods we can detect micro-diffusions in the Afghan War Diary data. By learning these diffusion effects we can improve the predictive performance and obtain plausible qualitative interpretations of conflict contagion. The proposed methodology can also be applied to  epidemic or environmental studies where sparse latent spatial diffusion models---linear diffusions or linear approximations---can be formulated. 

In the future we intend to explore the set of structures that lie between the  fast and less accurate spanning tree and the somewhat larger chordal structures employed in this work. Currently, we are working on improving the distributed scheduling presented in Section~\ref{SecScheduling}. This is the most important  area of further research as the proposed message-passing algorithm, by design, is particularly well suited to take advantage of distributed computing environments. 
The ease of parallelisation is a major strength of our approach w.r.t. block approaches relying on partial matrix inversions, as it is unclear how solution of the Takahashi equations could be distributed.
Another notable advantage of our approach, when compared to the block model, is that it more adaptable to online settings, that is, to cases where we have streamed data.

\newpage

\appendix

\begin{center}
{  {\LARGE Sparse Approximate Inference for Spatio-Temporal \\ Point Process Models} \\ {\small Supplementary Material}}
\end{center}

\section{Derivation of the message-passing algorithm for $q_{X}$}\label{SecMP}

We reiterate that the massage passing algorithm for the factor graph corresponding to the density $q_{X}$ (see Section 3.1 and Figure 2) is written as
\begin{align}
	\lambda_{t+1, j}^{0}(x_{t+1}^{j}) \propto & \: \psi_{t+1,j}(x_{t+1}^{j}), \label{EqnMargFree}
	\\
	\xi_{t+1}(\vect{x}_{t+1}) \propto & \int\!d\vect{x}_{t}\:  \Psi_{t, t+1}(\vect{x}_{t}, \vect{x}_{t+1}) \hat{\xi}_{t}(\vect{x}_{t}), \label{EqnXiFree}
	\\	
	\eta_{t}(\vect{x}_{t}) \propto & \int\!d\vect{x}_{t+1}\:  \Psi_{t, t+1}(\vect{x}_{t}, \vect{x}_{t+1}) \hat{\eta}_{t+1}(\vect{x}_{t+1}), \label{EqnEtaFree}
	\\
	\lambda_{t+1, j}^{l}(x_{t+1}^{j}) \propto & \: \int\! d\vect{x}_{t+1}^{\backslash j} \: \xi_{t+1}(\vect{x}_{t+1}) \eta_{t+1}(\vect{x}_{t+1}),  \prodlow{k \neq j } \lambda_{t+1, k}^{0}(x_{t+1}^{k})  \label{EqnLamFree}
	\\
	\hat{\xi}_{t}(\vect{x}_{t})  \propto & \:	\xi_{t}(\vect{x}_{t})  \prodlow{j} \lambda_{t, j}^{0}(x_{t}^{j})  \label{EqnXiHatFree},
	\\
	\hat{\eta}_{t+1}(\vect{x}_{t+1})  \propto & \: \eta_{t+1}(\vect{x}_{t+1}) \prodlow{j} \lambda_{t+1, j}^{0}(x_{t+1}^{j}).  \label{EqnEtaHatFree}
\end{align}
and the corresponding marginals are
\begin{align}
	q_{X}(\vect{x}_{t}, \vect{x}_{t+1}) \propto &\: \Psi_{t, t+1}(\vect{x}_{t}, \vect{x}_{t+1}) \hat{\xi}_{t}(\vect{x}_{t}) \hat{\eta}_{t+1}(\vect{x}_{t+1}), 
	\label{EqnTwoSliceFree}	
	\\
	q_{X}(x_{t+1}^{j}) \propto &  \: \psi_{t+1,j}(x_{t+1}^{j})\lambda_{t+1, j}^{l}(x_{t+1}^{j}),\label{EqnOneSliceMargFree} 
	\\
	q_{X}(\vect{x}_{t}) \propto & \: \xi_{t}(\vect{x}_{t})\eta_{t}(\vect{x}_{t})  \prodlow{j} \lambda_{t, j}^{0}(x_{t}^{j}). \label{EqnOneSliceFree}.
\end{align}

The messages $\lambda_{t+1, j}^{0}$, $\xi_{t+1}$, and $\eta_{t}$ are factor-to-variable messages while the messages $\lambda_{t+1, j}^{l}$, $\hat{\xi}_{t}$ and $\hat{\eta}_{t+1}$ are variable-to-factor messages. The message-passing algorithm computes the marginals $q_{X}(\vect{x}_{t}, \vect{x}_{t+1})$, $q_{X}(x_{t+1}^{j}) $ and $q_{X}(\vect{x}_{t})$ of $q_{X}$.

In the following we show how the message passing can be re-written into a form that helps us to introduce approximate message-passing algorithms. We multiply \eqref{EqnXiFree} by $\prodlow{j} \lambda_{t+1, j}^{0}(x_{t+1}^{j})$ and $\eta_{t+1}(\vect{x}_{t+1})$ and using \eqref{EqnXiHatFree},  \eqref{EqnEtaHatFree} and \eqref{EqnTwoSliceFree} we obtain
\begin{align*}
	\hat{\xi}_{t+1}(\vect{x}_{t+1}) \eta_{t+1}(\vect{x}_{t+1}) \propto \int\!d\vect{x}_{t} \: q_{X}(\vect{x}_{t}, \vect{x}_{t+1}).
\end{align*}
In a similar fashion, by multiplying both sides of \eqref{EqnEtaFree} by $\hat\xi_t(\vect{x}_t)$, we obtain
\begin{align*}
	\hat{\xi}_{t}(\vect{x}_{t}) \eta_{t}(\vect{x}_{t}) \propto \int\!d\vect{x}_{t+1}\: q_{X}(\vect{x}_{t}, \vect{x}_{t+1}).
\end{align*}
We rename $\alpha_{t+1} = \hat{\xi}_{t+1}$ and $\beta_{t+1} =  \eta_{t+1}$ to obtain the equations (16) and (17) in the main text. By using \eqref{EqnMargFree} and \eqref{EqnOneSliceMargFree},  and \eqref{EqnLamFree} and \eqref{EqnOneSliceFree}, respectively, we can write 
\begin{align*}
	\lambda_{t+1, j}^{0}(x_{t+1}^{j}) \lambda_{t+1, j}^{l}(x_{t+1}^{j})  = q_{X}(x_{t+1}^{j}) \quad \text{and} \quad \lambda_{t+1, j}^{0}(x_{t+1}^{j}) \lambda_{t+1, j}^{l}(x_{t+1}^{j})  = \int\! d\vect{x}_{t+1}^{\backslash j}  q_{X}(x_{t+1}^{j}).
\end{align*}
Using the identity $q_{X}(\vect{x}_{t+1}) = \int\!d\vect{x}_{t+1}q_{X}(\vect{x}_{t}, \vect{x}_{t+1})$  we arrive to the set of update equations presented in the main text.

\section{Algorithmic details of the chordal graph projections}

For chordal $G(\vect{f})$ the equations satisfying the optimality conditions can be solved as follows, see  \cite{Dahl2008} for further details.  Assume that the cliques $C_{1}, \ldots, C_{K}$ of the  $G(\vect{f})$'s junction tree are ordered such that if $C_i$ is an ancestor of $C_{j}$ then $i\leq j$. Let $S_j = C_j \cap \{C_1 \cup C_2 \cup \ldots \cup C_{j-1}\}$ and $R_j = C_j \setminus \{C_1 \cup C_2 \cup \ldots \cup C_{j-1}\}$. Let $\hat{\vect{Q}}$ be the precision matrix with constrained structure that we are solving for. Then  $\hat{\vect{Q}} = (\vect{I} +\vect{U})\vect{D}(\vect{I}+\vect{U})^{T}$, where $\vect{I}$ is the identity matrix, and $\vect{U}$ and $\vect{D}$ can be computed iteratively  from
\begin{eqnarray*}
	\vect{U}_{R_k, S_k} &=& - \vect{V}_{S_k,S_k}^{-1}\vect{V}_{S_k,R_k}, 
	\\
	\vect{D}_{R_k,R_k} &=& [\vect{V}_{R_k,R_k} - \vect{V}_{R_k,S_k} \vect{V}_{S_k,S_k}^{-1} \vect{V}_{S_k,R_k}]^{-1}. 
\end{eqnarray*}
The computational complexity scales with  $\sum_{k} \max\{\mid S_k \mid^{3}, \mid R_k \mid^{3}\}$. The size of the cliques depends on the structure of $G(\vect{f})$. 
See Section~{3.1.2} for further details. 



\section{Variational inference with expectation constraints}

In this section we detail the variational inference approach we propose, and derive the update equations used in the paper from an expectation constraints inference point of view. The conceptual advantage of the expectation constraints framework is that we start off from a global objective function that we the optimise (using Lagrange multipliers). The optimisation updates are identical to those implied by the approximate message passing scheme presented in the main text.

In Section \ref{sec:AppModel} we re-visit the model used in the main work, and extend it slightly to allow for the presence of another matrix $\vect{B}$ that plays the same role as $\vect{A}$ but for exogenous (known) inputs to the system. In Section \ref{SecVarInf} we then explicitly write down the variational posteriors. Finally, in Section in Section \ref{SecFreeOpt} we derive the approximate inference scheme, this time from the expectation constraints point of view.

\subsection{Model details}\label{sec:AppModel}
In this section we consider the model defined in the main text, made more general through the introduction of a set of  control variables  $\vect{U} = \{\vect{u}_{t}\}_{t}, \vect{u}_{t} \in \mathbb{R}^{m}$, that influence the field $\{\vect{x}_{t}\}_t$ through a set of (unknown) coefficients in $\vect{B} \in \mathbb{R}^{n\times m}$. We treat $\vect{Q}$ as a generic matrix (i.e., not necessarily a diagonal matrix), and use $p(\vect{Q} \mid \vect{\theta})$ to denote its prior distribution. We do not address inference of $\vect{Q}$ in detail, but in Section \ref{SecVarInf} we give the formulae for the free-form variational update. 

The model for which we derive the approximate inference can be written as 
\begin{align}\label{EqnJointModel}
 	p(\mathcal{Y}, \vect{X}, \vect{A}, \vect{Z}, \vect{B}, \vect{Q} \mid \vect{\theta}, \vect{U}) =\: & p(\vect{x}_{1})\prodlow{t} p(\vect{x}_{t+1}\mid \vect{x}_{t}, \vect{A}, \vect{B}, \vect{u}_{t}, \vect{Q}) \prodlow{j}\tilde{\psi}_{t+1,j}(x_{t+1}^{j}) 
	\\
	& \times \prodlow{i\sim j} p(a_{ij}\mid z_{ij}, \vect{\theta})p(z_{ij} \mid \vect{\theta})\prodlow{k,l}p(b_{kl} \mid \vect{\theta}) p(\vect{Q} \mid  \vect{\theta}), \nonumber
\end{align}
with the hyper-parameters $\vect{\theta} = \{\vect{m}_{1}, \vect{V}_{1}, v_{\text{slab}},  p_{\text{slab}}, v_{b}, \vect{\theta}_Q \}$ collecting the parameters of
\begin{align*}
 	p(\vect{x}_{1}) = &\: \mathcal{N}(\vect{x}_{1}; \vect{m}_{1}, \vect{V}_1),  
	\\
	p(\vect{x}_{t+1}\mid \vect{x}_{t}, \vect{A}, \vect{B}, \vect{u}_{t}, \vect{Q}) = & \: \mathcal{N}(\vect{x}_{t+1} ;\vect{A}\vect{x}_{t} + \vect{B}\vect{u}_{t}, \vect{Q}^{-1}),  
	\\
	p({a}_{ij} \mid z_{ij}, v_{\text{slab}}) = & \: \mathcal{N}(a_{ij}; 0,v_{\text{slab}})^{z_{ij}}\delta(a_{ij})^{1 - z_{ij}},
	\\
	p(z_{ij} \mid \vect{\theta}) = &\: {Ber}(z_{ij}; p_{\text{slab}}),  
	\\
	p(\vect{Q} | \vect{\theta}) = &\: p(\vect{Q} | \vect{\theta}_Q) .
\end{align*}
Since the hyper-parameters are not  explicitly relevant to the inferential framework, hereafter we omit them to keep the presentation simple.

\subsection{Free-form variational inference}\label{SecVarInf}

By using the (structured) variational Bayes method we can approximate the  joint density \\ $p(\vect{X}, \vect{A},\vect{Z}, \vect{B},  \vect{Q} \mid \mathcal{Y}, \vect{U}) $ with a factorised density $q_{X}(\vect{X})q_{AZ}(\vect{A}, \vect{Z})q_{B}(\vect{B})q_{Q}(\vect{Q})$. Variational methods approximate marginal densities by recasting the inference problem as an optimisation of the divergence $\D{}{\cdot}{p}$. Similar to the main text, for our model, the (structured variational) inference is formulated as the optimisation
\begin{equation}\label{EqnVarObj}
	\mathop{\text{minimise}}\limits_{q_{X},q_{AZ}, q_{Q}, q_{B}} \D{}{q_{X}(\vect{X})q_{AZ}(\vect{A}, \vect{Z})q_{Q}(\vect{Q})q_{B}(\vect{B})}{p(\vect{X}, \vect{A},\vect{Z}, \vect{B}, \vect{Q} \mid \mathcal{Y}, \vect{U})}.
\end{equation}
The stationarity condition of this minimisation problem yields the structured variational mean field updates
\begin{align}
	{[q_{X}(\vect{X})]}^{new} &\propto \exp\BC{ \BE{\log p(\vect{X}, \vect{A},\vect{Z}, \vect{Q}, \vect{B} \mid \mathcal{Y}, \vect{U})}_{q_{AZ},q_{Q}, q_{B}}}, \label{EqnDefX}
	\\
	{[q_{AZ}(\vect{A},\vect{Z})]}^{new} &\propto \exp\BC{ \BE{\log p(\vect{X}, \vect{A}, \vect{Z}, \vect{Q}, \vect{B} \mid \mathcal{Y}, \vect{U})}_{q_{X},q_{Q}, q_{B}}}, \label{EqnDefA}
	\\ 
	{[q_{B}(\vect{B})]}^{new} &\propto \exp\BC{ \BE{\log p(\vect{X}, \vect{A}, \vect{Z},  \vect{Q}, \vect{B} \mid \mathcal{Y}, \vect{U})}_{q_{X},q_{AZ}, q_{Q}}}, \label{EqnDefB}
	\\
	{[q_{Q}(\vect{Q})]}^{new} &\propto \exp\BC{ \BE{\log p(\vect{X}, \vect{A},\vect{Z},  \vect{Q}, \vect{B} \mid \mathcal{Y}, \vect{U})}_{q_{X},q_{AZ}, q_{B}}}. \label{EqnDefC}
\end{align}
This coordinate-wise descent is guaranteed to converge to a local optimum. However, due to the form of the likelihood and prior terms, some of the densities above can be analytically intractable. For example, in the case of the model presented in the paper, $q_{X}(\vect{X})$ is analytically intractable due to the non-Gaussian likelihood terms.  A standard approach to circumvent this intractability is to restrict the distributions of interest to parameterised forms and optimise the objective w.r.t. these parameters. In our case this translates to restricting $q_{X}$, $q_{AZ}$ and $q_{B}$ to the Gaussian family, $q_{Q}$ to an inverse Wishart (a product of Gammas in the diagonal case) and performing the optimisation w.r.t. their canonical/moment parameters.  

However, when viewing $q_{X}, q_{AZ}$ and $q_{B}$ as graphical models or factor graphs various versions of the expectation propagation algorithm can perform much better (faster convergence, higher accuracy). Therefore, we formulate a modified variational approach that results in (i)  the above variational updates for  $q_{X}, q_{AZ}$ and $q_{B}$ and (ii) expectation propagation algorithms for each model $q_{X}, q_{AZ}$ and $q_{B}$ to compute the expected values needed on the r.h.s of \eqref{EqnDefX}-\eqref{EqnDefC}.  To do this,  we embed the variational framework above into a variational framework with expectation constraints \citep{Heskes2005}. 


In order to introduce the approximate inference method from an expectation constraints point of view, we take a closer look at the form of the densities of \eqref{EqnDefX}-\eqref{EqnDefC} in the next section. In order to simplify the presentation, we introduce an alternative notation for expectations w.r.t. products of densities that is consistent with the $\BE{\cdot}_{q}$ notation of the paper.  Let $F$ be an arbitrary function of $n$ random variables $z_1, \ldots, z_n$ and let $q_1(z_1), \ldots, q_{n}(z_n)$ be their respective densities. Then, we use $\BE{F(z_1, \ldots, z_{i}, \ldots, z_n)}_{\setminus q_i}$ to denote $\BE{F(z_1, \ldots, z_{i}, \ldots, z_n)}_{q_1, \ldots, q_{i-1}, q_{i+1}, \ldots ,q_{n}}$. For example, in \eqref{EqnDefX} we write the expectation term as $\BE{\log p(\vect{X}, \vect{A},\vect{Z}, \vect{Q}, \vect{B} \mid \mathcal{Y}, \vect{U})}_{\setminus q_{X}}$.

\subsubsection{The variational posterior $q_{X}$}
From \eqref{EqnDefX} we obtain a $q_{X}$ that can be re-written as
\begin{align*}
	q_{X}(\vect{X}) \propto \prodlow{t}\Psi_{t,t+1}(\vect{x}_{t}, \vect{x}_{t+1}) \prodlow{j} \psi_{t+1,j}(x_{t+1}^{j}),
\end{align*}
where $\log \Psi_{t,t+1}(\vect{x}_{t}, \vect{x}_{t+1}) = \BE{\log \{ \mathcal{N}(\vect{x}_{t+1} \mid \vect{A}\vect{x}_{t}+\vect{B}\vect{u}_{t}, \vect{Q}^{-1})\times \exp(\vect{h}_{t+1}^{T} \vect{x}_{t+1} ) \}}_{\setminus q_{X}}$. This is the form of a latent Gaussian model, where the non-Gaussian terms depend only on the variables $x_{t}^{j}$.

The factor $\Psi_{t,t+1}$ is Gaussian with  canonical parameters $\BE{\vect{h}_{t,t+1}}_{\setminus q_{X}}$ and
$\BE{\vect{Q}_{t,t+1}}_{\setminus q_{X}}$ given by
\begin{align*}
	\vect{h}_{t,t+1} &= \left[\begin{array}{cc}
							 -\vect{A}^{T}\vect{Q}\vect{B}\vect{u}_{t} + \delta_{t,1}\vect{h}_{1}
							 \\
							 \vect{Q}\vect{B}\vect{u}_{t} +  \vect{h}_{t+1}
						\end{array} \right],
	\\
	\vect{Q}_{t,t+1} &= \left[\begin{array}{cc}
							\vect{A}^{T}\vect{Q}\vect{A}+\delta{t,1}\vect{Q}_{1} & - \vect{A}^{T}\vect{Q}
							\\ 
							-\vect{Q}\vect{A}  & \vect{Q}
						\end{array} \right].
\end{align*}
Here $\vect{h}_1 = \vect{V}_{1}^{-1}\vect{m}_{1}$ and $\vect{Q}_{1} = \vect{V}_{1}^{-1}$ denote the canonical parameters of the initial conditions and $\delta_{t,1}$ is the Kronecker delta function, that is, $\delta_{t,1} = 1$ if $t=1$ and $\delta_{t,1} = 0$ otherwise.

\subsubsection{The  variational posterior $q_{AZ}$}
Let $\vect{a}_{\cdot,j}$ denote the $j$-th column of $\vect{A}$, $\vect{A} = [\vect{a}_{\cdot, 1}, \ldots, \vect{a}_{\cdot,n}]$, and $[\vect{A}]_{c}$ be the column ordered vectorised form of $\vect{A}$, namely  $[\vect{A}]_{c}^{T} = [\vect{a}_{\cdot,1}^{T}, \ldots, \vect{a}_{\cdot,n}^{T}]$. Then it follows that
\begin{align*}
	q_{AZ}(\vect{A}, \vect{Z}) \propto \exp\BC{ \BE{\vect{h}_{A}}_{\setminus q_{AZ}}^{T} [\vect{A}]_{c} - \frac{1}{2}  [\vect{A}]_{c}^{T} \BE{\vect{Q}_{A}}_{\setminus q_{AZ}}  [\vect{A}]_{c} } \times \prodlow{ij}p_{0}(a_{ij},z_{ij} \mid \vect{\theta}),
\end{align*}
where 
\begin{align*}
	\vect{h}_{A} &= \BS{\vect{Q}\sumlow{t}\vect{x}_{t+1}\vect{x}_{t}^{T} - \vect{Q}\vect{B}\sumlow{t}\vect{u}_{t}\vect{x}_{t}^{T}}_{c}, 
	\\
	\vect{Q}_{A} &=  \sumlow{t} \vect{x}_{t}\vect{x}_{t}^{T}  \otimes \vect{Q}.
\end{align*}

This variational posterior represents that of a multivariate conditional Gaussian model, thus marginalisation and inference is generally intractable for high dimensional models because of the exponential growth w.r.t. the dimensionality of $[\vect{A}]_{c}$. However, when $\vect{A}$ and  $\vect{Q}$ are sparse  there is a significant decrease in the effective size of the model ($[\vect{A}]_{c}$ is of much smaller dimension than $n^2$, and under our assumptions its dimension scales with  $n$). Moreover, when $\vect{Q}$ is diagonal, the distribution $q_{AZ}(\vect{A}, \vect{Z})$ factorises over the rows $\vect{a}_{i}$ and $\vect{z}_{i}$ of $\vect{A}$ and $\vect{Z}$ respectively; that is, we have  $q_{AZ}(\vect{A}, \vect{Z}) = \prod_{i} q_{AZ}(\vect{a}_{i}, \vect{z}_{i})$. In the latter case the number of structural non-zeros in $\vect{a}_{i}$ is  typically lower than $10$, thus inference can be done exactly---in the worst case we have  $2^{10}$ mixture components of dimension $10$ when computing the marginals $q_{AZ}^{i}(\vect{a}_i)$ needed for the updates. For non-diagonal, but sparse $\vect{Q}$, we still have a sparse latent Gaussian model with mixture priors and expectation propagation can be applied \citep{CsatoMIXTURE2005, Lobato2013}. 


\subsubsection{The  variational posterior $q_{B}$}

The distribution $q_{B}$ has a similar structure to that of $q_{AZ}$  and is given by
\begin{align*}
	q_{B}(B) \propto \exp\BC{ \BE{\vect{h}_{B}}_{\setminus q_{B}}^{T} [\vect{B}]_{c} - \frac{1}{2}  [\vect{B}]_{c}^{T} \BE{\vect{Q}_{B}}_{\setminus q_{B}}  [\vect{B}]_{c} } \times \prodlow{ij}p(b_{ij} \mid \vect{\theta}),
\end{align*}
with 
\begin{align*}
	\vect{h}_{B} &= \BS{\vect{Q}\sumlow{t}(\vect{x}_{t+1} - \vect{A}\vect{x}_{t})\vect{u}_{t}^{T}}_{c}, 
	\\
	\vect{Q}_{B} &=  \sumlow{t}\vect{u}_{t}\vect{u}_{t}^{T} \otimes \vect{Q}.
\end{align*}
This variational posterior is once again that of a latent Gaussian model where the possible non-Gaussian terms are the priors $p(b_{ij}\mid \vect{\theta})$ and inference can be done by expectation propagation.

\subsubsection{The  variational posterior $q_{Q}$}
This distribution is somewhat more complicated than the rest because it involves a log-determinant term, but suitable choices for the structure of $\vect{Q}$ and the prior $p(\vect{Q} \vert \vect{\theta})$ can make inference analytically tractable. The distribution $q_{Q}$ has the form
\begin{align*}
	q_{Q}(\vect{Q}) \propto \exp\BC{ \frac{1}{2} \log \det \vect{Q} - \frac{1}{2} \tr (\vect{Q}^{T}\vect{H}_{Q})} \times p(\vect{Q} \mid \vect{\theta}),
\end{align*}
where
\begin{align*}
	\vect{H}_{Q} =& \sumlow{t}\BE{\vect{x}_{t+1}\vect{x}_{t+1}}_{q_X} - 2\tr\BC{\BE{\vect{A}}_{{q}_{AZ}} \sumlow{t} \BE{\vect{x}_{t}\vect{x}_{t+1}^{T}}_{q_{X}}} + \tr \BC{ \BE{\vect{A}^{T}\vect{A}}_{q_{AZ}} \sumlow{t}\BE{\vect{x}_{t}\vect{x}_{t}^{T}}_{q_X}} 
	\\
	& -2 \BE{\vect{B}}_{q_{B}}\sumlow{t}\vect{u}_{t}(\BE{\vect{x}_{t+1}}_{q_X} - \BE{\vect{A}}_{q_{AZ}}\BE{\vect{x}_{t}}_{q_X})^{T} + \BE{\vect{B}^{T}{\vect{B}}}_{q_{B}}\sumlow{t}\vect{u}_{t}\vect{u}_{t}^{T}.
\end{align*}
In the model presented in the main text we use a diagonal structure with a Gamma prior, rendering inference analytically tractable.

\subsection{Variational inference with expectation constraints}\label{SecFreeOpt}
The objective in  \eqref{EqnVarObj} can be written as
\begin{align}\label{EqnExactFE}
 F(q_{X},q_{AZ},  q_{B} , q_{Q}) =& -\BE{\log  p(\vect{X}, \vect{A},\vect{Z}, \vect{Q}, \vect{B} \mid \mathcal{Y},  \vect{\theta})}_{q_{X},q_{AZ},  q_{B} , q_{Q}} 
 \\	&+ \BE{\log q_{X}(\vect{X})}_{q_{X}} + \BE{\log q_{AZ}(\vect{A},\vect{Z} )}_{q_{AZ}} + \BE{\log q_{B}(\vect{B})}_{q_{B}} +\BE{\log q_{Q}(\vect{Q})}_{q_{Q}}. \nonumber
\end{align}
With the exception of $q_{Q}$, all the above distributions are those of latent Gaussian models where exact inference can be intractable due to the non-Gaussian prior or likelihood terms. 
As mentioned above, instead of direct parametric optimisation we choose to recast the problem as an optimisation problem over approximate marginals. For each density $q_{X}, q_{AZ}$ and $q_{B}$ in Section \ref{SecqX}--\ref{SecqB} we introduce a set of approximate marginals and a corresponding entropy approximation. This results in an approximation of \eqref{EqnExactFE} that is optimised w.r.t. the approximate marginals. The resulting optimisation algorithm is  equivalent to the approximate inference algorithm introduced in the main text.

\begin{figure}
\begin{center}
	\begin{tabular}{cc}
		\resizebox{!}{0.25\textheight}{\includegraphics{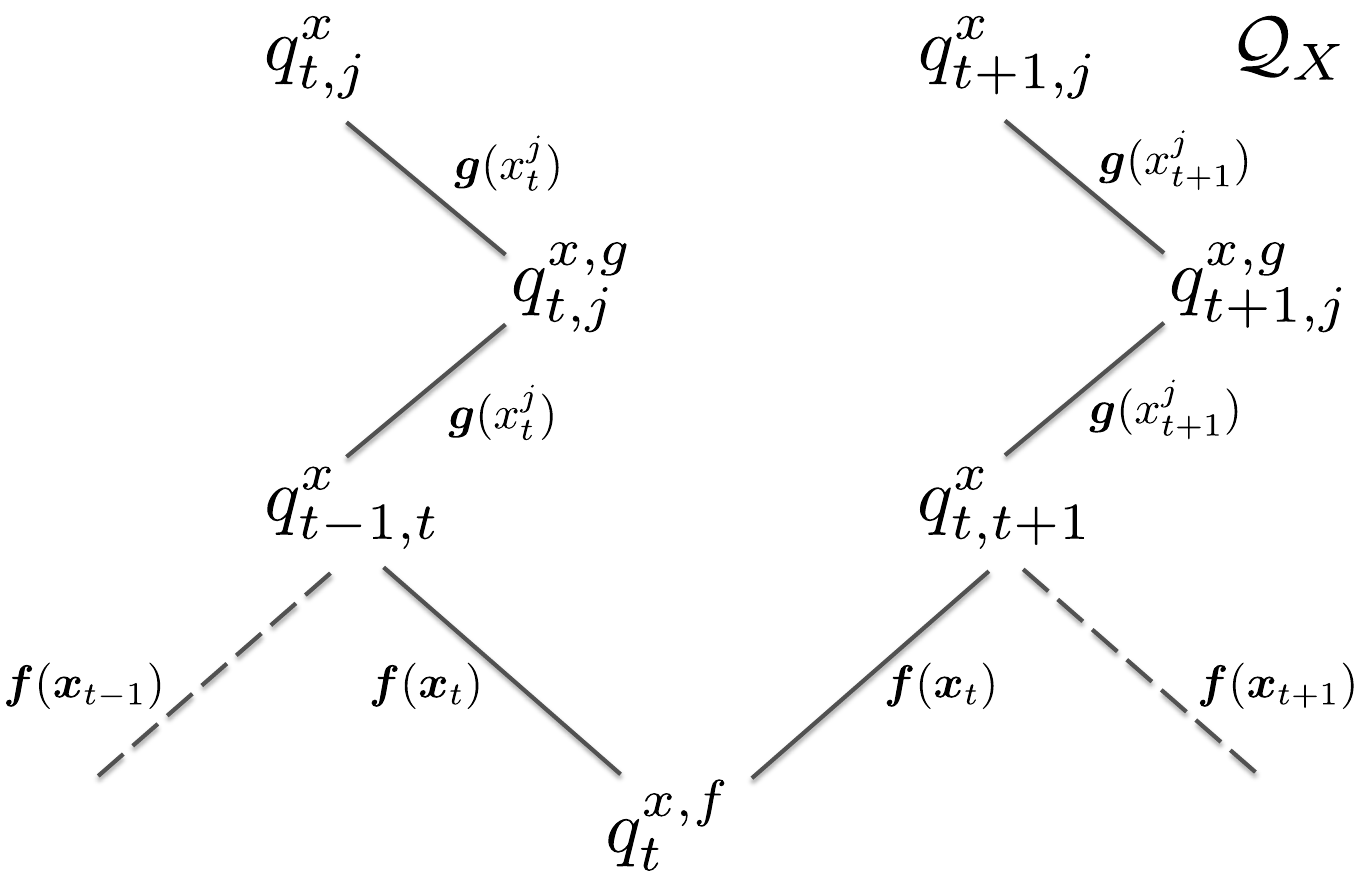}}		\hspace{0.5in}
		&
		\resizebox{!}{0.25\textheight}{\includegraphics{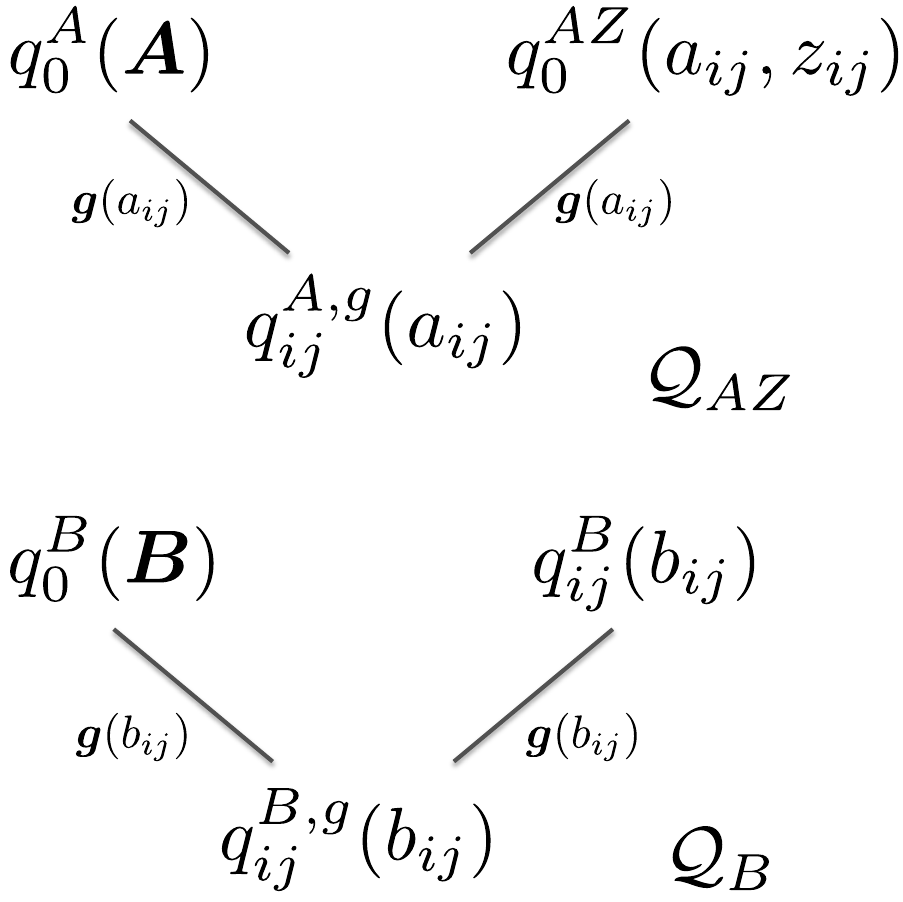}}
	\end{tabular}
\end{center}
\caption{\small Illustration of the sets of approximate marginals and the imposed expectation constraints. The left panel illustrates the set of approximate marginals $\mathcal{Q}_{X}$ and the corresponding expectation constraints over $\vect{g}(x_{t}^{j})$ and $\vect{f}(\vect{x}_{t})$. The panels on the right illustrate the approximate marginals for the sets $\mathcal{Q}_{AZ}$ and $\mathcal{Q}_{B}$ and the expectation constraints over $\vect{g}(a_{ij}) = (a_{ij}, -a_{ij}^{2}/2)$ and $\vect{g}(b_{ij}) = (b_{ij}, -b_{ij}^{2}/2)$, respectively.}
\label{FigAM}
\end{figure}

\subsubsection{Approximate marginals for $q_{X}$} \label{SecqX}

The distribution $q_{X}$ has a chain-structured factor graph with additional non-Gaussian terms (recall Figure 2 in the main text). \cite{YediWillFree2001} showed that message-passing algorithms on certain factor graph models can be formulated as fixed point iterations resulting from  the constrained optimisation of variational objectives derived from the variational free energy. We follow the line of thought in \cite{YediWillFree2001}  and  \cite{Heskes2005} to formulate approximate inference in $q_{X}$. By following the factorisation of $q_{X}$, we define a corresponding tree-structured set of approximate marginals 
\begin{align*}
	\mathcal{Q}_{X} = \{ \{q^{x}_{t,t+1}\}_{t}, \{q^{x,f}_{t}\}_{t}, \{q^{x}_{t,j}\}_{t,j}, \{q^{g}_{t,j}\}_{t,j} \},
\end{align*}
where we assign the  $q_{t,t+1}^{x}(\vect{x}_{t}, \vect{x}_{t+1})$ to the Gaussian factor $\Psi_{t,t+1}(\vect{x}_{t}, \vect{x}_{t+1})$,  and $q^{x}_{t+1, j}(\vect{x}_{t+1})$ to the non-Gaussian factor $\psi_{t+1,j}(\vect{x}_{t+1})$. 

Through the set $\mathcal{Q}_{X}$ we intend to represent a tree-structured distribution
\begin{align}\label{eq:qXtree}
 \frac{\prod_{t}q^{x}_{t,t+1} (\vect{x}_{t}, \vect{x}_{t+1})}
	{\prod_{t}q^{x,f}_{t+1}(\vect{x}_{t+1})} \times \prodlow{t,j} \frac{q^{x}_{t+1, j} (\vect{x}_{t+1}) }{q^{x,g}_{t+1,j}(\vect{x}_{t+1})}.
\end{align}
In contrast to the standard variational Bayes approach, the approximate marginals we introduce are not assumed to be independent (they share variables) and thus we need to impose some sort of local consistency constraints over their shared variables. Requiring them to have matching marginals, such as in \cite{YediWillFree2001},  would lead to the same intractability problem we encounter with the message-passing algorithm of Section~\ref{SecMP}.  For this reason, we relax the consistency constraints to expectation constraints as in \cite{Heskes2005}. Motivated by the latent Gaussian nature of $q_{X}$, we limit ourselves to imposing first- and second-order expectation constraints over the shared variables. For  $q^{x}_{t,t+1}$, $q^{x}_{t+1,t+2}$ and $q^{x,g}_{t+1,j}$ we postulate expectation constraints w.r.t. $\vect{g}(x_{t+1}^{j}) = (x_{t+1}^{j} ,-[x_{t+1}^{j}]^2/2)$ to deal with non-Gaussianity, while for 
$q^{x}_{t-1,t}$, $q^{x}_{t,t+1}$ and $q^{x,f}_{t}$ we introduce expectation constraints over a sparse Gaussian Markov random field $\vect{f}(\vect{x}_{t}) = (\{x_t^j\}_i, \{-[x_t^j]^2/2\}_i,\{-x_t^i x_t^{j}/2\}_{i\sim j})$ where $i \sim j$ denotes an adjacency relation in a graph $G(\vect{f})$ specified in the main text. Note that a fully connected $G(\vect{f})$ corresponds to constraining across all first- and second-order sufficient statistics. 
Formally, we express these expectation constraints as 
\begin{equation} \label{EqnECX1}
	\BE{\vect{f}(\vect{x}_{t+1})}_{q^{x}_{t,t+1}} =  \BE{\vect{f}(\vect{x}_{t+1})}_{q^{x,f}_{t+1}} \quad \text{and} \quad \BE{\vect{f}(\vect{x}_{t+1})}_{q^{x}_{t+1,t+2}} =  \BE{\vect{f}(\vect{x}_{t+1})}_{q^{x,f}_{t+1}}, 
\end{equation}
and
\begin{equation}\label{EqnECX2}
	\BE{\vect{g}(\vect{x}_{t+1})}_{q^{x}_{t,t+1}} =  \BE{\vect{g}(\vect{x}_{t+1})}_{q^{x,g}_{t+1,j}} \quad \text{and} \quad  \BE{\vect{g}(\vect{x}_{t+1})}_{q^{x}_{t+1,j}} =  \BE{\vect{g}(\vect{x}_{t+1})}_{q^{x,g}_{t+1,j}}. 
\end{equation}
The approximate marginals in $\mathcal{Q}_{X}$ and the corresponding constraints are illustrated in Figure~\ref{FigAM}. 

The tree structure of the model \eqref{eq:qXtree} implies that the entropy term in \eqref{EqnExactFE} can be approximated as
\begin{align*}
	-\tilde{H}(\mathcal{Q}_{X}) = \sumlow{t}\BE{\log q^{x}_{t,t+1}}_{q^{x}_{t,t+1}} -  \sumlow{t}\BE{\log q^{x,f}_{t}}_{q^{x,f}_{t}} + \sumlow{t,j}\BE{\log q^{x}_{t,j}}_{q^{x}_{t,j}} -  \sumlow{t,j}\BE{\log q^{x,g}_{t,j}}_{q^{x,g}_{t,j}}.
\end{align*}
Note that if the the marginals in $\mathcal{Q}_{AZ}$ were exact, the entropy $\tilde{H}(\mathcal{Q}_{X})$ would also be exact.

To encode the expectation constraints in \eqref{EqnECX1} and \eqref{EqnECX2} we formulate the Lagrange multiplier terms 

\vspace{-0.15in}
{\small
\begin{align*}
	&C(\mathcal{Q}_{X}, \Lambda_{X}) =
								\\ & \quad  
									\sumlow{t}   \vect{\lambda}^{\beta}_{t+1} \cdot \BS{ \BE{\vect{f}(\vect{x}_{t+1})}_{q^{x,f}_{t+1}} - \BE{\vect{f}(\vect{x}_{t+1})}_{q^{x}_{t,t+1}} } 
									+ \sumlow{t}   \vect{\lambda}^{\alpha}_{t+1} \cdot \BS{ \BE{\vect{f}(\vect{x}_{t+1})}_{q^{x,f}_{t+1}} - \BE{\vect{f}(\vect{x}_{t+1})}_{q^{x}_{t+1,t+2}} } + \nonumber
								\\& \quad
								 \sumlow{t,j} \vect{\lambda}^{0}_{t+1,j}\cdot\BS{ \BE{\vect{g}(\vect{x}_{t+1})}_{q^{x,g}_{t+1,j}} - \BE{\vect{g}(\vect{x}_{t+1})}_{q^{x}_{t,t+1}} }
								 + \sumlow{t,j}	\vect{\lambda}^{l}_{t+1,j}\cdot \BS{\BE{\vect{g}(\vect{x}_{t+1})}_{q^{x,g}_{t+1,j}} - \BE{\vect{g}(\vect{x}_{t+1})}_{q^{x}_{t+1,j}}},
\end{align*}}
where $\vect{a} \cdot \vect{b}$ is shorthand for $\vect{a}^T \vect{b}$. We use $\Lambda_{X}$ to refer to the set of Lagrange multipliers 
$\big\{ \{ \vect{\lambda}^{\beta}_{t}\}_{t}, \{ \vect{\lambda}^{\alpha}_{t}\}_{t}, \{ \vect{\lambda}^{0}_{t,j}\}_{t,j}, \{ \vect{\lambda}^{l}_{t,j}\}_{t,j}  \big\}$.  Note that further multiplier constraints should be added to ensure that all approximate marginals normalise to $1$. However, since these are irrelevant to our derivation,  we omit them to keep the notation simple.

\subsubsection{Approximate marginals for $q_{AZ}$}

In the case of $q_{AZ}$ we have a multivariate conditional Gaussian model which is again tree-structured. For this reason, we use a set of approximate marginals
\begin{align*}
	\mathcal{Q}_{AZ} = \{ q^{A}_{0}, \{ q^{AZ}_{ij} \}_{ij}, \{ q^{A,g}_{ij} \}_{ij} \},
\end{align*}
where we associate $q^{A}_{0}$ with the exponentiated quadratic form in $q_{AZ}$ and we associate $q^{AZ}_{ij}(a_{ij}, z_{ij})$ with the prior $p_{0}(a_{ij}, z_{ij} \mid \vect{\theta})$.  The densities $\{§q^{A,g}_{ij} \}_{ij}$ are associated with the beliefs over the variables.

Similar to the approach presented in Section \ref{SecqX}, we introduce first- and second-order expectation constraints. That is, we choose $\vect{g}(a_{ij}) = (a_{ij}, -a_{ij}^2/2)$ and we postulate the constraints $\BE{\vect{g}(a_{ij})}_{q^{A,g}_{ij}} =\BE{\vect{g}(a_{ij})}_{q^{A}_{0}}$ and $\BE{\vect{g}(a_{ij})}_{q^{A,g}_{ij}} =\BE{\vect{g}(a_{ij})}_{q^{AZ}_{ij}}$. The approximate marginals in $\mathcal{Q}_{AZ}$ and the corresponding constraints are illustrated in Figure~\ref{FigAM}. 

The set of marginal densities in $\mathcal{Q}_{AZ}$ can be viewed to define a tree-structured density
\begin{align*}
	  q^{A}_{0}(\vect{A}) \prodlow{ij} \frac{q^{AZ}_{ij}(a_{ij}, z_{ij})} {q^{A,g}_{ij}(a_{ij})}.
\end{align*}
We define the corresponding entropy approximation as
\begin{align*}
	-\tilde{H}(\mathcal{Q}_{AZ}) = \BE{ \log q^{A}_{0}}_{q^{A}_{0}} + \sumlow{ij} \BE{ \log q^{AZ}_{ij}}_{q^{AZ}_{ij}} - \sumlow{ij} \BE{ \log q^{A,g}_{ij}}_{q^{A,g}_{ij}}.
\end{align*}
The Lagrangian terms for the expectation constraints can be written as
\begin{align*}
	C(\mathcal{Q}_{AZ}, \Lambda_{AZ}) = \sumlow{ij} \vect{\lambda}^{A,g}_{ij} \cdot \BS{ \BE{\vect{g}(a_{ij})}_{q^{A,g}_{ij}}  - \BE{\vect{g}(a_{ij})}_{q^{AZ}_{ij}}} 
										+ \sumlow{ij} \vect{\lambda}^{A,0}_{ij} \cdot \BS{ \BE{\vect{g}(a_{ij})}_{q^{A,g}_{ij}}  - \BE{\vect{g}(a_{ij})}_{q^{A}_{0}}}. 
\end{align*}
Note that when $\vect{Q}$ is diagonal, $q^{A}_{0}$ factorises. Therefore, the inference problem decouples for each row of $\vect{A}$ leading to $n$ sets of approximate marginals $\mathcal{Q}_{AZ}^{i}$ and  corresponding entropy approximations and Lagrangian terms respectively. 
Similarly to the previous section, we collect the Lagrange multipliers $\vect{\lambda}^{A,g}_{ij}$ and $\vect{\lambda}^{A,0}_{ij}$ in  $\Lambda_{AZ} = \big\{ \{ \vect{\lambda}^{A,g}_{ij}\}_{ij}, \{\vect{\lambda}^{A,0}_{ij}\}_{ij}\big\}$.

\subsubsection{Approximate marginals for $q_{B}$} \label{SecqB}
The distribution $q_{B}$ is that of a latent Gaussian model similar to $q_{AZ}$ with $\vect{z}$ marginalised. We define the set of approximate marginals
\begin{align*}
	\mathcal{Q}_{B} = \{ q^{B}_{0}, \{ q^{B}_{ij} \}_{ij}, \{ q^{B,g}_{ij} \}_{ij} \},
\end{align*}
assign $q^{B}_{0}$ to the exponentiated quadratic form in $q_{B}$ and $q^{B}_{ij}(b_{ij})$ to the prior $p_{0}(b_{ij} \mid \vect{\theta})$.  We define the entropy approximation and the expectation constraints in a similar fashion as in case of $q_{AZ}$. The entropy approximation reads as
\begin{align*}
	-\tilde{H}(\mathcal{Q}_{B}) = \BE{ \log q^{B}_{0}}_{q^{B}_{0}} + \sumlow{ij} \BE{ \log q^{B}_{ij}}_{q^{B}_{ij}} - \sumlow{ij} \BE{ \log q^{B,g}_{ij}}_{q^{B,g}_{ij}},
\end{align*}
and the Lagrangian term corresponding to the expectation constraints can be written as
\begin{align*}
	C(\mathcal{Q}_{B}, \Lambda_{B}) = \sumlow{ij} \vect{\lambda}^{B,g}_{ij} \cdot \BS{ \BE{\vect{g}(b_{ij})}_{q^{B,g}_{ij}}  - \BE{\vect{g}(b_{ij})}_{q^{B}_{ij}}} 
										+ \sumlow{ij} \vect{\lambda}^{B,0}_{ij} \cdot \BS{ \BE{\vect{g}(b_{ij})}_{q^{B,g}_{ij}}  - \BE{\vect{g}(b_{ij})}_{q^{B}_{0}}}. 
\end{align*}
The approximate marginals in $\mathcal{Q}_{B}$ and the corresponding constraints are illustrated in Figure~\ref{FigAM}. 

\subsubsection{Free energy optimisation and (approximate) message passing} \label{SecFreeEnergyOpt}
By using the families of approximate marginals and the corresponding entropy approximation defined above, we formulate a new variational objective based on \eqref{EqnExactFE} that is a function of the approximate marginals in $\mathcal{Q}_{X}$, $\mathcal{Q}_{AZ}$ and $\mathcal{Q}_{B}$. Both the expectation and the entropy terms in \eqref{EqnExactFE} are approximated by using approximate marginals.
By adding the corresponding constraints to this new objective,  we arrive to the Lagrangian
\begin{align}\label{EqnBigL}
	L(\mathcal{Q}_{X}, \mathcal{Q}_{AZ}, \mathcal{Q}_{Q}, \mathcal{Q}_{B}, &\: \Lambda_{X}, \Lambda_{AZ}, \Lambda_{B}) = 
				- \BE{\log p(\mathcal{Y}, \vect{X}, \vect{A}, \vect{Q}, \vect{B} \mid \vect{U} )}_{\mathcal{Q}_{X}, \mathcal{Q}_{A}, \mathcal{Q}_{Q}, \mathcal{Q}_{B}}
				\\ &
				- \tilde{H}(\mathcal{Q}_{X})  -  \tilde{H}(\mathcal{Q}_{A}) -  H(q_{Q}) - \tilde{H}(\mathcal{Q}_{B}) \nonumber
				\\ &
				+ C(\mathcal{Q}_{X}, \Lambda_{X}) + C(\mathcal{Q}_{A}, \Lambda_{A})  + C(\mathcal{Q}_{B}, \Lambda_{B}) + \text{\small normalisation constraints}. \nonumber
\end{align}
Taking expectations of the term $\log p(\vect{x}_{t+1} \vert \vect{x}_{t}, \vect{A}, \vect{B}, \vect{u}_t, \vect{Q})$ w.r.t $\mathcal{Q}_{X}$ means taking the expectations w.r.t. the corresponding member of $q_{t,t+1}^{x}$ of $\mathcal{Q}_{X}$, that is, 
$$\BE{ \log p(\vect{x}_{t+1} \vert \vect{x}_{t}, \vect{A},  \vect{B}, \vect{u}_t,\vect{Q})}_{\mathcal{Q}_{X}} = \BE{\log p(\vect{x}_{t+1} \vert \vect{x}_{t}, \vect{A},  \vect{B}, \vect{u}_t, \vect{Q})}_{q_{t,t+1}^{x}}.$$ Similarly, $$\BE{ \log p(\vect{x}_{t+1} \vert \vect{x}_{t}, \vect{A},  \vect{B}, \vect{u}_t,\vect{Q})}_{\mathcal{Q}_{AZ}}= \BE{\log p(\vect{x}_{t+1} \vert \vect{x}_{t}, \vect{A},  \vect{B}, \vect{u}_t, \vect{Q})}_{q_{0}^{A}}$$ and 
$\BE{ \log p(a_{ij}, z_{ij})}_{\mathcal{Q}_{AZ}} = \BE{ \log p(a_{ij}, z_{ij})}_{{q}^{AZ}_{ij}}$.

We perform the optimisation of the Lagrangian in $\eqref{EqnBigL}$ by (i) setting to zero its differentials w.r.t. the approximate marginals in $\mathcal{Q}_{X}, \mathcal{Q}_{AZ}, \mathcal{Q}_{Q}$, and $\mathcal{Q}_{B}$ and multipliers in $\Lambda_{X}, \Lambda_{AZ}, \Lambda_{B}$ and (ii) formulating a fixed point iteration based on these optimality conditions.
From the optimality conditions w.r.t the members of the families $\mathcal{Q}_{X}, \mathcal{Q}_{AZ}$ and $\mathcal{Q}_{B}$ we derive the form of the corresponding approximating densities. The  conditions corresponding to the expectation constraints are then used to define the fixed point iteration corresponding to the message-passing algorithm introduced in the main text.  The iteration typically converges to a (local) optimum and results in good quality approximations. When this is not the case, more elaborate optimisation algorithms such as the ones in \cite{HeskesZoeter2002}, \cite{Heskes2003} and  \cite{OpperWinther2005} can be applied to minimise the free energy. 

\begin{itemize}
\item[(1)] For the members of the set $\mathcal{Q}_{X}$ we proceed as follows. We take the differentials of $L$ w.r.t. the approximate marginals in $\mathcal{Q}_{X}$ and set them to zero. From this we obtain the approximate marginals
\begin{align}
	q^{x}_{t,t+1}(\vect{x}_{t}, \vect{x}_{t+1}) & \propto  \exp \Big \{  \BE{ \log\Psi_{t, t+1}(\vect{x}_{t}, \vect{x}_{t+1}) }_{\setminus \mathcal{Q}_{X}} \nonumber
	\\  & \quad 	+ \vect{\lambda}^{\alpha}_{t} \cdot \vect{f}(\vect{x}_{t}) + \vect{\lambda}^{\beta}_{t+1} \cdot \vect{f}(\vect{x}_{t+1}) + \sumlow{j} \vect{\lambda}^{0}_{t+1,j} \cdot \vect{g}(\vect{x}_{t+1})\Big\}, \nonumber
	\\
	q^{x}_{t+1, j}({x}_{t+1}^{j}) & \propto  \psi_{t+1,j}(x_{t+1}^{j}) \times \exp \{ \vect{\lambda}^{l}_{t+1,j} \cdot \vect{g} (x_{t+1}^{j})\}, \nonumber
	\\
	q^{x,g}_{t+1, j}({x}_{t+1}^{j}) & \propto   \exp \{( \vect{\lambda}^{0}_{t+1,j} + \vect{\lambda}^{l}_{t+1,j} ) \cdot \vect{g} ({x}_{t+1}^{j})\},	\label{EqnGM2}
	\\
	q^{x,f}_{t+1}(\vect{x}_{t+1}) & \propto   \exp \{( \vect{\lambda}^{\alpha}_{t+1} + \vect{\lambda}^{\beta}_{t+1} ) \cdot \vect{f}(\vect{x}_{t+1})\}.	  \label{EqnGM1}
\end{align}

We note that both $q^{x,g}_{t+1, j}({x}_{t+1}^{j}) $ and $q^{x,f}_{t+1}(\vect{x}_{t+1}) $ are Gaussian densities. We define the moment matching KL projection as
\begin{equation*}
	\text{Project}[p(\vect{z}); \vect{f}] = \mathop{\text{argmin}}\limits_{\theta} \D{}{p(\vect{z})}{ \exp(\vect{\theta}\cdot\vect{f}(\vect{z}) - \log Z(\vect{\theta}))}.
\end{equation*}
Note the slight notational difference between this definition and the $\text{Project}[\cdot; \mathcal{N}_{\vect{f}}]$ given in the main text: for convenience, here we operate on the canonical parameters, that is, 
for a given $p(\vect{z})$, $\text{Project}[p(\vect{z}); \vect{f}] $ is the (Gaussian) canonical parameter corresponding to distribution $\text{Project}[p(\vect{z}); \mathcal{N}_{\vect{f}}]$. $\text{Project}[p(\vect{z}); \vect{g}] $ is defined similarly by projecting to the univariate Gaussian defined by $\vect{g}(z) = (z, -z^2/2)$. Thus, it corresponds to the functional $\text{Project}[\tilde{q}_{t,t+1}(x_{t+1}^{j}); \mathcal{N}]$ and $\text{Project}[\tilde{q}_{t+1, j}(x_{t+1}^{j}); \mathcal{N}]$ in the main text.

By setting to zero the differentials w.r.t. the Lagrange multipliers in $\Lambda_{X}$, we recover the expectation constraints in \eqref{EqnECX1} and \eqref{EqnECX2}. Inspecting \eqref{EqnECX2} and noting that  \eqref{EqnGM2} is Gaussian, we rewrite the constraints as
\begin{align*}
\vect{\lambda}^{0}_{t+1,j} + \vect{\lambda}^{l}_{t+1,j}  = \text{Project}[q^{x}_{t+1, j}(\vect{x}_{t+1}) ; \vect{g}] \quad \text{and} \quad \vect{\lambda}^{0}_{t+1,j} + \vect{\lambda}^{l}_{t+1,j}  = \text{Project}[q^{x}_{t, t+1}(\vect{x}_{t+1}) ; \vect{g}]. 
\end{align*}
Similarly, using \eqref{EqnECX1} and and  \eqref{EqnGM1} we have
\begin{align*}
	\vect{\lambda}^{\alpha}_{t+1} + \vect{\lambda}^{\beta}_{t+1} = \text{Project}[q^{x}_{t,t+1}(\vect{x}_{t+1}) ; \vect{f}] ,  
	 \quad \text{and} \quad 
	\vect{\lambda}^{\alpha}_{t+1} + \vect{\lambda}^{\beta}_{t+1}= \text{Project}[q^{x}_{t+1,t+2}(\vect{x}_{t+1}) ; \vect{f}] .
\end{align*}

We use these optimality conditions to define the fixed point updates 
\begin{align}
	{[\vect{\lambda}_{t+1,j}^{g}]}^{new} &= \text{Project}[q^{x}_{t,t+1}(\vect{x}_{t+1}) ; \vect{g}] - \vect{\lambda}^{0}_{t+1,j}, \label{EqnSSfirst}
	\\
	{[\vect{\lambda}_{t+1,j}^{0}]}^{new} &= \text{Project}[q^{x}_{t+1, j}(\vect{x}_{t+1}) ; \vect{g}] - \vect{\lambda}^{l}_{t+1,j},	 
	\\
	{[\vect{\lambda}^{\alpha}_{t+1}]}^{new} &= \text{Project}[q^{x}_{t,t+1}(\vect{x}_{t+1}) ; \vect{f}] - \vect{\lambda}^{\beta}_{t+1}, 	
	\\
	{[\vect{\beta}^{\alpha}_{t+1}]}^{new} &= \text{Project}[q^{x}_{t+1,t+2}(\vect{x}_{t+1}) ; \vect{f}] - \vect{\lambda}^{\alpha}_{t+1}	, \label{EqnSSlast}	
\end{align}
which are equivalent to the approximate message passing updates defined in Section 3 of the paper. Note that these updates are expressed in terms of canonical parameters as opposed to the functional formulation in the paper. For example, $\vect{\lambda}_{t+1}^{\alpha}$ and $\vect{\lambda}_{t+1}^{\beta}$ are the canonical parameters of the forward and backward messages $\alpha_{t+1}(\vect{x}_{t+1})$ and $\beta_{t+1}(\vect{x}_{t+1})$, respectively. The matrix algebraic operations used to perform these updates are detailed in the main text. 

\item[(2)] For the members of the set $\mathcal{Q}_{AZ}$ setting the differentials w.r.t. approximate marginals yields
\begin{align}
	q^{A}_{0}([\vect{A}]_{c}) &\propto  \exp \Big \{ [\vect{A}]_{c}^{T}\BE{\vect{h}_{A}}_{\setminus \mathcal{Q}_{A}} - \frac{1}{2}[\vect{A}]_{c}^{T}\BE{\vect{Q}_{A}}_{\setminus \mathcal{Q}_{A}} [\vect{A}]_{c} + \sumlow{ij}\vect{\lambda}^{A,0}_{ij} \cdot \vect{g}(a_{ij}) \Big\}, \label{EqnMixQ0}
	\\
	q^{AZ}_{ij}(a_{ij}, z_{ij}) &\propto p_{0}(a_{ij}, z_{ij} \mid \vect{\theta}) \times \exp\BC{ \vect{\lambda}^{A,g}_{ij} \cdot \vect{g}(a_{ij})}, \label{EqnMixUni}
	\\
	q^{A,g}_{ij}(a_{ij}) &\propto \exp\BC{( \vect{\lambda}^{A,0}_{ij} + \vect{\lambda}^{A,g}_{ij}) \cdot \vect{g}(a_{ij}) }. \nonumber
\end{align}
Following the same arguments as above we arrive to the update equations
\begin{align}
	{[\vect{\lambda}^{A,g}_{ij}]}^{new} &= \text{Project}[q^{A}_{0}(a_{ij}) ; \vect{g}] - \vect{\lambda}^{A,0}_{ij}, \label{EqnMixEP2}
	\\
	{[\vect{\lambda}^{A,0}_{ij}]}^{new} &= \text{Project}[q^{AZ}_{ij}(a_{ij}) ; \vect{g}] - \vect{\lambda}^{A,g}_{ij}. \label{EqnMixEP1}
\end{align}
We compute the update as follows. (i) In \eqref{EqnMixEP2} we compute the marginal mean and variance parameters of the multivariate Gaussian distribution in \eqref{EqnMixQ0}, compute the corresponding univariate (Gaussian) canonical parameters and perform the update. (ii) In \eqref{EqnMixEP1} we compute the mean and variance of the marginals $q^{AZ}_{ij}(a_{ij})$ by summing out  $z_{ij}$ in  \eqref{EqnMixUni}. We then compute the corresponding univariate (Gaussian) canonical parameters and perform the update. The complexity is dominated by the partial matrix inversion needed to compute the marginal variances in \eqref{EqnMixQ0}. 

\item[(3)] For the members of the set $\mathcal{Q}_{B}$ we have
\begin{align}
	q^{B}_{0}([\vect{B}]_{c}) &\propto  \exp \Big \{ [\vect{B}]_{c}^{T}\BE{\vect{h}_{B}}_{\setminus \mathcal{Q}_{B}} - \frac{1}{2}[\vect{B}]_{c}^{T} \BE{\vect{Q}_{B}}_{\setminus \mathcal{Q}_{B}} [\vect{B}]_{c} + \sumlow{ij}\vect{\lambda}^{B,0}_{ij} \cdot \vect{g}(b_{ij}) \Big\}, \nonumber
	\\
	q^{B}_{ij}(b_{ij}) &\propto p_{0}(b_{ij} \mid \vect{\theta}) \times \exp\BC{ \vect{\lambda}^{B,g}_{ij} \cdot \vect{g}(b_{ij})},  \nonumber
	\\
	q^{B,g}_{ij}(b_{ij}) &\propto \exp\BC{( \vect{\lambda}^{B,0}_{ij} + \vect{\lambda}^{B,g}_{ij}) \cdot \vect{g}(b_{ij}) },  \nonumber
\end{align}
and the corresponding expectation constraints lead to the update equations
\begin{align}
	{[\vect{\lambda}^{B,g}_{ij}]}^{new} &= \text{Project}[q^{B}_{0}(b_{ij}) ; \vect{g}] - \vect{\lambda}^{B,0}_{ij}. \label{EqnBEP1}
	\\
	{[\vect{\lambda}^{B,0}_{ij}]}^{new} &= \text{Project}[q^{B}_{ij}(b_{ij}) ; \vect{g}] - \vect{\lambda}^{B,g}_{ij}. \label{EqnBEP2}
\end{align}
Computations of these updates are similar to those of $\mathcal{Q}_{AZ}$, with the difference that here the we use univariate quadrature when  \eqref{EqnBEP2} is analytically intactable. 
\end{itemize}

We run the fixed point updates according to the structured variational Bayes approach, that is, for each set of marginals $\mathcal{Q}_{X}$, $\mathcal{Q}_{AZ}$, $\mathcal{Q}_{B}$ we run the corresponding updates till convergence and then we proceed to the next set of approximate marginals. In other words, (i) we run \eqref{EqnSSfirst}--\eqref{EqnSSlast} till convergence according to the various scheduling options described in the main text, then (ii) run \eqref{EqnMixEP2}--\eqref{EqnMixEP1} till convergence or do the exact marginalisation in case of diagonal $\vect{Q}$ (see main text)  (iii) we run  \eqref{EqnBEP1}--\eqref{EqnBEP2} till convergence, and finally (iv) we update $q_{Q}$ . We repeat the cycle (i)-(iv) till convergence is achieved. 


This particular scheduling of the updates was chosen to correspond to the variational Bayes approach. However, other scheduling choices might be more efficient, for example, the extension of the dynamic scheduling proposed in the main text to all the updates \eqref{EqnSSfirst}-\eqref{EqnSSlast}, \eqref{EqnMixEP2}-\eqref{EqnMixEP1} and  \eqref{EqnBEP1}-\eqref{EqnBEP2}, can be considered as a good starting point. 

Note that when the inference can be done exactly as it is the case for $q_{AZ}^{i}$ or Gaussian  $p(b_{ij} \mid \vect{\theta})$, then the projection operations are simply replaced by the corresponding free form variational update as presented in Section~\ref{SecVarInf}. This corresponds to replacing the approximate entropies with the exact ones and  omitting the introduction of the families of approximate marginals and the corresponding expectation constraints.  

\subsection{Notes on convergence}

Convergence of the proposed algorithm is not guaranteed.  However, it has been shown that when the non-Gaussian terms $\psi_{t,j}(x_{t}^{j})$ are log-concave (and $q_{X}$ is unimodal), then the block version of the model for $\mathcal{Q}_{X}$ does not diverge \citep{Seeger2008}. Given the implicit links between the blocked and distributed/dynamic inference approaches, this property should also hold for the method we present in the paper.  A similar comment can be made for the distributions in  $\mathcal{Q}_{B}$ when $p(b_{ij} \mid \vect{\theta})$ are log-concave. 

The approximate inference method for $q_{AZ}$ is known to be less robust; it has, however, been applied successfully in several situations such as robust regression \citep{CsatoMIXTURE2005} and variables selection  \citep[e.g.,][]{Lobato2013}. Robustness is aided when using techniques designed to stabilise the fixed point update in \eqref{EqnMixEP2} and \eqref{EqnMixEP1}. In one such technique, known as damping \citep[e.g.,][]{Heskes2004a}, for $\epsilon \in [0, 1]$ one replaces the update in \eqref{EqnMixEP1} with the update 
\begin{align*}
	{[\vect{\lambda}^{A,0}_{ij}]}^{new} &= (1-\epsilon){[\vect{\lambda}^{A,0}_{ij}]} +  \epsilon \big(\text{Project}[q^{AZ}_{ij}(a_{ij}) ; \vect{g}] - \vect{\lambda}^{A,g}_{ij}\big).
\end{align*}

Another method employs a modified entropy approximation and, for $\eta \in [0,1]$, results in the updates  
\begin{align*}
	{[\vect{\lambda}^{A,0}_{ij}]}^{new} &= \frac{1}{\eta} \big( \text{Project}[q^{AZ}_{ij}(a_{ij}) ; \vect{g}] -  \vect{\lambda}^{A,g}_{ij} \big),
	\\
	{[\vect{\lambda}^{A,g}_{ij}]}^{new} &= \text{Project}[q^{A}_{0}(a_{ij}) ; \vect{g}] - \eta \vect{\lambda}^{A,0}_{ij}.
\end{align*}
This technique is related to fractional belief, power expectation  propagation and convexified entropy approximations \citep{WiegHesk2003, WainJaakWill2003, Minka2004, Heskes:2006}. If none helps we can replace the spike and slab prior with other sparsity promoting priors such as the half-Cauchy prior \citep[e.g.,][]{Armagan2011}. The resulting variational inference algorithm  fits naturally into our framework. 

The only convergence problems we encountered for the model presented in the main text pertained to the {\it diag} method with $\mathcal{Q}_{X}$.  In some models with strong correlations in $\tilde{q}_{t,t+1}(\vect{x}_t, \vect{x}_{t+1})$  we applied a damping of $\epsilon=0.9$ for the temporal messages $\alpha_t$ and $\beta_t$ to improve convergence. Generally we experience fast convergence for all expectation propagation algorithms for the various models (e.g., $q_X$ and $q_{AZ}$). It seems that the (known) slow convergence of the  variational Bayes approach dominates the overall convergence when updating the forms of $q_{X}, q_{AZ}, q_{B}$ and $q_{Q}$.


\section{Scheduling options}
\begin{figure}
	\begin{center}
		\resizebox{!}{0.15\textheight}{\includegraphics{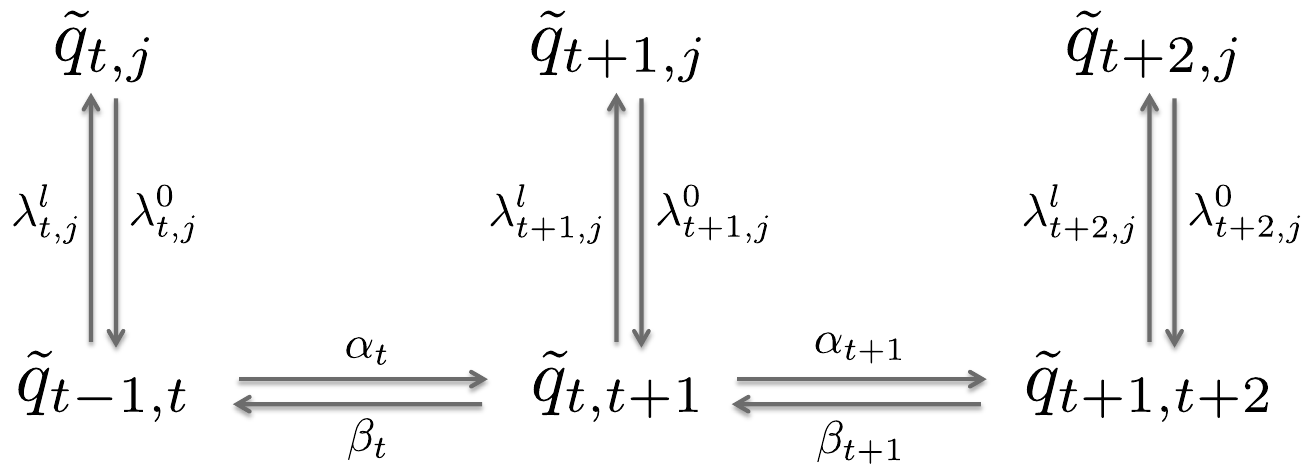}}
	\end{center}
	\caption{\small An illustration of the approximate marginals and  messages.} 
	\label{FigSched}
\end{figure}

Figure~\ref{FigSched} illustrates the approximate marginals in $\mathcal{Q}_{X}$ and the messages used in the corresponding message-passing algorithm. The scheduling options from the main text can be summarised as follows.

\subsubsection*{Dynamic scheduling}
\begin{itemize}
	\item[(i)] pick the message with the largest last update in absolute value from all $\alpha_{t+1}, \beta_{t}, \lambda^{l}_{t+1, j}$ and $\lambda^{0}_{t+1, j}$ 
	\item[(ii)] choose from the following steps
		\begin{itemize}
			\item if $\lambda^{0}_{t+1, j}$ is chosen 
				\begin{itemize}
					\item[-] compute  $\tilde{q}_{t,t+1}$  and  update $\lambda_{t+1, j}^{l}$, $\alpha_{t+1}$ and $\beta_t$  
					\item[-] compute $\tilde{q}_{t+1, j}$  and update $\lambda_{t+1, j}^{0}$
				\end{itemize}	
			\item if $\lambda^{l}_{t+1, j}$ is chosen, 
				\begin{itemize}
					\item[-] compute $\tilde{q}_{t+1, j}$  and update  $\lambda_{t+1, j}^{0}$ 
					\item[-] compute $\tilde{q}_{t,t+1}$  and  update $\lambda_{t+1, j}^{l}$, $\alpha_{t+1}$ and $\beta_t$  
				\end{itemize}
			\item if $\alpha_{t+1}$ is chosen
				\begin{itemize}
					\item[-] compute $\tilde{q}_{t+1,t+2}$  and update $\lambda_{t+2, j}^{l}$, $\alpha_{t+2}$ and $\beta_{t+1}$  
					\item[-] compute $\tilde{q}_{t,t+1}$  and update $\lambda_{t+1, j}^{l}$, $\beta_{t}$ and  $\alpha_{t+1}$ 
				\end{itemize}	
			\item if $\beta_{t}$ is chosen
				\begin{itemize}
					\item[-] compute $\tilde{q}_{t-1,t}$  and update $\lambda_{t, j}^{l}$,  $\beta_{t-1}$ and $\alpha_{t}$  
					\item[-] update $\tilde{q}_{t,t+1}$  and update $\lambda_{t+1, j}^{l}$, $\alpha_{t+1}$ and $\beta_{t}$ 
				\end{itemize}	
		\end{itemize}
	\item[(iii)] repeat (i) and (ii) until the largest update in step (i) is below a threshold
\end{itemize}

\subsubsection*{Sequential scheduling}
\begin{itemize}
	\item[(i)] run until convergence: compute $\tilde{q}_{t,t+1}$ and update $\lambda^{l}_{t+1, j}$ for all $j \in \{1, \ldots, n\}$,  compute $\tilde{q}_{t+1, j}$ and update $\lambda^{0}_{t+1, j}$
	\item[(ii)] update the $\alpha_{t+1}$ or $\beta_t$ message depending on whether doing a forward or a backward step 
	\item[(iii)] repeat (i) and (ii) in a forward-backward fashion  till the absolute or relative update in all $\alpha_{t+1}, \beta_{t}, \lambda^{l}_{t+1, j}$ and $\lambda^{0}_{t+1, j}$ is below a threshold
\end{itemize}
Note that after the initial forward sweep, the iteration in step (i) converges in a few steps.

\subsubsection*{Static Scheduling}
\begin{itemize}
	\item[(i)] run till convergence in a forward-backward fashion: compute $\tilde{q}_{t,t+1}$ and update $\alpha_{t+1}, \beta_{t}$ and $\lambda^{l}_{t+1, j}$; updating backward messages $\beta_t$ in a forward iteration is not necessary and likewise for the forward messages $\alpha_{t+1}$ in the backward iteration
	\item[(ii)] for all $j \in \{1, \ldots, n\}, t \in \{1, \ldots, T\} $ update $\tilde{q}_{t, j}$ and $\lambda^{0}_{t, j}$
	\item[(iii)] repeat (i) and (iii) till the absolute or relative update in all $\alpha_{t+1}, \beta_{t}, \lambda^{l}_{t+1, j}$ and $\lambda^{0}_{t+1, j}$ is below a threshold
\end{itemize}


\bibliographystyle{plainnat}
\bibliography{jasa-saistpp-arxiv}

\end{document}